\documentclass[journal]{IEEEtran}
\usepackage{graphicx}
\usepackage{amsmath, amssymb}
\DeclareMathOperator*{\argmax}{arg\,max}
\usepackage{caption, subcaption}
\usepackage{multirow}
\usepackage{soul,color}
\usepackage{cite}
\usepackage{hyperref}
\begin{document}
%%%%%%%%%%%%%%%%%%%%%%%%%%%%%
\title{Transformer-Driven Active Transfer Learning for Cross-Hyperspectral Image Classification}
%%%%%%%%%%%%%%%%%%%%%%%%%%%%%%%
\author{Muhammad Ahmad, Francesco Mauro, Manuel Mazzara, Salvatore Distefano, Adil Mehmood Khan, Silvia Liberata Ullo
\thanks{M. Ahmad is with SDAIA-KFUPM, Joint Research Center for Artificial Intelligence (JRCAI), King Fahd University of Petroleum and Minerals, Dhahran, 31261, Saudi Arabia. (e-mail: mahmad00@gmail.com).}
\thanks{F. Mauro is with the Engineering Department, University of Sannio, Benevento 82100, Italy.}
\thanks{M. Mazzara is with the Institute of Software Development and Engineering, Innopolis University, Innopolis, 420500, Russia.}
\thanks{S. Distefano is with Dipartimento di Matematica e Informatica-MIFT, University of Messina, 98121 Messina, Italy.}
\thanks{A.M. Khan is with the School of Computer Science, University of Hull, Hull HU6 7RX, UK.}
\thanks{S. L. Ullo is with the Engineering Department, University of Sannio, Benevento 82100, Italy.}
}
%%%%%%%%%%%%%%%%%%%%%%%%%%%%%%%
\markboth{Journal of \LaTeX\ Class Files}
{Ahmad \MakeLowercase{\textit{et al.}}}
%%%%%%%%%%%%%%%%%%%%%%%%%%%%%%%
\maketitle
%%%%%%%%%%%%%%%%%%%%%%%%%%%%%
\begin{abstract}
Hyperspectral image (HSI) classification presents inherent challenges due to high spectral dimensionality, significant domain shifts, and limited availability of labeled data. To address these issues, we propose a novel Active Transfer Learning (ATL) framework built upon a Spatial-Spectral Transformer (SST) backbone. The framework integrates multistage transfer learning with an uncertainty-diversity-driven active learning mechanism that strategically selects highly informative and diverse samples for annotation, thereby significantly reducing labeling costs and mitigating sample redundancy. A dynamic layer freezing strategy is introduced to enhance transferability and computational efficiency, enabling selective adaptation of model layers based on domain shift characteristics. Furthermore, we incorporate a self-calibrated attention mechanism that dynamically refines spatial and spectral weights during adaptation, guided by uncertainty-aware feedback. A diversity-promoting sampling strategy ensures broad spectral coverage among selected samples, preventing overfitting to specific classes. Extensive experiments on benchmark cross-domain HSI datasets demonstrate that the proposed SST-ATL framework achieves superior classification performance compared to conventional approaches. The source code is publicly available at \url{https://github.com/mahmad000/ATL-SST}.
\end{abstract}
%%%%%%%%%%%%%%%%%%%%%%%%%%%%%
%%%%%%%%%%%%%%%%%%%%%%%%%%%%%
\section{Introduction}

\IEEEPARstart{H}{yperspectral} images (HSIs) capture detailed spectral signatures across hundreds of contiguous and narrow wavelength bands, offering rich information that enables accurate classification and identification of diverse materials and land cover types \cite{6555921}. This fine spectral resolution has established HSIs as a vital tool across numerous domains, including remote sensing \cite{BUTT2024103773}, environmental monitoring \cite{stuart2022high}, food quality and safety inspection \cite{saleem2020prediction, khan2021hyperspectral, ayaz2020hyperspectral}, forensic analysis \cite{9870277}, and biomedical imaging \cite{zulfiqar2021hyperspectral, 103389}. However, despite these widespread applications, achieving robust HSI classification remains a formidable challenge. The high dimensionality of spectral data, the limited availability of labeled samples, and significant spectral-spatial variability due to varying acquisition conditions collectively hinder generalization performance \cite{HAJAJ2024101218, GUERRI2024100062}. These issues become particularly pronounced in cross-domain classification tasks, where domain shifts between datasets severely limit model transferability.

In recent years, transformer-based architectures have garnered increasing interest for HSI classification \cite{10681622, 10472541, Ahmad03042025}, owing to their powerful self-attention mechanisms that capture long-range spectral and spatial dependencies. This capability leads to richer and more discriminative feature representations \cite{10571998, 11037730}. Spatial-Spectral Transformers (SSTs), in particular, have emerged as a promising alternative to traditional convolutional neural networks (CNNs). SSTs can model intricate spectral-spatial interactions by operating on image patches, and inherently scale well to high-resolution HSI data \cite{10379176, 10400402}. Unlike CNNs, SSTs eliminate the need for handcrafted feature extractors or complex pooling strategies by learning hierarchical representations directly from raw pixels \cite{9627165}. Furthermore, their interpretability via attention maps enhances trust in model decisions by highlighting salient regions that drive classification outcomes \cite{10387571, Ahmad30062025}.

Building on this foundation, several SST-based models have been proposed better to harness the spectral and spatial properties of HSIs. For instance, Shi et al. \cite{10384612} introduced a dual-branch transformer architecture that separately captures spectral and spatial dependencies, allowing for more effective multi-scale contextual learning. Nevertheless, deploying SSTs for large-scale HSI tasks presents several challenges. First, the self-attention mechanism incurs quadratic computational complexity with respect to sequence length, posing scalability concerns for longer inputs \cite{huang2024sparse}. Second, unlike CNNs, SSTs lack built-in translation invariance, which may affect the consistency of spatial feature extraction \cite{sun2024image}. Third, tokenizing HSIs into fixed-size patches can result in the loss of fine-grained information, particularly in small or irregular structures. Moreover, SSTs typically rely on large amounts of labeled data to achieve strong generalization, making them vulnerable to overfitting under low-label regimes \cite{9874815}. Given the high cost and effort involved in acquiring labeled HSI samples \cite{JIA2021179, 1011453472291}, this reliance on fully supervised learning is often impractical. Consequently, strategies that leverage both labeled and unlabeled data, such as semi-supervised and active learning (AL), have become increasingly important for reducing annotation demands while maintaining classification accuracy \cite{AHMAD2020163712,101007}.

AL, in particular, has gained significant traction in the HSI community as a means of selectively labeling the most informative samples \cite{9774342, 10647404}. A number of effective approaches have been introduced to address this challenge \cite{Karaca18042025}. Liu et al. \cite{9186822} proposed MDL4OW, a multitask framework that combines classification and reconstruction to detect unknown classes via reconstruction errors. Liao et al. \cite{10231146} introduced CGE-AL, a graph-embedding-based AL method that optimizes sample selection using graph CNNs and uncertainty estimation. Zhao et al. \cite{10167502} designed MAT-ASSAL, which integrates a multi-attention transformer with adaptive superpixel segmentation to preserve spatial structures during query selection. Similarly, Wang et al. \cite{10012443} presented a co-auxiliary learning strategy, coupling pseudo-labeling with diverse sample selection to ensure both uncertainty and representativeness. Liu et al. \cite{10268909} further advanced the field by introducing an adversarial domain alignment framework that incorporates contrastive learning, enhancing robustness and cross-domain generalization under few-shot conditions.

Despite these promising directions, deep AL methods still face key limitations when applied to HSIs. The inherent scarcity of labeled data constrains the diversity of queried samples, often resulting in suboptimal generalization \cite{10695121}. Additionally, complex spectral-spatial correlations present in HSIs are not always effectively captured by conventional AL techniques \cite{10423035}. To address these gaps, the research community has started to explore active transfer learning (ATL) frameworks that combine the strengths of AL and transfer learning. In such frameworks, knowledge learned from well-labeled source domains can be transferred to improve sample selection and feature extraction in sparsely labeled target domains. This approach is especially valuable in dynamic remote sensing scenarios, where spectral characteristics can vary substantially over time and location. By enabling domain adaptation with minimal retraining, ATL offers a viable pathway to enhance HSI classification with limited annotation overhead.

Motivated by these insights, this work introduces an SST-based ATL framework (SST-ATL) specifically designed for cross-domain HSI classification. The proposed SST-ATL framework directly tackles critical issues such as limited label availability, spectral-spatial heterogeneity, and domain shifts, providing a comprehensive and adaptive solution. The key contributions of this work are summarized below:

We propose \textbf{SST-ATL}, a unified ATL framework for HSI classification that integrates an SST with a multistage fine-tuning strategy and a dynamic layer freezing mechanism. This design allows for efficient cross-domain adaptation under annotation scarcity by selectively updating model layers based on domain shift characteristics, thereby enhancing generalization while reducing computational overhead. To further improve robustness against spectral variability, a novel self-calibrated attention refinement module is introduced, which dynamically adjusts spatial and spectral attention weights using uncertainty-guided feedback. In addition, we develop a \textbf{hybrid query strategy} that jointly optimizes informativeness, diversity, and spectral coverage through a combined uncertainty-diversity sampling criterion and a spectral diversity-enforcing selection scheme. Extensive evaluations across six standard HSI benchmarks demonstrate that SST-ATL consistently outperforms CNN-, Transformer-, and state-space model-based baselines by \textbf{+2.5\% to +3.8\% OA} under constrained annotation budgets. The complete implementation is publicly available at \url{https://github.com/mahmad000/ATL-SST}.

%%%%%%%%%%%%%%%%%%%%%%%%%%%%%
\section{Related Works}

Recent advancements in HSI classification have explored several paradigms, including transformer-based models, AL, and their integration into ATL. We categorize and review the literature under these themes. Table \ref{SOTA} summarizes each method’s key contributions, limitations, and the improvements offered by our proposed SST-ATL framework.

\paragraph{Transformer-Based HSI Classification} Transformer-based models, particularly SSTs, have shown exceptional capacity in modeling long-range spectral-spatial dependencies. S2FTNet \cite{10163753} introduces a dual-branch SST using separate spectral and spatial attention modules, achieving 3–5\% gains over CNNs. Zhang et al. \cite{10196016} propose ELS2T, a resource-efficient SST variant with reduced embedding dimensions. Zhang et al. \cite{ZHANG2023104983} further design a multi-range SST that attends to spectral-spatial intervals at multiple scales. Ahmad et al. \cite{10399798} tackle spectral redundancy using a wavelet-based compression (WaveFormer), which reduces FLOPs while preserving spectral detail. Despite these innovations, most transformer-based methods require extensive labeled data and struggle with domain shifts. These limitations motivate the integration of adaptive learning strategies such as AL and ATL to improve generalization under low-label regimes.

%%%%%%%%%%%%%%%%%%%%%%%%%%%%%
\begin{table*}[!hbt]
\centering
\caption{Comparative Summary of Recent HSI Classification Approaches.}
\resizebox{\textwidth}{!}{\begin{tabular}{p{2.0cm}|p{4.5cm}|p{4.8cm}|p{4.9cm}} \hline
\textbf{Method} & \textbf{Key Contributions} & \textbf{Limitations} & \textbf{Proposed SST-ATL} \\ \hline

Shi et al. \cite{10384612} & Dual-branch SST for decoupled spectral-spatial attention and multi-scale context modeling & Quadratic attention cost, lack of translation invariance, and overfitting risk under low labels & Lightweight SST, dynamic freezing for low-label adaptation, and computationally efficient \\ \hline 

Zhang et al. \cite{ZHANG2023104983} & Multi-range attention with heads attending to different spectral-spatial scales & Fusion overhead and redundant computation across branches & Self-calibrated attention mechanism with uncertainty guidance optimizes attention allocation \\ \hline 

Ahmad et al. \cite{10399798} & Wavelet-based downsampling before SST to reduce spectral redundancy & Optimal wavelet scale selection critical; limited adaptivity to scene heterogeneity & Integrates dynamic spectral diversity sampling with reversible compression \\ \hline 

Liao et al. \cite{10231146} & Graph embedding-based AL with class-wise sample selection via uncertainty in graph parameters & Graph construction is expensive and scales poorly; lacks spectral adaptivity & SST-ATL avoids explicit graphs and embeds spectral–spatial diversity in query mechanism \\ \hline 

Zhao et al. \cite{10167502} & Multi-attention transformer with adaptive superpixels for active querying & Relies heavily on superpixel quality; segmentation biases impact results & Patch-free querying driven by calibrated attention and uncertainty \\ \hline 

Wang et al. \cite{10012443} & Co-auxiliary learning with pseudo-labeling and uncertainty-based diversity & Requires training auxiliary models; complexity grows with iterations & No auxiliary network needed; SST-ATL optimizes a unified loss with query feedback \\ \hline 

Liu et al. \cite{10268909} & Adversarial domain alignment using contrastive loss to enhance cross-domain generalization & Training instability due to adversarial components; contrastive loss tuning is sensitive & Domain adaptation via dynamic freezing and progressive fine-tuning improves robustness \\ \hline 

Di et al. \cite{rs15030752} & Dual-module AL with adversarial and inter-class uncertainty sampling & Increased training complexity and instability in contrastive pairing under limited labels & Simpler and stable sample selection via joint uncertainty–diversity strategy \\ \hline 

Zhang et al. \cite{10196016} & Efficient SST (ELS2T) with reduced heads and embeddings for resource-limited settings & Reduced capacity on complex datasets; performance drops in high-class overlap & SST-ATL retains model expressivity while compressing via reversible transforms \\ \hline 

Deng et al. \cite{8520902} & First ATL framework using sparse autoencoders and active querying & Based on outdated autoencoder architecture; lacks attention-based adaptability & Transformer-based ATL with uncertainty feedback and fine-grained spectral focus \\ \hline 

Lin et al. \cite{8531707} & AL-driven transfer learning without prior target knowledge & Evaluation limited to three datasets; weak robustness to large domain shifts & Evaluated across diverse benchmarks; freezing strategy tailored to shift magnitude \\ \hline 

Ahmad et al. \cite{9903062} & Deep AL with 3D CNN and query function using fuzziness, Breaking ties, and MI & Computational cost of 3D CNNs high; lacks long-range dependency modeling & SST-ATL models long-range features with fewer parameters using hierarchical attention \\ \hline 

Yang et al. \cite{8100722} & Conceptual integration of AL and transfer learning for label efficiency & High-level design; lacks architectural or query function details & SST-ATL provides end-to-end framework with defined transfer, freezing, and query modules \\ \hline 

Cao et al. \cite{8978543} & CNN + AL with MRF smoothing for better spatial consistency & Lacks global attention modeling and depends on fixed post-processing & Self-attention offers adaptive spatial consistency without external smoothers \\ \hline 

He et al. \cite{8913592} & Heterogeneous transfer learning from RGB to HSI with mapping and attention & Pretrained RGB CNNs underutilize spectral richness; weak generalization under large shifts & SST-ATL learns spectral-attentive features directly from HSI with built-in domain adaptation \\ \hline
\end{tabular}}
\label{SOTA}
\end{table*}
%%%%%%%%%%%%%%%%%%%%%%%%%%%%%

\paragraph{Active Learning Strategies for HSI} AL has been leveraged to reduce annotation costs by prioritizing informative sample selection. Zhao et al. \cite{10167502} use adaptive superpixel segmentation with a multi-attention transformer for sample querying. Liao et al. \cite{10231146} propose class-wise graph-based uncertainty sampling via graph embeddings, though with high computational costs. Wang et al. \cite{10012443} combine pseudo-labeling with co-auxiliary learning to improve sample representativeness. Di et al. \cite{rs15030752} propose ALSN, an adversarially trained Siamese network employing dual uncertainty criteria (AUAL, ICUAL), showing strong label efficiency but with increased model complexity. Cao et al. \cite{8978543} integrate Markov Random Field smoothing with CNN-based AL to improve spatial coherence, while Ahmad et al. \cite{9903062} adopt multiple sample selection criteria, including fuzziness, mutual information, and breaking ties in a 3D CNN setting. These methods, while effective, often rely on heuristic sampling or are computationally intensive.

\paragraph{Active Transfer Learning for HSI} The fusion of AL and transfer learning forms ATL frameworks, aimed at learning transferable representations under limited labels. Deng et al. \cite{8520902} were among the first to propose ATL for HSI using stacked sparse autoencoders. Lin et al. \cite{8531707} extend this by introducing an AL-driven transfer learning pipeline that does not require prior knowledge of the target domain. Yang et al. \cite{8100722} provide a conceptual overview of ATL for HSI, but lack architectural detail. He et al. \cite{8913592} focus on heterogeneous ATL from RGB to HSI using pretrained CNNs, but struggle with spectral misalignment. Our previous work \cite{9903062} shows that ATL with 3D CNNs can achieve strong results, but at the cost of computational complexity and limited domain adaptation flexibility. 

In contrast, our SST-ATL framework offers a principled fusion of SSTs, AL, and ATL by incorporating dynamic layer freezing, uncertainty-calibrated attention, and hybrid uncertainty-diversity querying, providing a robust and computationally efficient solution to cross-domain HSI classification.

%%%%%%%%%%%%%%%%%%%%%%%%%%%%%%%
\begin{figure*}[!hbt]
    \centering
    \includegraphics[width=0.99\linewidth]{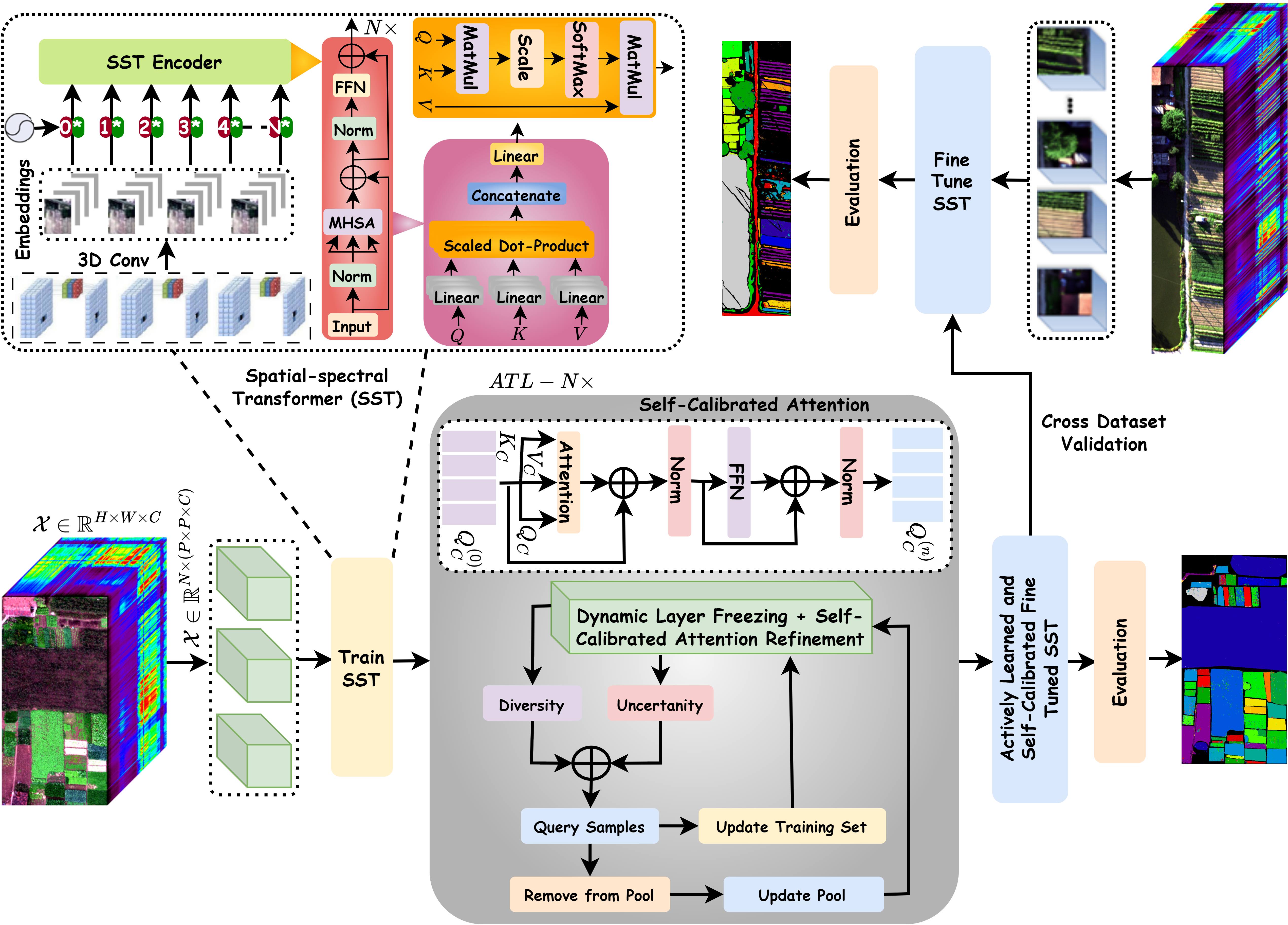}
    \caption{Overview of the proposed ATL-SST framework for HSI classification. The model first trains an SST encoder by extracting 3D patch embeddings and learning spatial-spectral representations via multi-head self-attention (MHSA). AL iteratively refines the training set by querying the most informative samples based on uncertainty and diversity criteria, while dynamic layer freezing and self-calibrated attention refinement enhance model generalization and robustness. A fine-tuned SST model is evaluated to achieve efficient and accurate cross-dataset classification with minimal labeled data.}
    \label{Model}
\end{figure*}
%%%%%%%%%%%%%%%%%%%%%%%%%%%%%%%

%%%%%%%%%%%%%%%%%%%%%%%%%%%%%%%
\section{Proposed Methodology}

Given an HSI cube $\mathbf{X} \in \mathbb{R}^{M \times N \times k}$, where $M$ and $N$ are the spatial dimensions and $k$ is the number of spectral bands, we first extract overlapping spatial windows of size $W \times W$ from $\mathbf{X}$. These windows provide localized spatial-spectral context for feature encoding via a transformer-based model. Each window is further divided into smaller subpatches used for embedding.

%%%%%%%%%%%%%%%%%%%%%%%%%%%%%%%
\subsection{Spectral-Spectral Transformer (SST)}

The SST backbone comprises four stages: patch embedding, positional encoding, self-attention-based encoder blocks, and the final classification head.

\paragraph{\textbf{Patch Embedding}} Each spatial window of size $W \times W \times k$ is processed by a 3D convolutional layer to extract $p \times p$ subpatches. Let $\mathbf{W}_e$ be the learnable convolutional kernel bank. The $i$-th embedded subpatch $\mathbf{P}_i$ is computed as:

\begin{equation}
\mathbf{P}_i(u,v,c) = \sum_{m=0}^{p-1} \sum_{n=0}^{p-1} \sum_{q=0}^{k-1} \mathbf{X}(s_u+m, s_v+n, q) \cdot \mathbf{W}_e(m,n,q,c)
\end{equation}
where $(s_u, s_v)$ are the coordinates of the $i$-th subpatch, $p$ is the subpatch size, $c$ indexes output channels, and $N_p = \left(\frac{W}{p}\right)^2$ is the number of subpatches in a window. These subpatches are projected into a $d$-dimensional feature space.

\paragraph{\textbf{Positional Encoding}} To inject spatial information into the embeddings, a sinusoidal positional encoding $\mathbf{E}_{\text{pos}}$ is added:

\begin{align}
\mathbf{E}_{\text{pos}}(i, 2j) &= \sin\left(\frac{i}{10000^{2j/d}}\right) \\
\mathbf{E}_{\text{pos}}(i, 2j+1) &= \cos\left(\frac{i}{10000^{2j/d}}\right)
\end{align}

The encoded patch sequence is $\mathbf{Z} = \mathbf{P} + \mathbf{E}_{\text{pos}}$.

\paragraph{\textbf{Transformer Encoder Block}} Each block applies multi-head self-attention followed by feed-forward processing:

\begin{align}
\text{MultiHead}(\mathbf{Q}, \mathbf{K}, \mathbf{V}) &= \text{Concat}(\text{head}_1, \dots, \text{head}_h)\mathbf{W}_o \\
\text{Attention}(\mathbf{Q}, \mathbf{K}, \mathbf{V}) &= \text{Softmax}\left(\frac{\mathbf{QK}^T}{\sqrt{d_k}}\right)\mathbf{V}
\end{align}
where $\mathbf{Q}_i = \mathbf{Z}_i \mathbf{W}_Q$, $\mathbf{K}_i = \mathbf{Z}_i \mathbf{W}_K$, $\mathbf{V}_i = \mathbf{Z}_i \mathbf{W}_V$, and $d_k$ is the attention key dimension. The output of one encoder block is:

\begin{align}
\mathbf{Z}' &= \text{LayerNorm}(\mathbf{Z} + \text{Dropout}(\text{Attention}(\mathbf{Z}))) \\
\mathbf{Z}'' &= \text{LayerNorm}(\mathbf{Z}' + \text{Dropout}(\mathbf{F}(\mathbf{Z}')))
\end{align}
where $\mathbf{F}$ is a feedforward network:

\begin{equation}
\mathbf{F}(\mathbf{Z}_i) = \text{ReLU}(\mathbf{Z}_i \mathbf{W}_1 + b_1)\mathbf{W}_2 + b_2
\end{equation}

To improve robustness against domain-specific spectral distortions, we introduce uncertainty-aware attention scaling. Let $\mathcal{U}_i$ be the entropy-based uncertainty for the $i$-th token. The calibrated attention is given by:

\begin{equation}
\mathbf{A}_{\text{cal}} = \text{Softmax}\left(\frac{\mathbf{QK}^T}{\sqrt{d_k}}\right) \cdot (1 + \lambda \cdot \mathcal{U})
\end{equation}
where $\lambda$ is a hyperparameter controlling uncertainty influence. This guides the model to focus more on spatial-spectral tokens with higher ambiguity, leading to better adaptation. To inject global context, a learnable token $\mathbf{Q}_c \in \mathbb{R}^d$ attends to the encoded patches:

\begin{equation}
\mathbf{A}_{\text{cross}} = \text{Softmax}\left(\frac{\mathbf{Q}_c \mathbf{K}_c^T}{\sqrt{d}}\right) \mathbf{V}_c
\end{equation}

\paragraph{\textbf{Classification Head}} The aggregated features pass through a two-layer MLP with softmax:

\begin{align}
\mathbf{O} &= \text{ReLU}(\mathbf{Z}'' \mathbf{W}_3 + b_3) \\
\mathbf{O}_{\text{final}} &= \text{Softmax}(\mathbf{O} \mathbf{W}_4 + b_4)
\end{align}
where $\mathbf{O}_{\text{final}} \in \mathbb{R}^C$ and $C$ is the number of classes.

%%%%%%%%%%%%%%%%%%%%%%%%%%%%%%%
\subsection{Active Learning}

\paragraph{\textbf{Hybrid Query Strategy}} We define $X_{\text{pool}}$ as the set of samples and $\text{query\_size}$ as the number of samples to be queried in each AL round. Our strategy combines:

\begin{itemize}
    \item \textbf{Uncertainty Sampling:}
        \begin{equation}
            \mathcal{U}(\mathbf{x}_i) = -\max(p(y|\mathbf{x}_i))
        \end{equation}
    \item \textbf{Diversity Sampling:}
        \begin{equation}
            \mathcal{D}(\mathbf{x}_i) = \argmax_{S \subseteq X_{\text{pool}}, |S| = \text{query\_size}} \text{Diversity}(S)
        \end{equation}
        where
        \begin{equation}
            \text{Diversity}(S) = \frac{1}{m(m-1)} \sum_{j=1}^{m} \sum_{\substack{k=1\\k \ne j}}^{m} d_{jk}
        \end{equation}
        and $d_{jk}$ is the Euclidean distance between pixel spectra in local neighborhoods of size $n_{\text{neighborhood}}^2$.
\end{itemize}

\paragraph{\textbf{Query Index Selection}} For each candidate $\mathbf{x}_{(h,w)}$, we compute:

\begin{equation}
    \text{diversity\_metric}(h,w) = \frac{1}{m(m-1)} \sum_{j \ne k} d_{jk}
\end{equation}

The top-$\text{query\_size}$ entries with the highest diversity are selected:
\begin{equation}
    \text{query\_indices} = \left\{ \pi(k) \mid k \in [HW - \text{query\_size} + 1, HW] \right\}
\end{equation}
where
\begin{equation}
    \pi = \text{argsort}(d)
\end{equation}

%%%%%%%%%%%%%%%%%%%%%%%%%%%%%%%
\subsection{Dynamic Layer Freezing}

To facilitate efficient domain adaptation, we introduce a dynamic layer freezing policy. During transfer to a new target domain, SST layers are selectively frozen or unfrozen based on sensitivity to domain shift. We estimate shift magnitude using the maximum mean discrepancy (MMD) of intermediate features. Layers with low variance across domains are frozen, reducing computational cost while retaining useful knowledge.

%%%%%%%%%%%%%%%%%%%%%%%%%%%%%%%
\subsection{Model Training and Evaluation}

The model is trained using categorical cross-entropy:

\begin{equation}
\mathcal{L} = -\sum_{i=1}^{N} \sum_{c=1}^{C} y_{ic} \log(\hat{y}_{ic})
\end{equation}

Evaluation metrics include OA, AA, and the kappa coefficient:

\begin{equation}
\kappa = \frac{p_o - p_e}{1 - p_e}
\end{equation}
where $p_o$ is the observed agreement and $p_e$ is expected agreement by chance.

%%%%%%%%%%%%%%%%%%%%%%%%%%%%%%%
\section{Experimental Settings and Datasets}

The experimental configuration for SST-ATL is designed to ensure clarity, reproducibility, and robust performance evaluation. Following the preprocessing of the HSI cubes and corresponding ground truths, patch-based datasets are generated. Each HSI is divided into overlapping spatial windows of size $8 \times 8$ (i.e., input patch size), which serve as local contextual inputs for the SST backbone. The data are partitioned into training, pool, and test sets with a split ratio of 1\%, 49\%, and 50\%, respectively. Specifically, 1\% of the samples are randomly selected for initial training, 49\% are allocated to the pool for AL queries, and 50\% are reserved exclusively for testing. For diversity-based query selection, a neighborhood size of $n_{\text{neighborhood}} = 3$ is used, resulting in $9$ spectral vectors per neighborhood to compute diversity scores, and the query percentage is set to 0.02\%. 

Model training is conducted under a set of carefully tuned hyperparameters. The learning rate is initialized at 0.001 with a decay factor of $1 \times 10^{-6}$, and the Adam optimizer is employed to facilitate convergence. Training proceeds for 50 epochs with a batch size of 56. The backbone architecture comprises four transformer layers, each featuring eight attention heads. The model dimensionality ($d_{\text{model}}$) is set to 54, and the feedforward network dimensionality is configured as $4 \times 64 = 256$. A dropout rate of 0.1 is applied to mitigate overfitting, while layer normalization is stabilized with an epsilon value of $1 \times 10^{-6}$.

The experimental evaluation in this study was conducted using six widely recognized hyperspectral datasets: Qingyun (QUH), WHU-Hi-HongHu (HH), WHU-Hi-HanChuan (HC), Salinas (SA), University of Houston (UH), and Pavia University (PU). Each dataset encompasses a variety of land cover classes with distinct characteristics, enabling a comprehensive assessment across diverse scenes. For HH and HC, a rich set of agricultural and urban classes was included, while SA predominantly featured vegetation-related categories. UH captured a complex urban environment, and PU primarily focused on urban materials and structures. The QUH dataset significantly expanded the scale and diversity of the evaluation, offering a substantial number of samples across natural and man-made categories. 

%%%%%%%%%%%%%%%%%%%%%%%%%%%%%%%
\section{ATL-SST Effects on the same dataset}

\begin{figure}[!hbt]
    \centering
    \begin{subfigure}{0.24\textwidth}
	\includegraphics[width=0.99\textwidth]{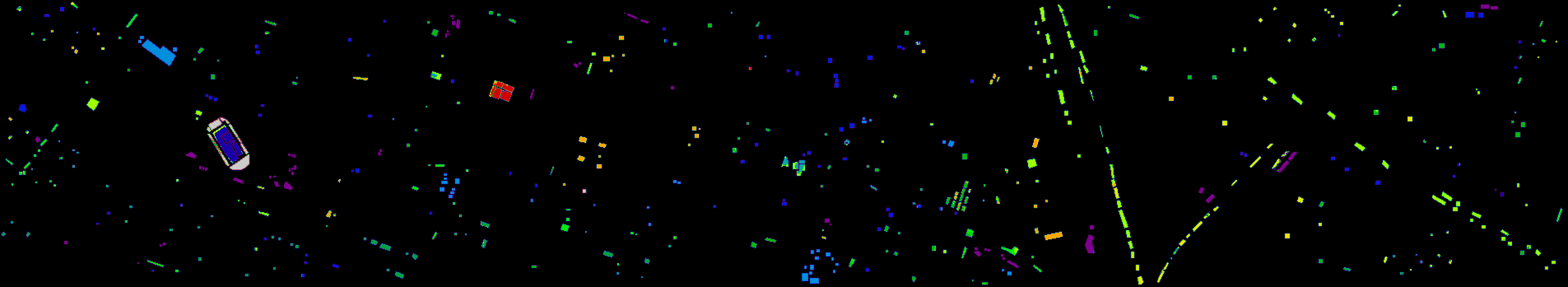}
	\caption*{75 training samples} 
    \end{subfigure}
    \begin{subfigure}{0.24\textwidth}
	\includegraphics[width=0.99\textwidth]{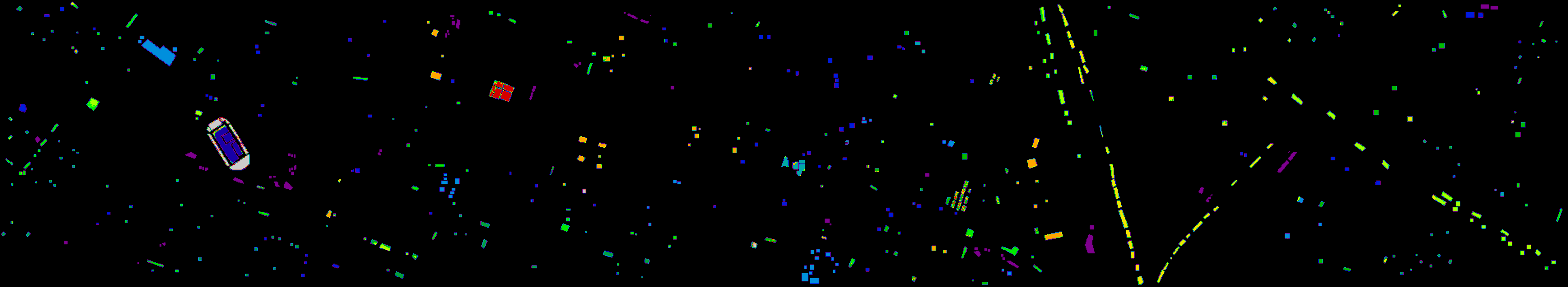}
	\caption*{223 training samples}
    \end{subfigure}
    \begin{subfigure}{0.24\textwidth}
	\includegraphics[width=0.99\textwidth]{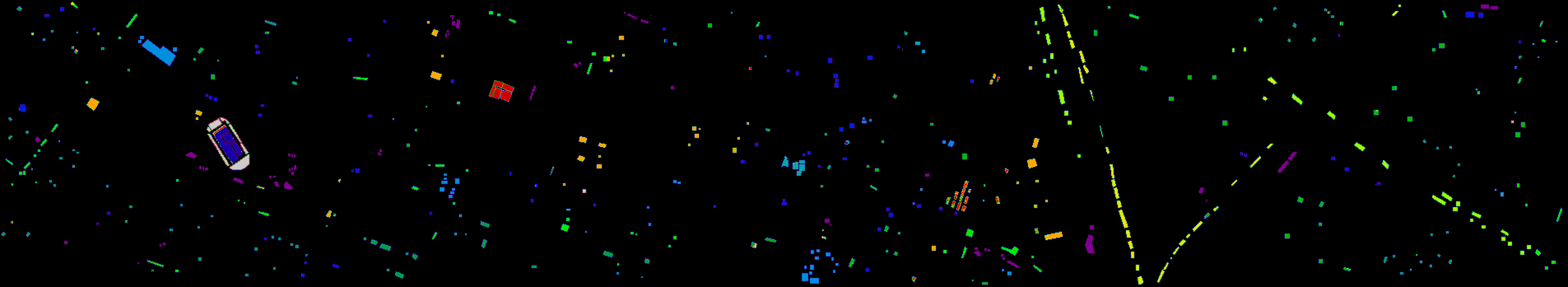}
	\caption*{368 training samples}
    \end{subfigure} 
    \begin{subfigure}{0.24\textwidth}
	\includegraphics[width=0.99\textwidth]{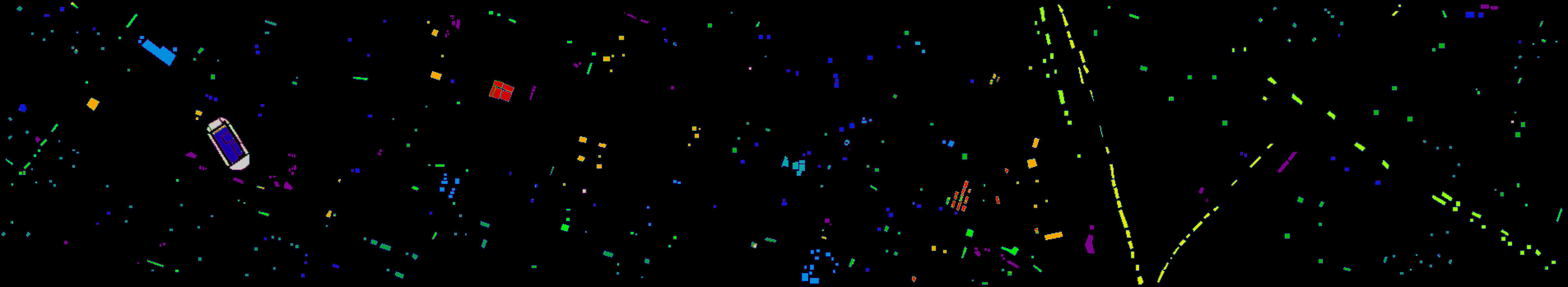}
	\caption*{510 training samples}
    \end{subfigure}
    \begin{subfigure}{0.24\textwidth}
	\includegraphics[width=0.99\textwidth]{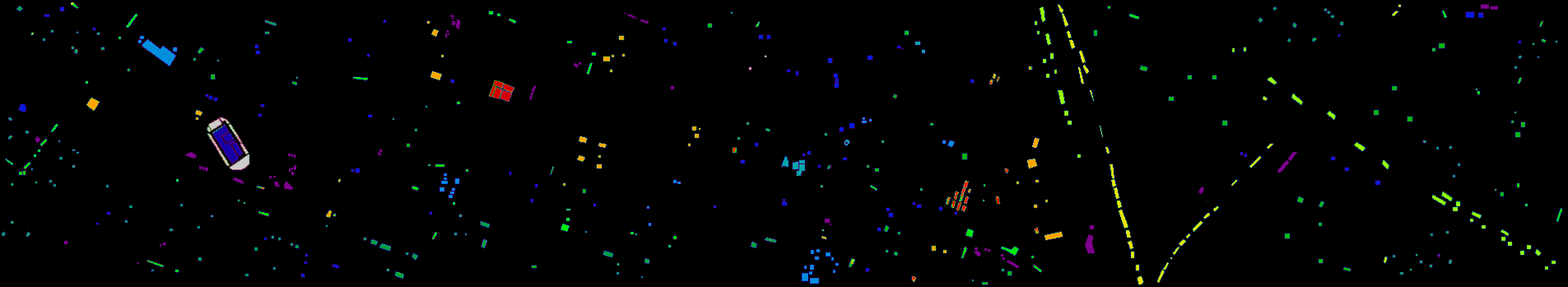}
	\caption*{650 training samples}
    \end{subfigure}
    \begin{subfigure}{0.24\textwidth}
	\includegraphics[width=0.99\textwidth]{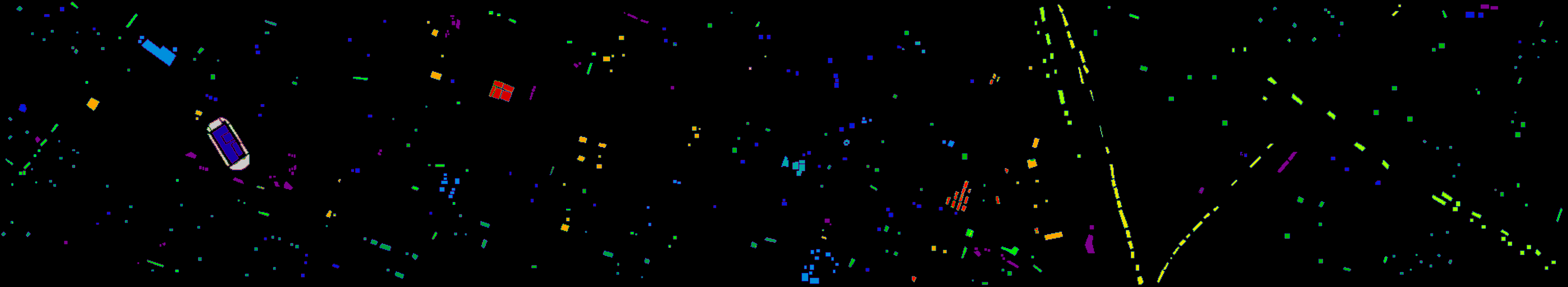}
	\caption*{787 training samples}
    \end{subfigure}
\caption{\textbf{UH dataset}: Ground truth maps corresponding to varying numbers of training samples.}
\label{UHGT}
\end{figure}

The results presented in Table \ref{UHT} highlight the consistent improvement in classification accuracy as the number of training samples increases. Table \ref{UHT} also provides computational metrics, showcasing the model’s efficiency. Training times remain stable at around 44s across iterations, despite the increase in sample size, indicating scalability. The testing time is consistently around 3.4s, demonstrating the method's suitability for real-time applications. The reported FLOPs (110,336) and parameter count (836,559) reflect the model’s balance between computational efficiency and capacity, making it practical. Qualitative improvements are visually evident in Figure \ref{UHGT}, which presents ground truth maps for varying training sample sizes. With fewer training samples (e.g., 75 and 223), the maps exhibit substantial noise and misclassification. As the training sample size increases, the maps become progressively refined, reflecting improved spatial and spectral consistency.

%%%%%%%%%%%%%%%%%%%%%%%%%%%%%%%
\begin{table}[!hbt]
    \centering
    \caption{\textbf{UH Dataset}: Accuracy improvements as the sample count increases per iteration. Additionally, average training and testing times (in seconds) and the total FLOPs and parameter count provide a comprehensive overview of computational requirements.}
    \begin{tabular}{c|cccccc} \hline 
        \multirow{2}{*}{\textbf{Class}} & \multicolumn{6}{c}{\textbf{Number of Training Samples}} \\ \cline{2-7}
        & 75 & 223 & 368 & 510 & 650 & 787 \\ \hline 

        $\kappa$ & 71.27 & 84.98 & 91.00 & 94.96 & 96.17 & 97.38 \\ 
        OA & 73.49 & 86.13 & 91.68 & 95.34 & 96.46 & 97.57 \\ 
        AA & 71.02 & 84.32 & 90.35 & 94.27 & 95.57 & 96.97 \\ \hline 
            
        Train (s) & 50.52 & 44.63 & 44.98 & 44.34 & 44.96 & 47.39 \\
        Test (s) & 4.43 & 3.45 & 3.45 & 3.35 & 3.44 & 3.44 \\ \hline 
        Flops & \multicolumn{6}{c}{\textbf{110336}} \\
        Param & \multicolumn{6}{c}{\textbf{836559}} \\ \hline 
    \end{tabular}
    \label{UHT}
\end{table}
%%%%%%%%%%%%%%%%%%%%%%%%%%%%%%%
%%%%%%%%%%%%%%%%%%%%%%%%%%%%%%%

\begin{figure}[!hbt]
    \centering
    \begin{subfigure}{0.07\textwidth}
	\includegraphics[width=0.99\textwidth]{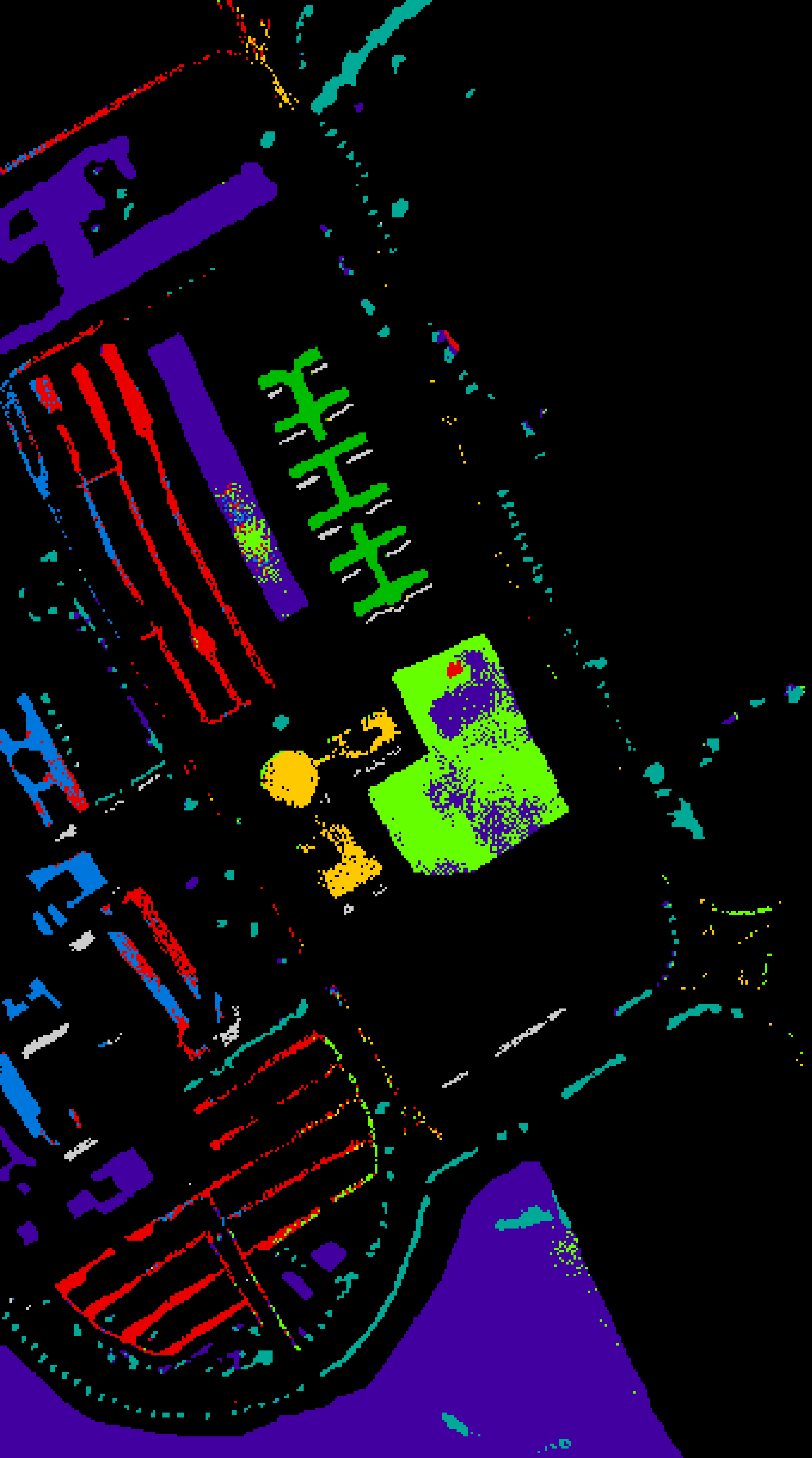}
	\caption*{213} 
    \end{subfigure}
    \begin{subfigure}{0.07\textwidth}
	\includegraphics[width=0.99\textwidth]{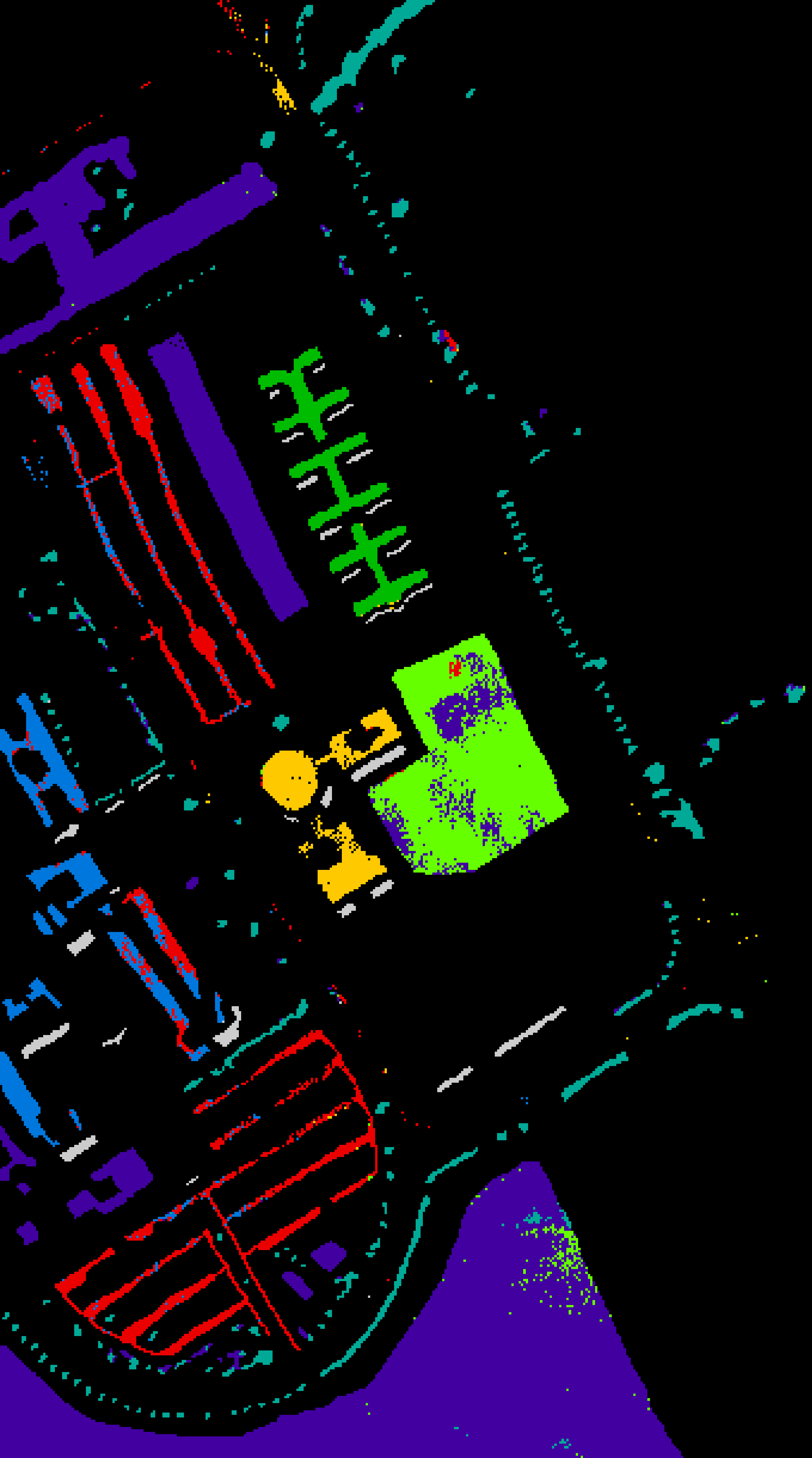}
	\caption*{636}
    \end{subfigure}
    \begin{subfigure}{0.07\textwidth}
	\includegraphics[width=0.99\textwidth]{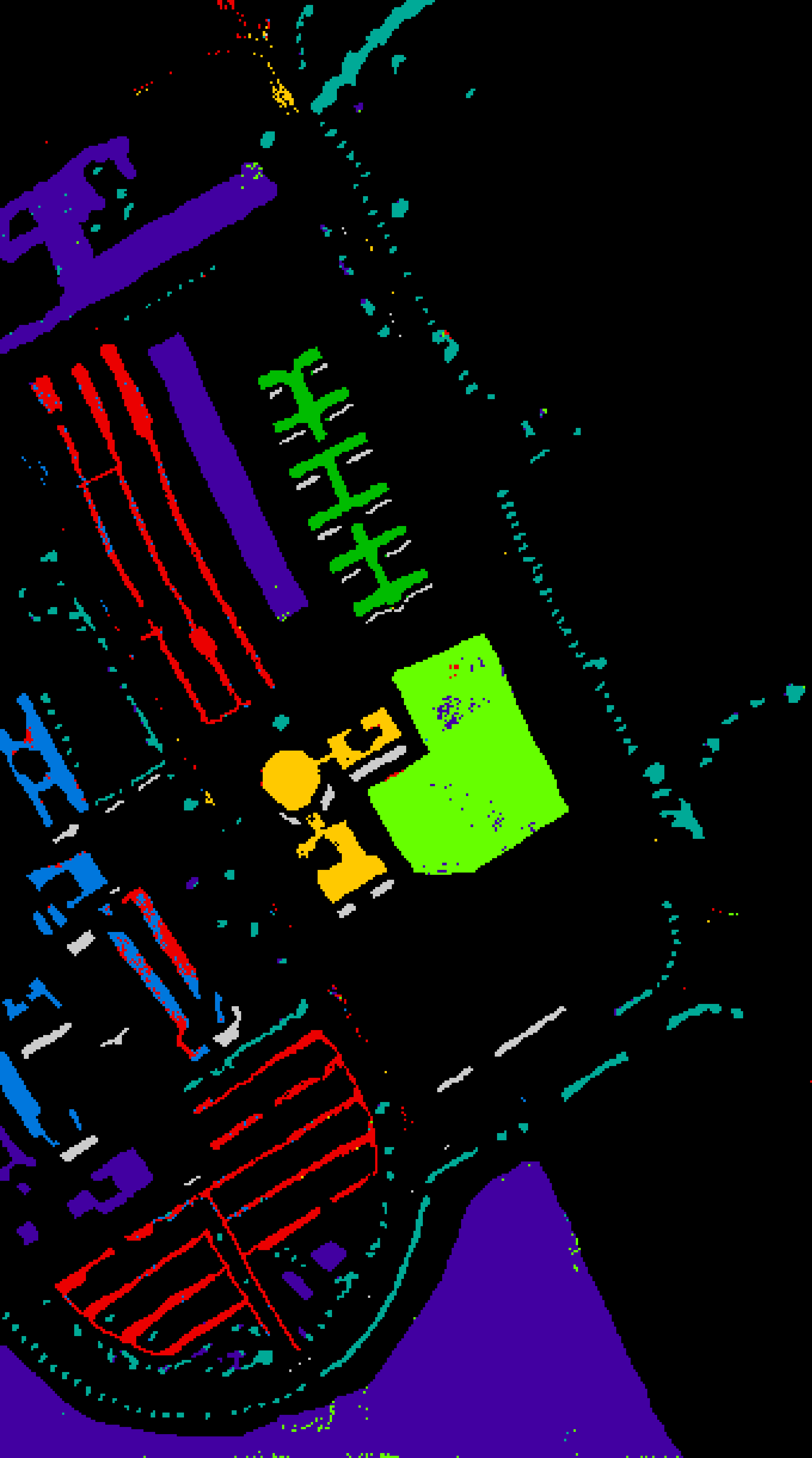}
	\caption*{1051}
    \end{subfigure} 
    \begin{subfigure}{0.07\textwidth}
	\includegraphics[width=0.99\textwidth]{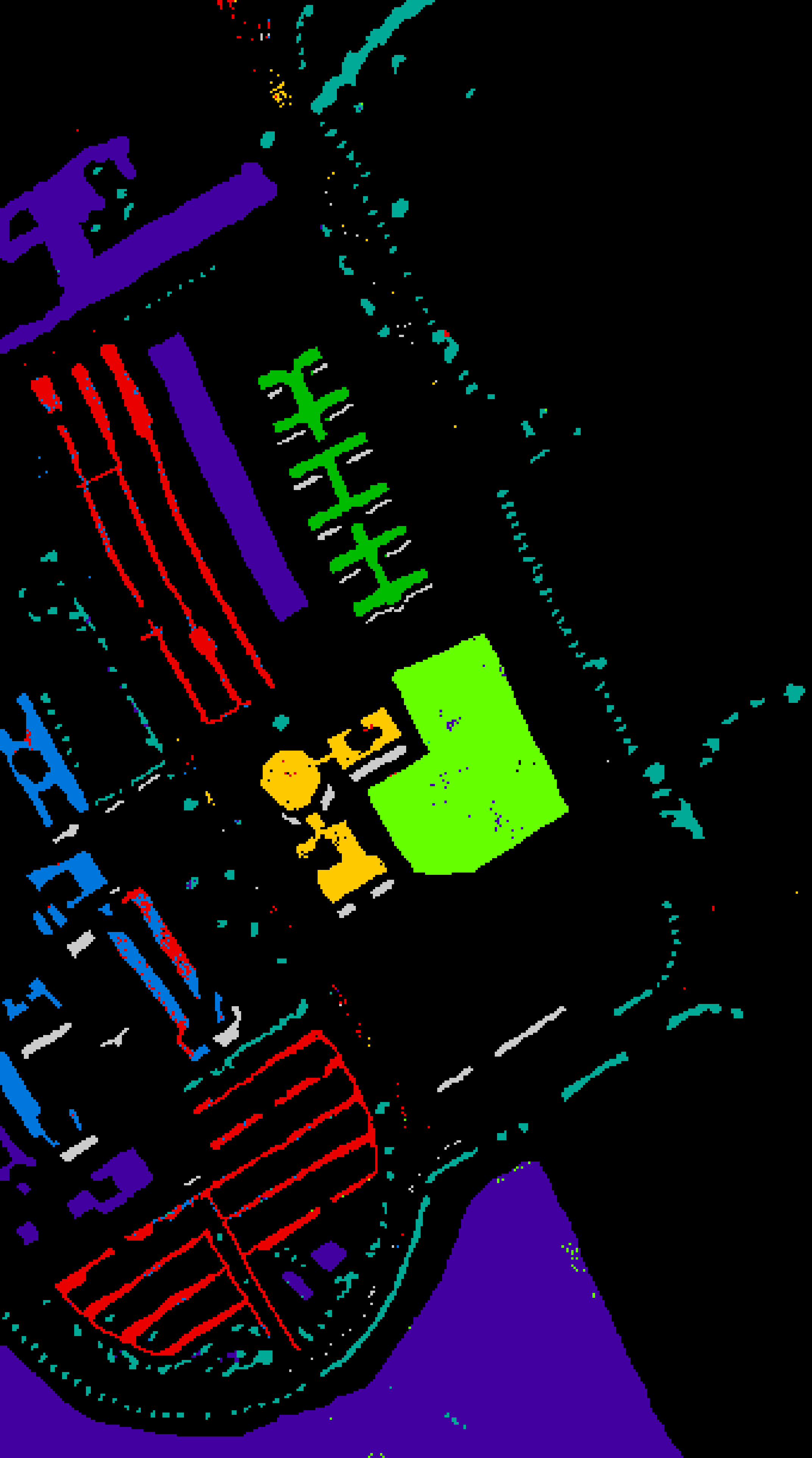}
	\caption*{1457}
    \end{subfigure}
    \begin{subfigure}{0.07\textwidth}
	\includegraphics[width=0.99\textwidth]{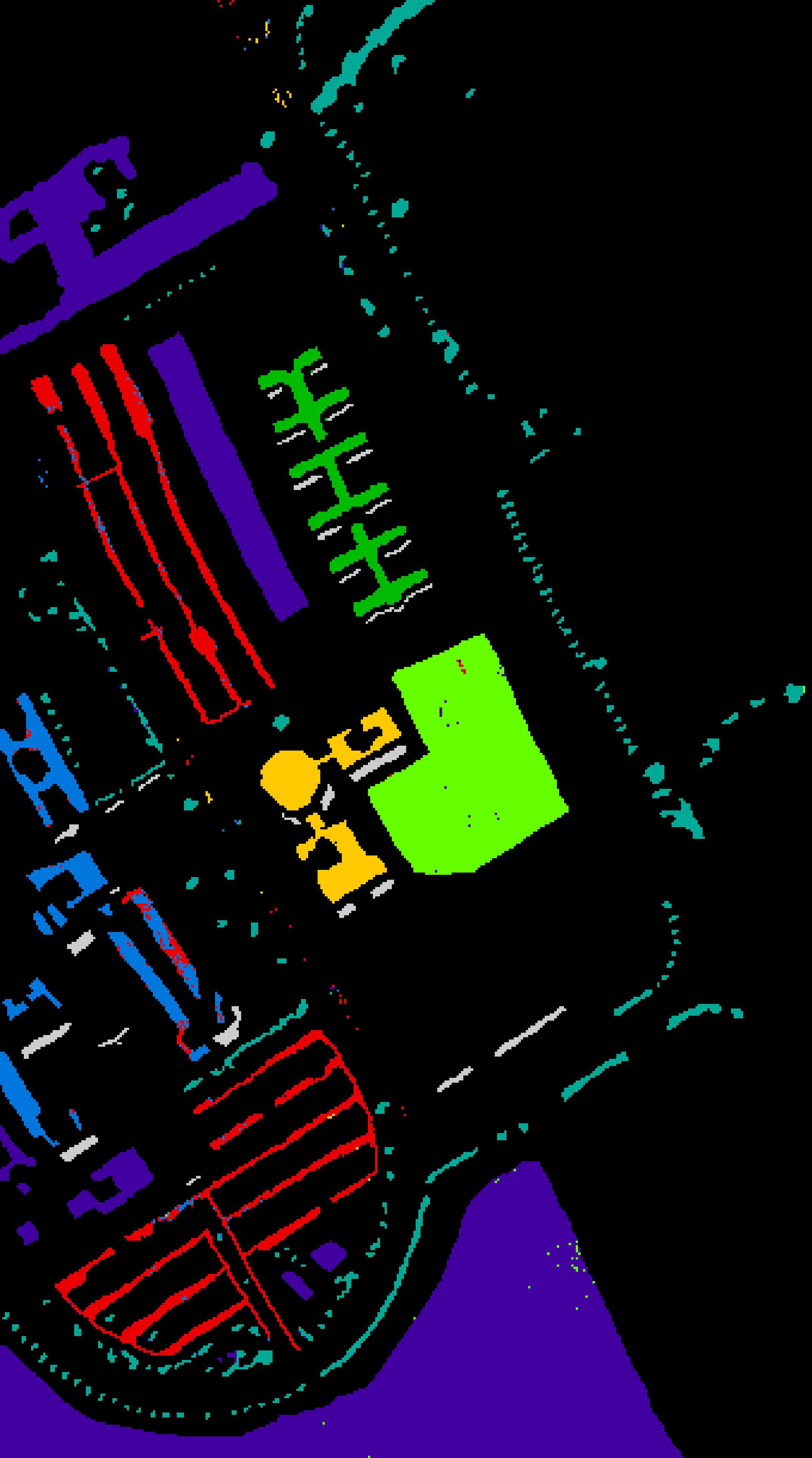}
	\caption*{1855}
    \end{subfigure}
    \begin{subfigure}{0.07\textwidth}
	\includegraphics[width=0.99\textwidth]{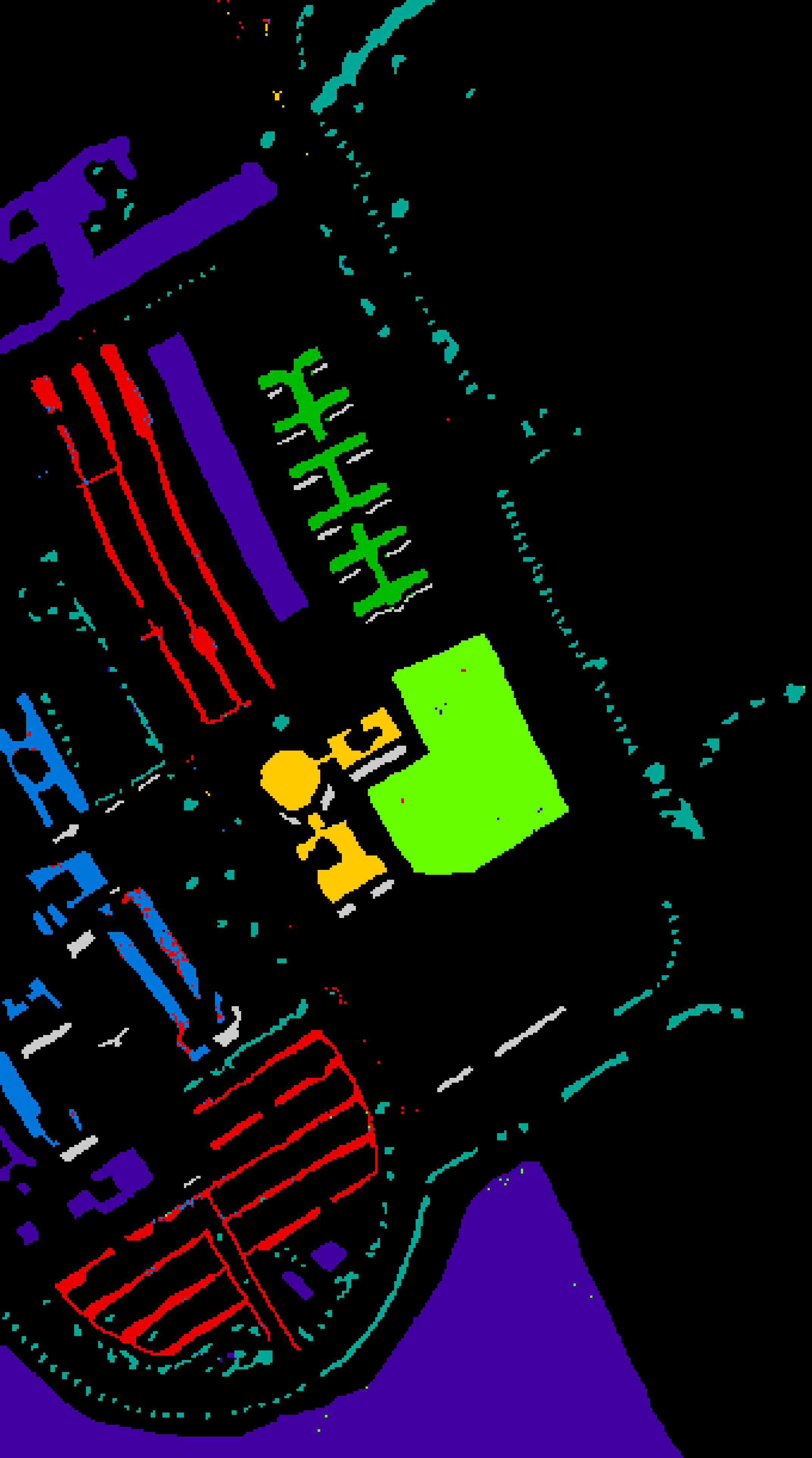}
	\caption*{2245}
    \end{subfigure}
\caption{\textbf{PU dataset}: Ground truth maps corresponding to varying numbers of training samples.}
\label{PUGT}
\end{figure}

The results presented in Table \ref{PUT} and \ref{PUGT} collectively highlight the model's performance progression on the PU dataset as training samples increase. Training and testing times remain relatively stable across sample sizes, indicating computational efficiency; a total of 108,800 FLOPs and 835,017 parameters suggests a well-optimized model that balances accuracy and computational overhead. The accompanying Figure further substantiates these findings by plotting OA, AA, and $\kappa$ scores against sample counts, illustrating a clear upward trend with convergence near 99\% as sample sizes approach 2245. 

%%%%%%%%%%%%%%%%%%%%%%%%%%%%%%%
\begin{table}[!hbt]
    \centering
    \caption{\textbf{PU Dataset}: Accuracy improvements as the sample count increases per iteration.}
    \begin{tabular}{c|cccccc} \hline 
        \multirow{2}{*}{\textbf{Class}} & \multicolumn{6}{c}{\textbf{Number of Training Samples}} \\ \cline{2-7}
        & 213 & 636 & 1051 & 1457 & 1855 & 2245 \\ \hline 

        $\kappa$ & 82.69 & 90.62 & 95.65 & 96.98 & 97.85 & 98.51 \\
        OA & 87.11 & 92.98 & 96.72 & 97.72 & 98.38 & 98.87 \\
        AA & 80.26 & 90.20 & 95.10 & 96.35 & 97.35 & 98.17 \\ \hline 
            
        Train (s) & 118.11 & 117.25 & 115.99 & 119.21 & 119.78 & 124.78 \\ 
        Test (s) & 9.87 & 9.26 & 9.08 & 8.79 & 9.07 & 9.09 \\ \hline 
        Flops & \multicolumn{6}{c}{\textbf{108800}} \\
        Param & \multicolumn{6}{c}{\textbf{835017}} \\ \hline 
    \end{tabular}
    \label{PUT}
\end{table}
%%%%%%%%%%%%%%%%%%%%%%%%%%%%%%%
%%%%%%%%%%%%%%%%%%%%%%%%%%%%%%%

\begin{figure}[!hbt]
    \centering
    \begin{subfigure}{0.07\textwidth}
	\includegraphics[width=0.99\textwidth]{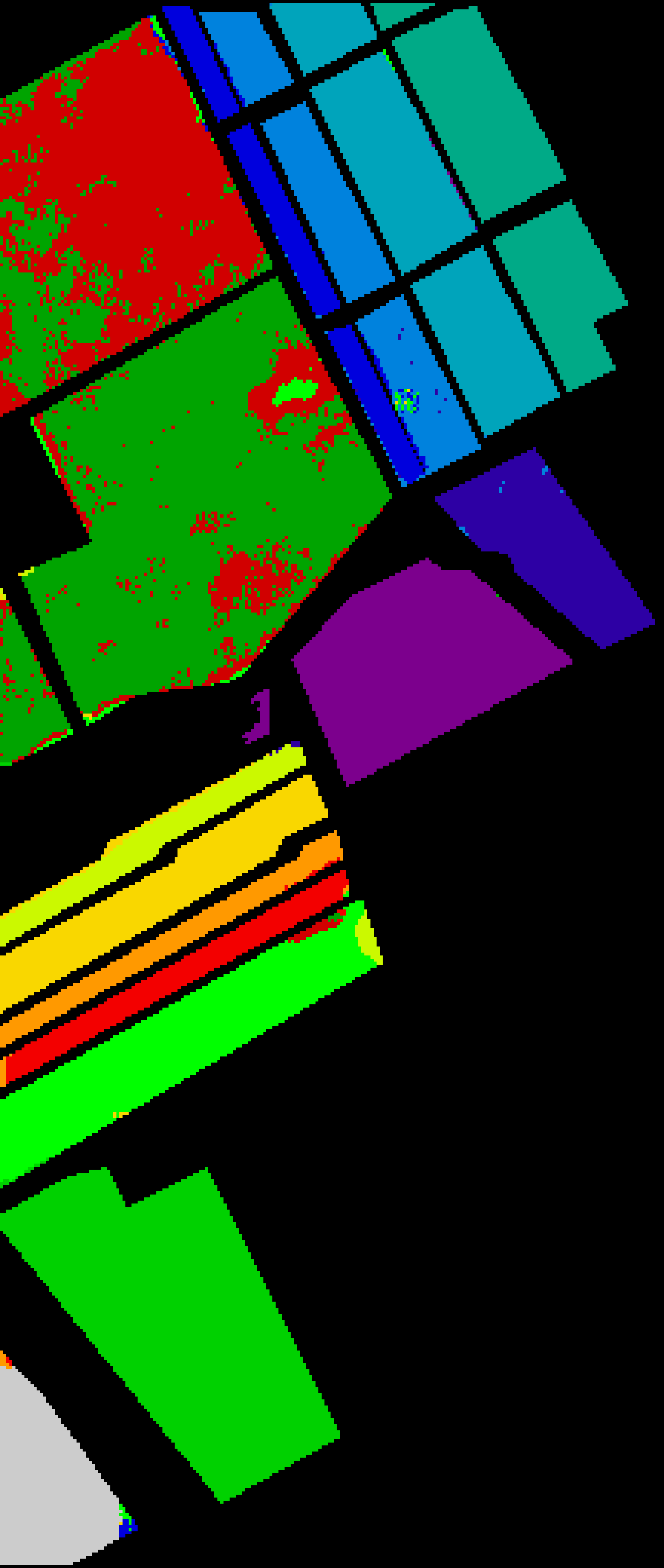}
	\caption*{270} 
    \end{subfigure}
    \begin{subfigure}{0.07\textwidth}
	\includegraphics[width=0.99\textwidth]{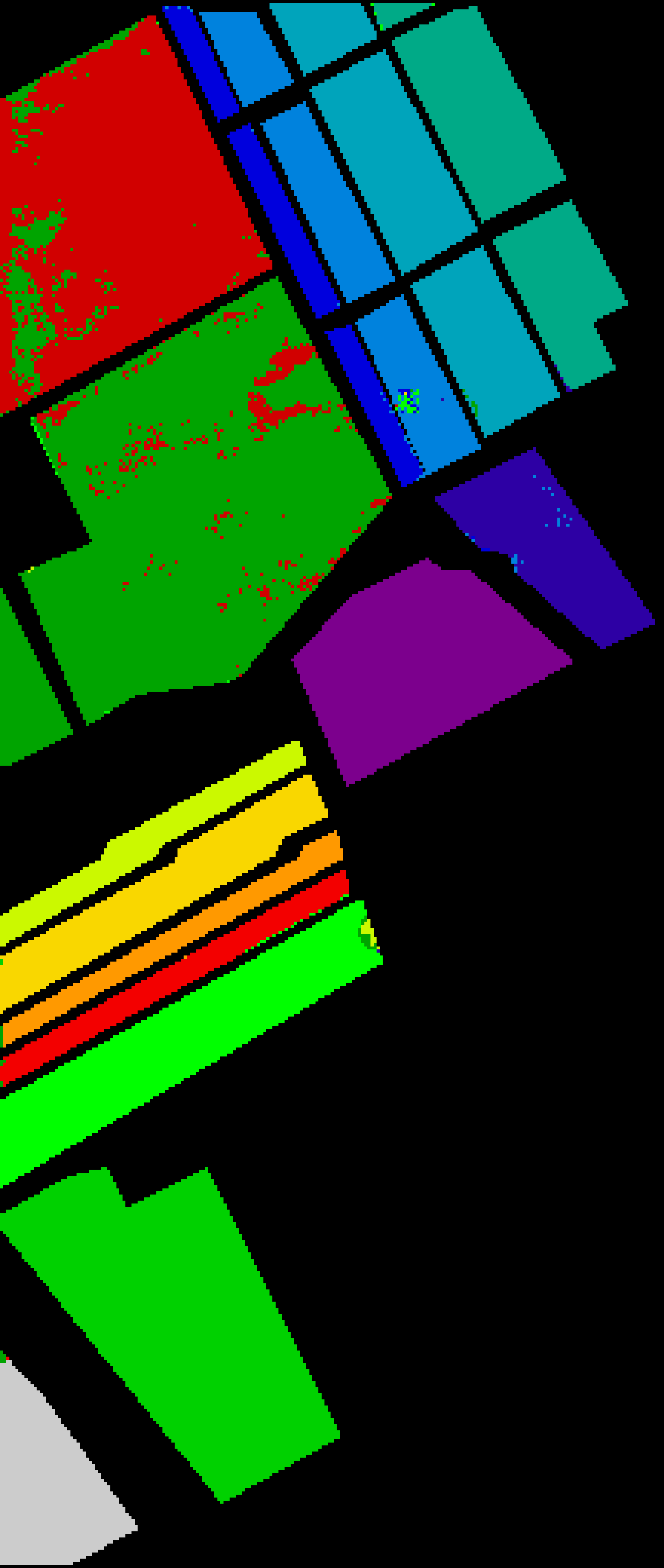}
	\caption*{805}
    \end{subfigure}
    \begin{subfigure}{0.07\textwidth}
	\includegraphics[width=0.99\textwidth]{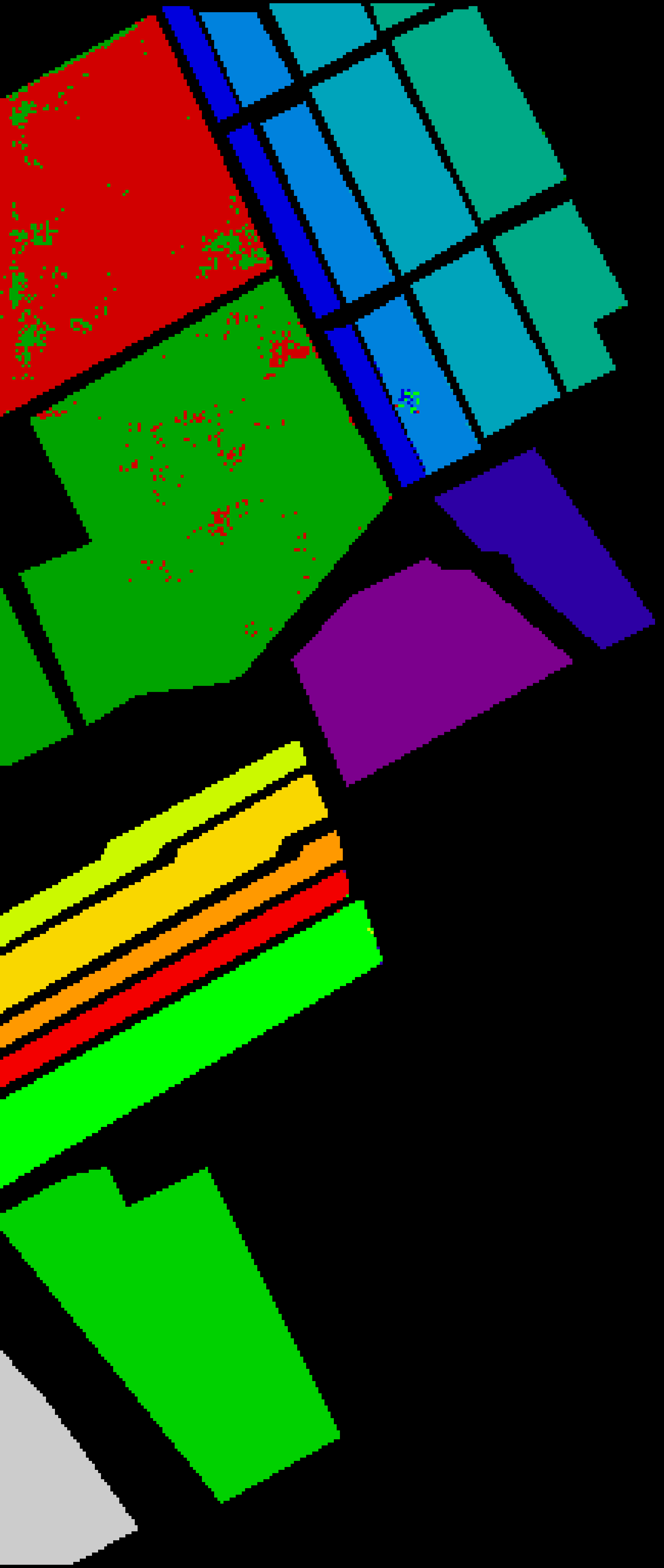}
	\caption*{1330}
    \end{subfigure} 
    \begin{subfigure}{0.07\textwidth}
	\includegraphics[width=0.99\textwidth]{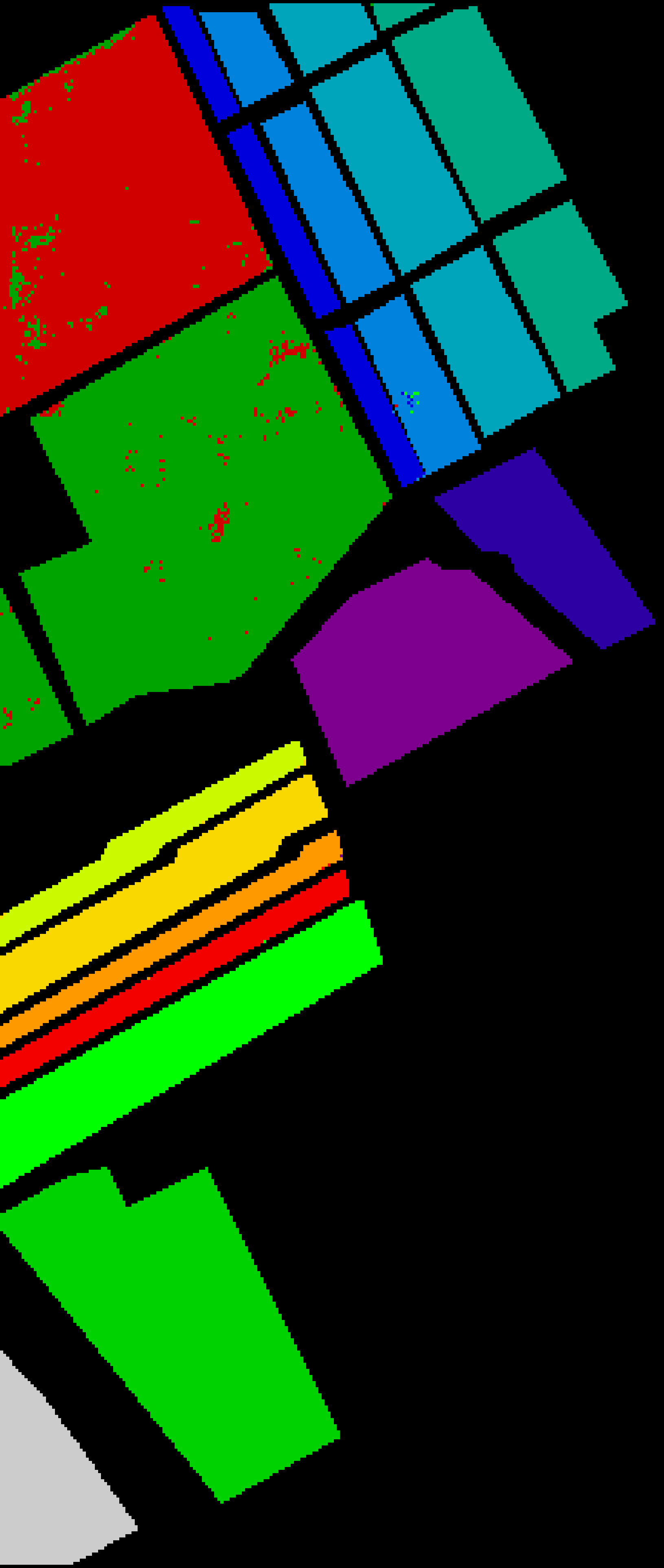}
	\caption*{1844}
    \end{subfigure}
    \begin{subfigure}{0.07\textwidth}
	\includegraphics[width=0.99\textwidth]{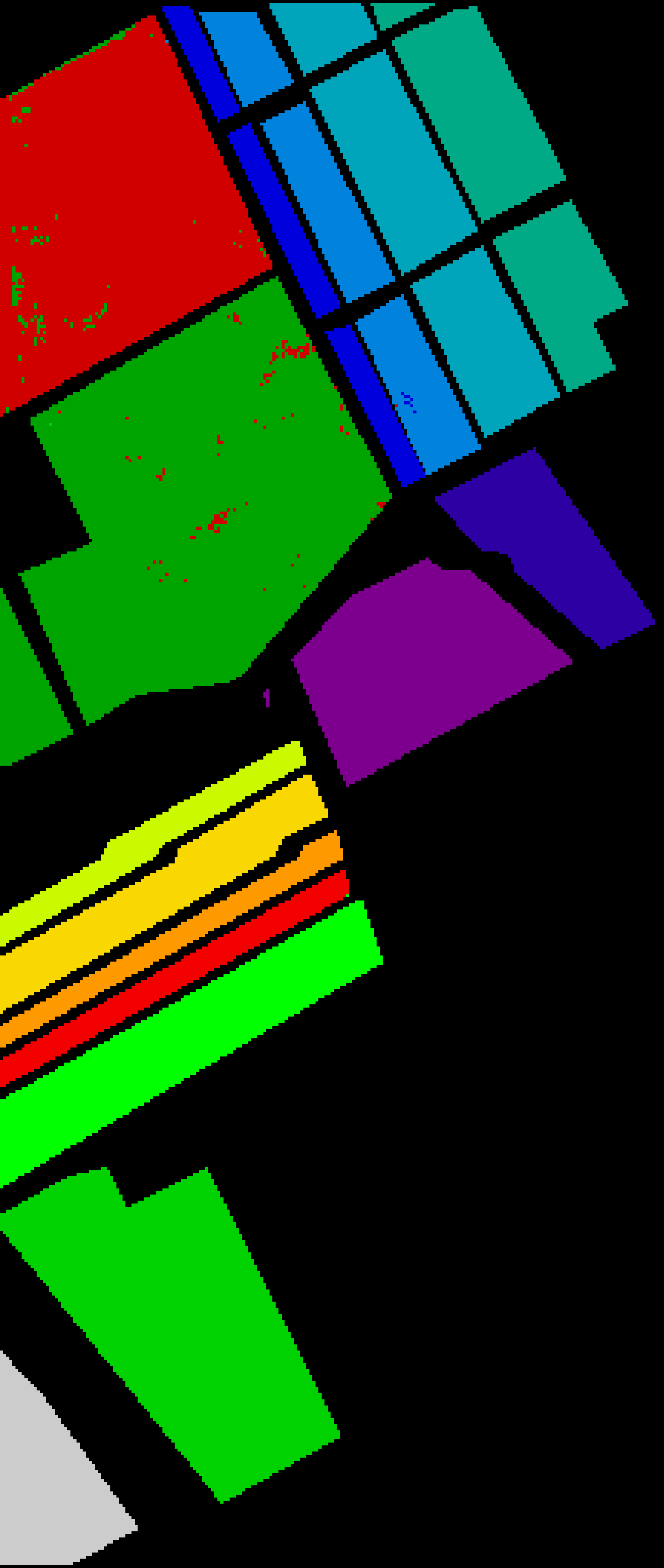}
	\caption*{2348}
    \end{subfigure}
    \begin{subfigure}{0.07\textwidth}
	\includegraphics[width=0.99\textwidth]{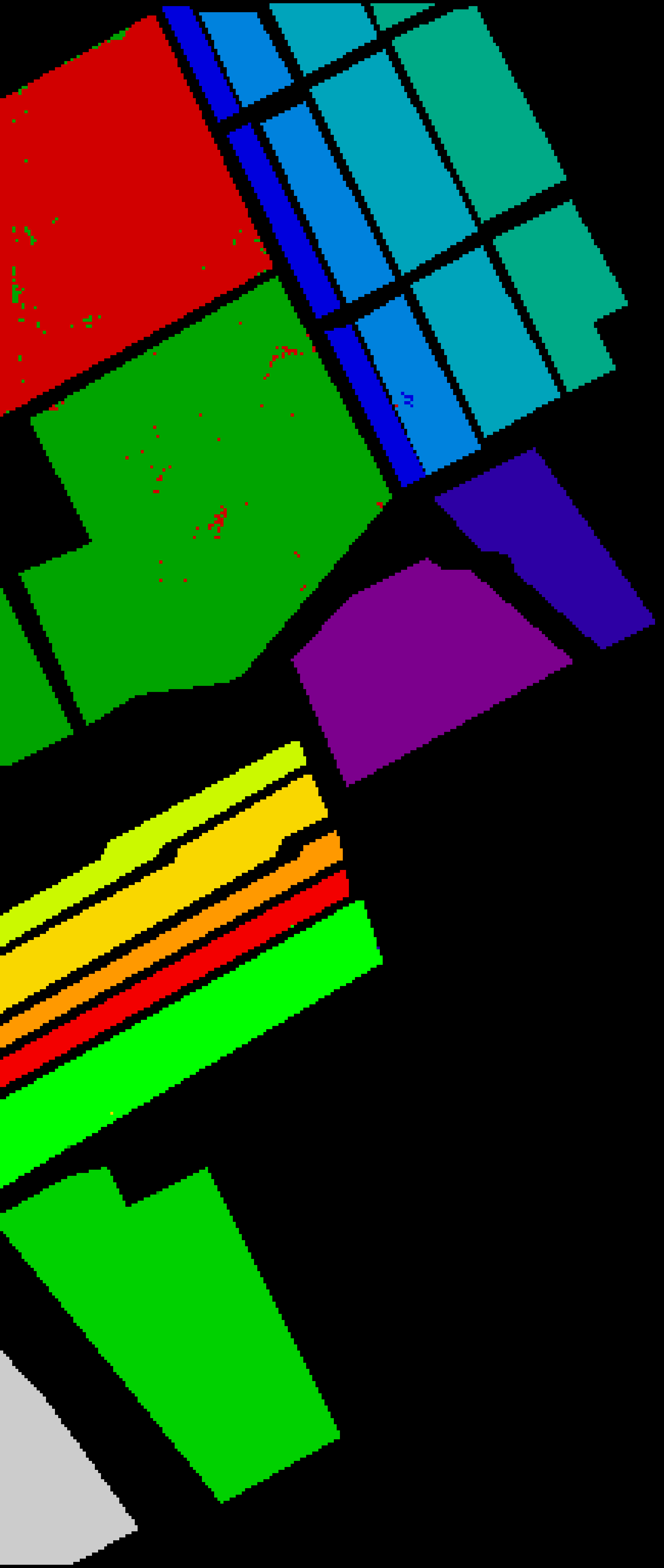}
	\caption*{2842}
    \end{subfigure}
\caption{\textbf{SA dataset}: Ground truth maps corresponding to varying numbers of training samples.}
\label{SAGT}
\end{figure}

Table \ref{SAT} and \ref{SAGT} thoroughly analyze the model’s performance, computational complexity, and the scalability of its classification accuracy on the SA dataset. Although training times gradually increase as sample sizes grow, testing times remain relatively stable, underscoring the model's scalability and efficiency. Furthermore, Table \ref{SAT} provides computational metrics, including FLOPs and parameter counts, which quantify the model’s complexity at approximately 110,592 FLOPs and 836,816 parameters. These values confirm the model’s high computational efficiency, which is an essential consideration for large-scale applications. Finally, Figure \ref{SAGT} provides ground truth maps generated with different training sample sizes, visually showcasing classification performance improvements.

%%%%%%%%%%%%%%%%%%%%%%%%%%%%%%%
\begin{table}[!hbt]
    \centering
    \caption{\textbf{SA Dataset}: Accuracy improvements as the sample count increases per iteration.}
    \begin{tabular}{c|cccccc} \hline 
        \multirow{2}{*}{\textbf{Class}} & \multicolumn{6}{c}{\textbf{Number of Training Samples}} \\ \cline{2-7}
        & 270 & 805 & 1330 & 1844 & 2348 & 2842 \\ \hline 
        
        $\kappa$ & 91.55 & 96.59 & 97.99 & 98.62 & 99.08 & 99.42 \\
        OA & 92.41 & 96.94 & 98.20 & 98.76 & 99.17 & 99.48 \\
        AA & 95.47 & 98.30 & 99.15 & 99.41 & 99.62 & 99.76 \\ \hline 

        Train (s) & 149.00 & 144.89 & 149.44 & 152.20 & 155.20 & 157.85 \\ 
        Test (s) & 12.44 & 11.40 & 11.60 & 11.32 & 10.69 & 11.49 \\ \hline 
        Flops & \multicolumn{6}{c}{\textbf{110592}} \\
        Param & \multicolumn{6}{c}{\textbf{836816}} \\ \hline 
    \end{tabular}
    \label{SAT}
\end{table}
%%%%%%%%%%%%%%%%%%%%%%%%%%%%%%%
%%%%%%%%%%%%%%%%%%%%%%%%%%%%%%%

\begin{figure}[!hbt]
    \centering
    \begin{subfigure}{0.07\textwidth}
	\includegraphics[width=0.99\textwidth]{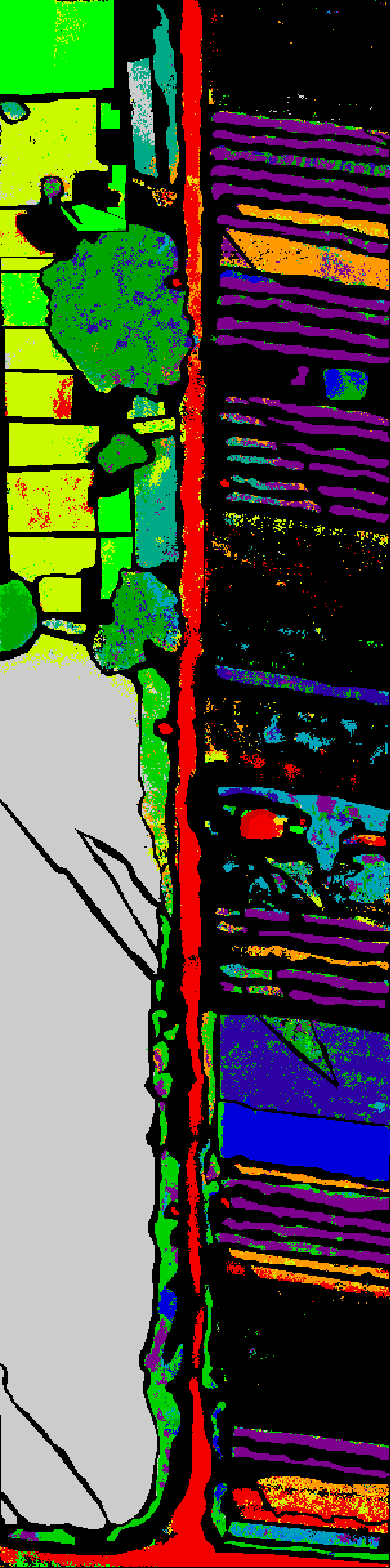}
	\caption*{1287} 
    \end{subfigure}
    \begin{subfigure}{0.07\textwidth}
	\includegraphics[width=0.99\textwidth]{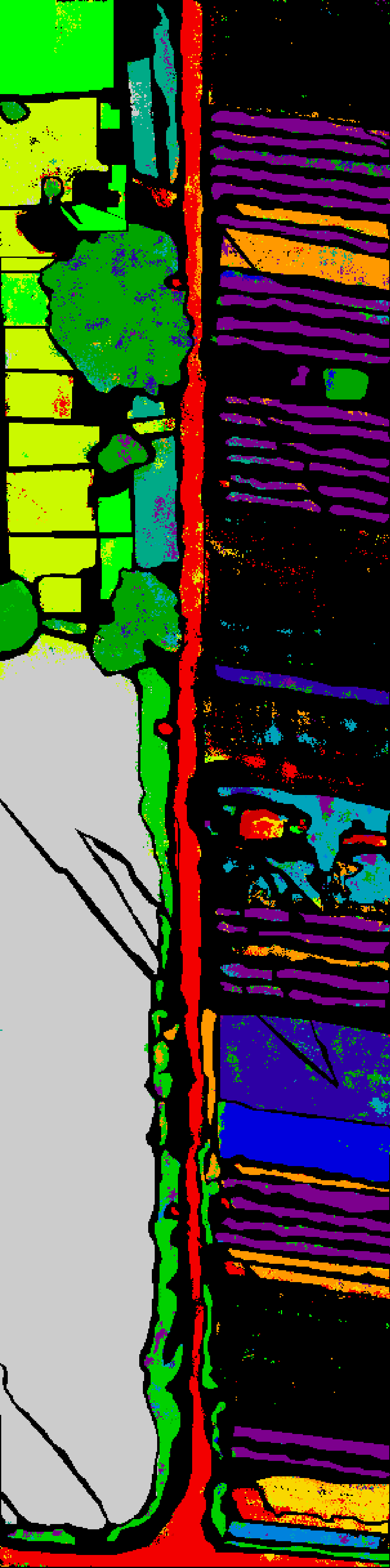}
	\caption*{3836}
    \end{subfigure}
    \begin{subfigure}{0.07\textwidth}
	\includegraphics[width=0.99\textwidth]{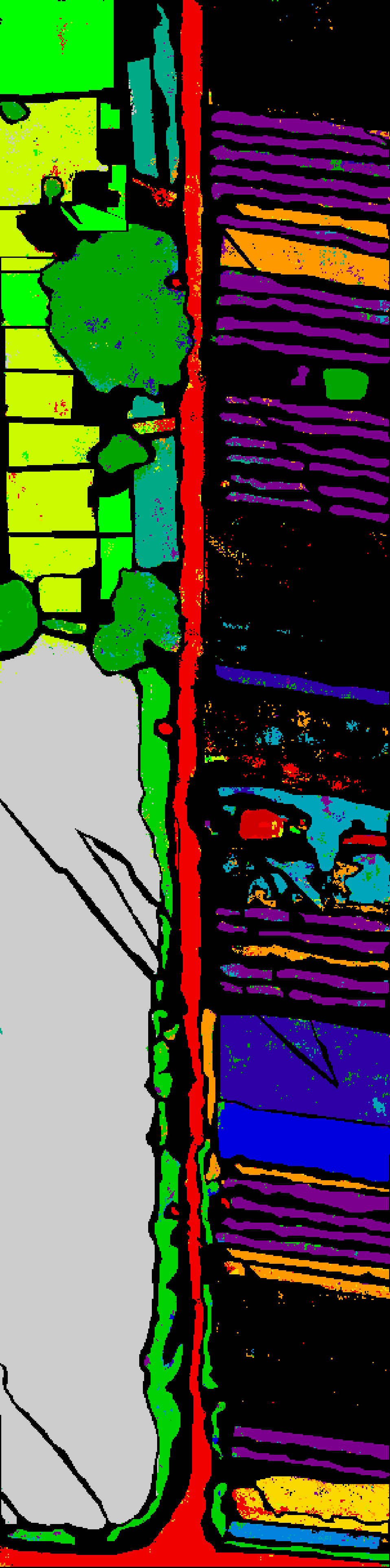}
	\caption*{6334}
    \end{subfigure} 
    \begin{subfigure}{0.07\textwidth}
	\includegraphics[width=0.99\textwidth]{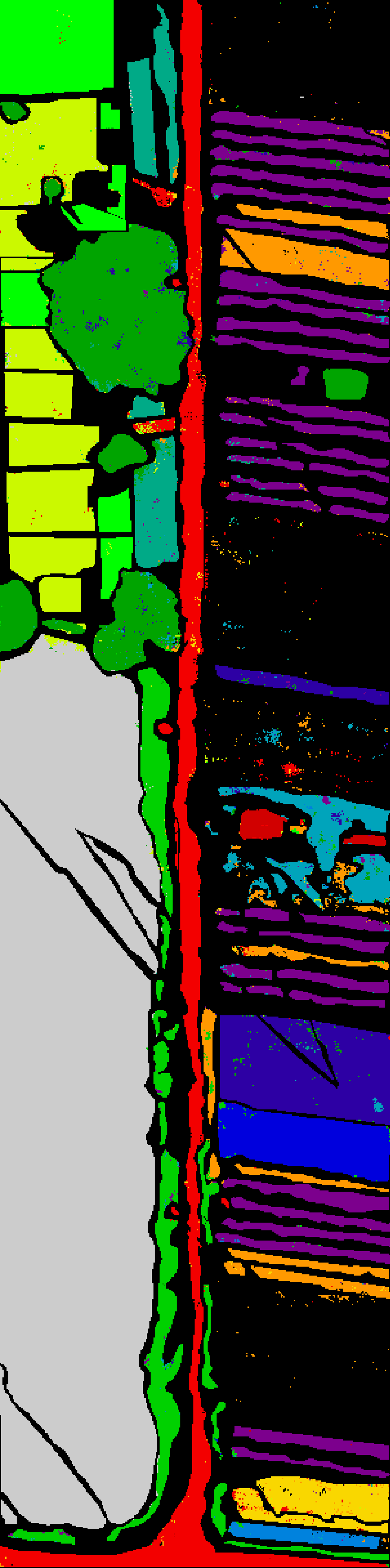}
	\caption*{8782}
    \end{subfigure}
    \begin{subfigure}{0.07\textwidth}
	\includegraphics[width=0.99\textwidth]{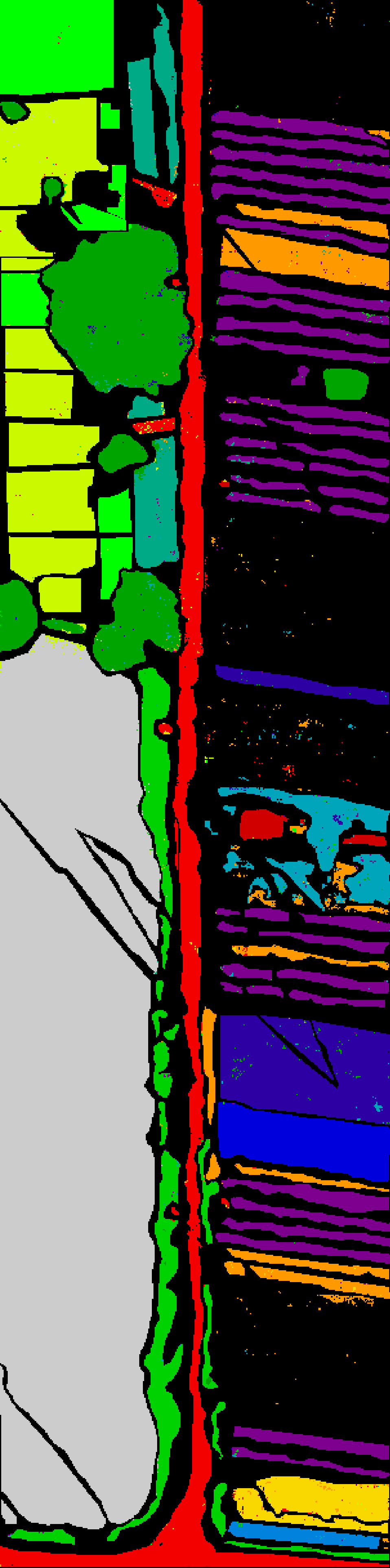}
	\caption*{11181}
    \end{subfigure}
    \begin{subfigure}{0.07\textwidth}
	\includegraphics[width=0.99\textwidth]{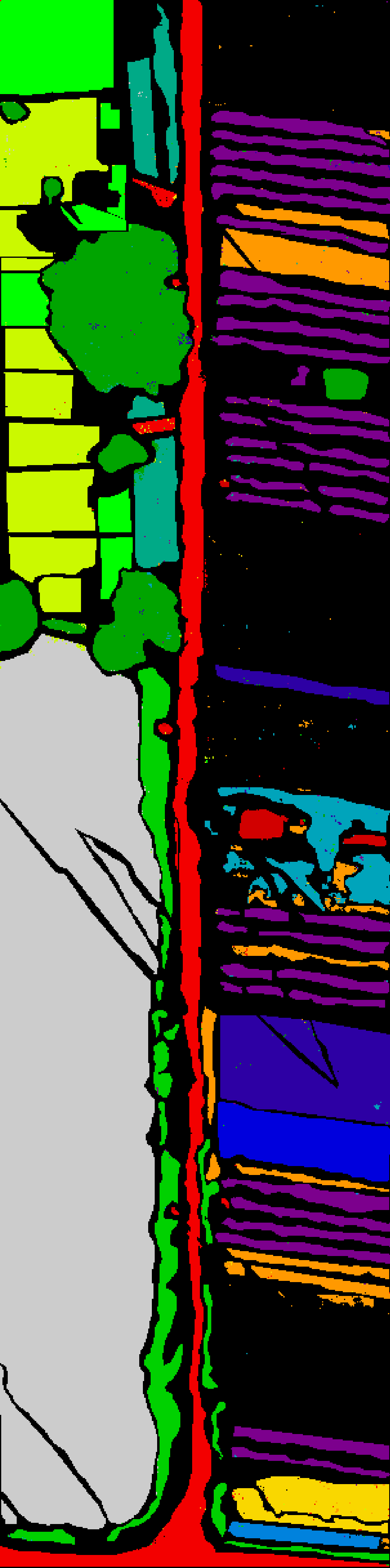}
	\caption*{13532}
    \end{subfigure}
\caption{\textbf{HC dataset}: Ground truth maps corresponding to varying numbers of training samples.}
\label{HCGT}
\end{figure}

Table \ref{HCT} provides a comprehensive overview of the HC dataset's classification performance and computational cost. The computational metrics reveal a gradual increase in training time with more samples, reflecting the added processing requirements, while testing times remain relatively stable. Furthermore, the reported FLOPs and parameter counts provide insight into the model's computational efficiency, balancing accuracy with resource usage. Figure \ref{HCGT} showcases the ground truth maps for each training sample size, visually representing the model's learning progress. With 1,287 samples, the classification appears fragmented, reflecting limited learning. As the sample size increases, the maps exhibit progressively finer details, and class boundaries become more accurate and well-defined. 

%%%%%%%%%%%%%%%%%%%%%%%%%%%%%%%
\begin{table}[!hbt]
    \centering
    \caption{\textbf{HC Dataset}: Accuracy improvements as the sample count increases per iteration.}
    \begin{tabular}{c|cccccc} \hline 
        \multirow{2}{*}{\textbf{Class}} & \multicolumn{6}{c}{\textbf{Number of Training Samples}} \\ \cline{2-7}
        & 1287 & 3836 & 6334 & 8782 & 11181 & 13532 \\ \hline 

        $\kappa$ & 84.71 & 91.39 & 94.78 & 96.89 & 98.12 & 98.83 \\
        OA & 86.97 & 92.65 & 95.54 & 97.34 & 98.39 & 99.00 \\
        AA & 70.46 & 84.04 & 90.67 & 94.33 & 96.36 & 97.60 \\ \hline 

        Train (s) & 523.75 & 544.15 & 547.76 & 598.53 & 618.34 & 648.47 \\ 
        Test (s) & 53.42 & 50.17 & 51.15 & 55.11 & 54.07 & 54.52 \\ \hline 
        Flops & \multicolumn{6}{c}{\textbf{110592}} \\
        Param & \multicolumn{6}{c}{\textbf{836816}} \\ \hline 
    \end{tabular}
    \label{HCT}
\end{table}
%%%%%%%%%%%%%%%%%%%%%%%%%%%%%%%
%%%%%%%%%%%%%%%%%%%%%%%%%%%%%%%

\begin{figure}[!hbt]
    \centering
    \begin{subfigure}{0.07\textwidth}
	\includegraphics[width=0.99\textwidth]{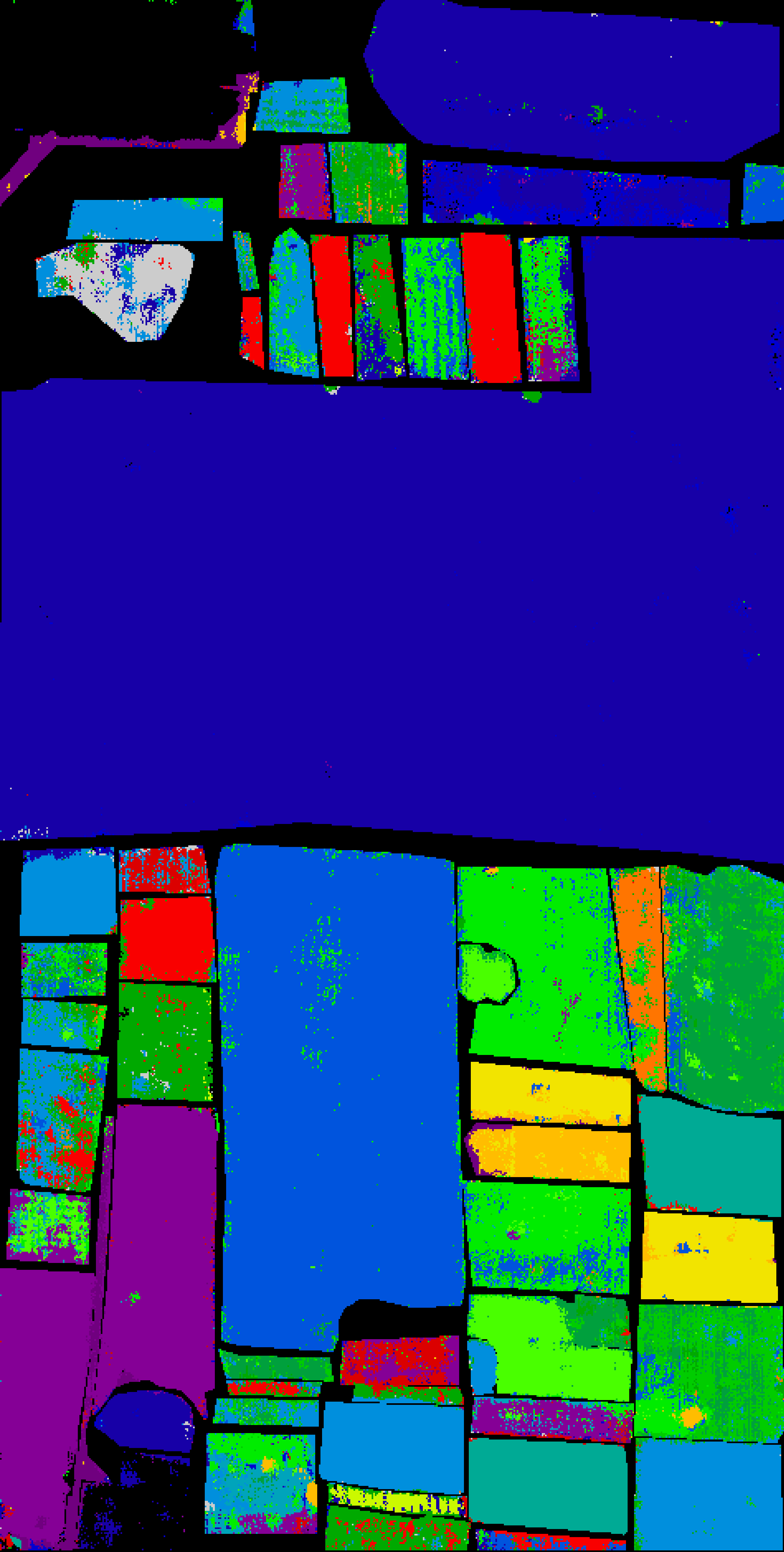}
	\caption*{1933} 
    \end{subfigure}
    \begin{subfigure}{0.07\textwidth}
	\includegraphics[width=0.99\textwidth]{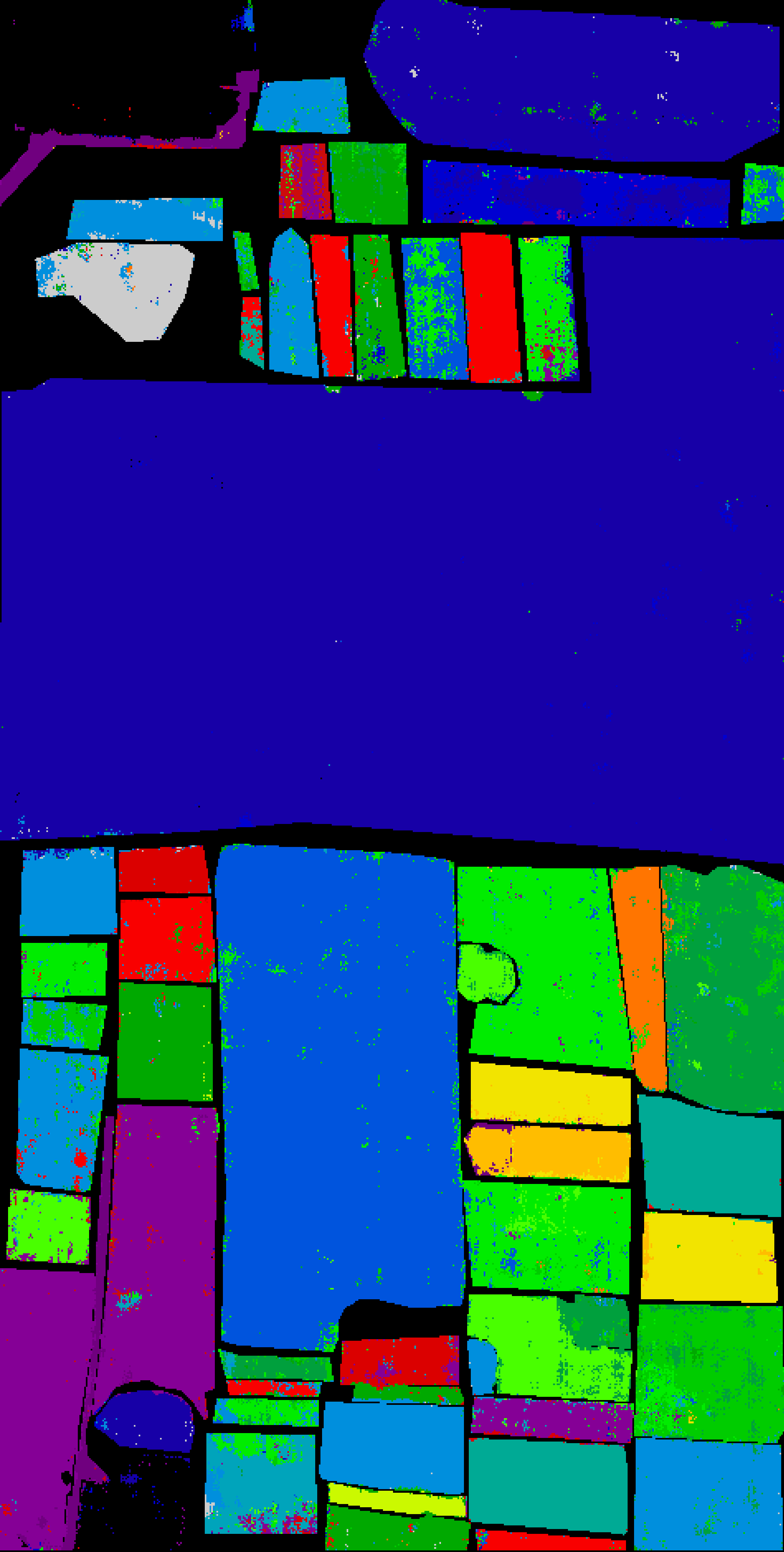}
	\caption*{5761}
    \end{subfigure}
    \begin{subfigure}{0.07\textwidth}
	\includegraphics[width=0.99\textwidth]{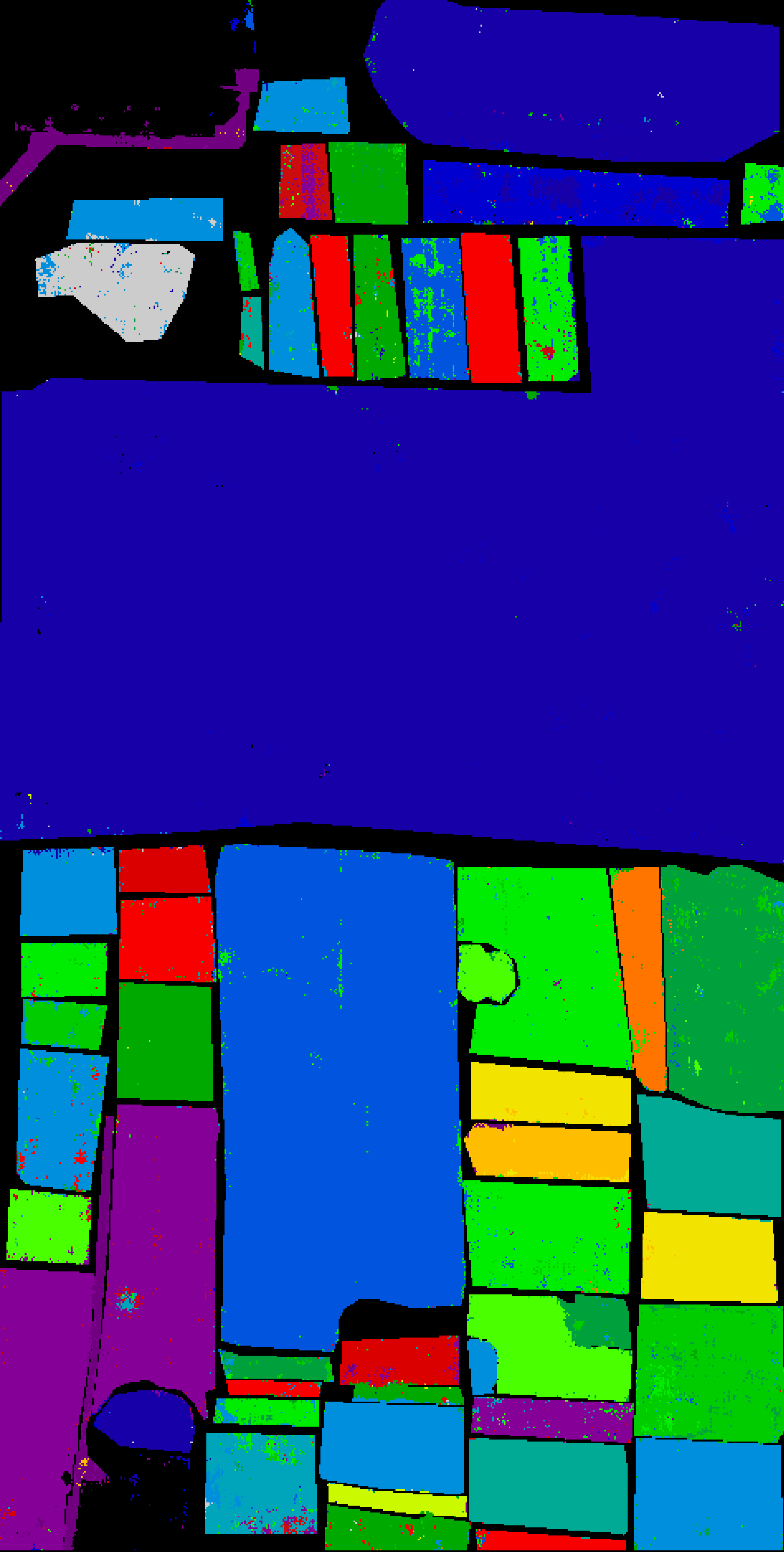}
	\caption*{9512}
    \end{subfigure} 
    \begin{subfigure}{0.07\textwidth}
	\includegraphics[width=0.99\textwidth]{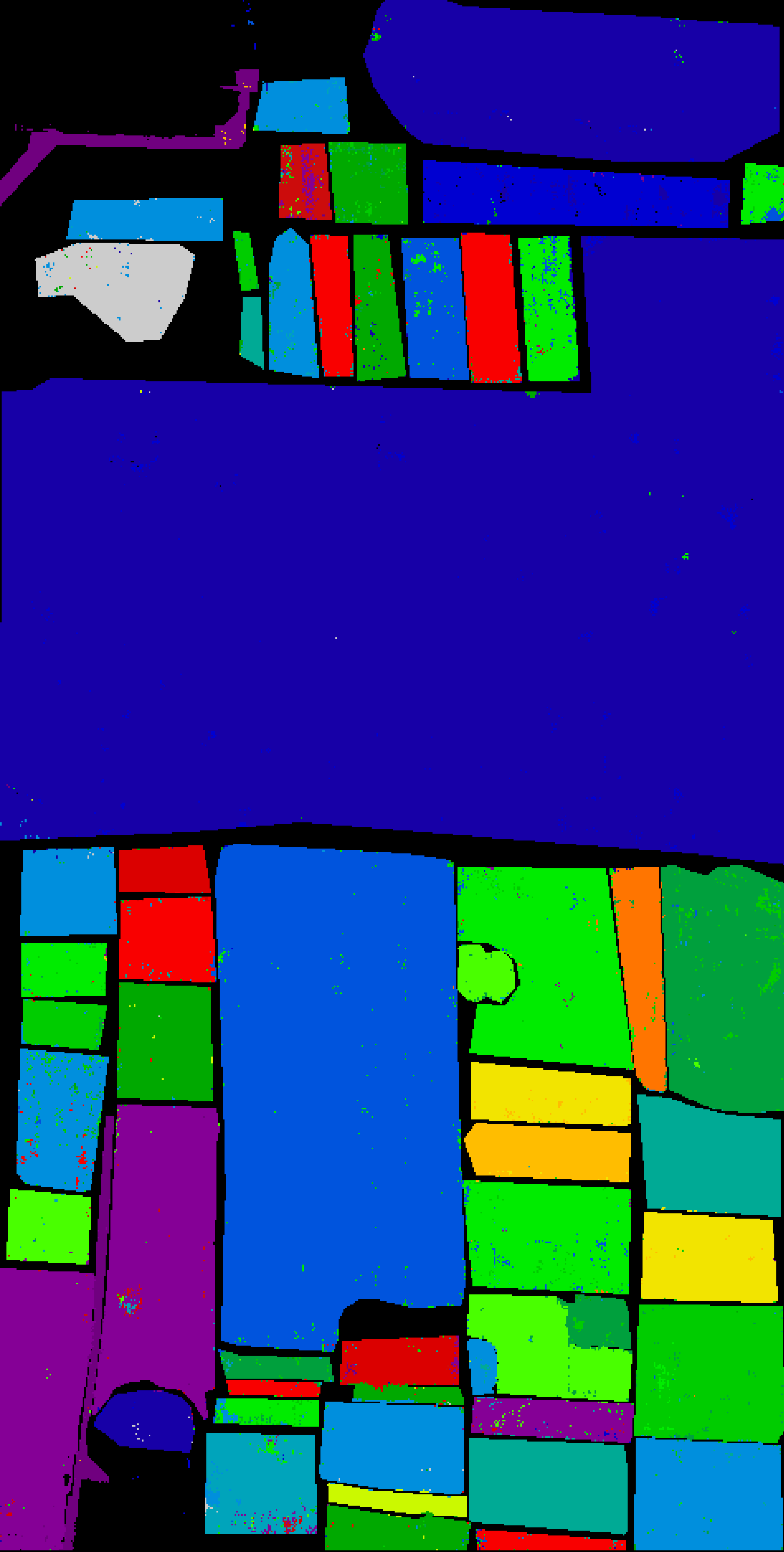}
	\caption*{13188}
    \end{subfigure}
    \begin{subfigure}{0.07\textwidth}
	\includegraphics[width=0.99\textwidth]{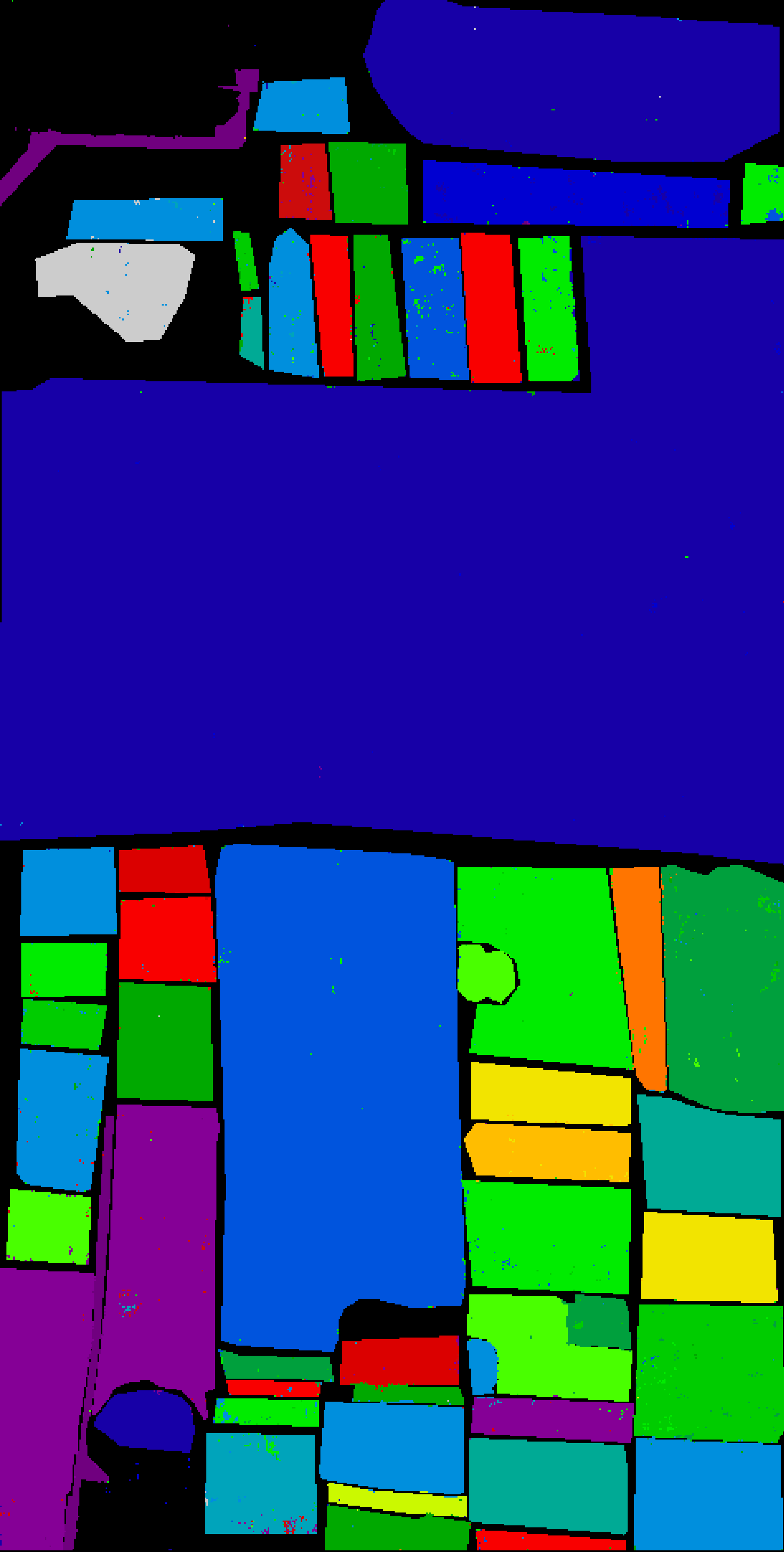}
	\caption*{16791}
    \end{subfigure}
    \begin{subfigure}{0.07\textwidth}
	\includegraphics[width=0.99\textwidth]{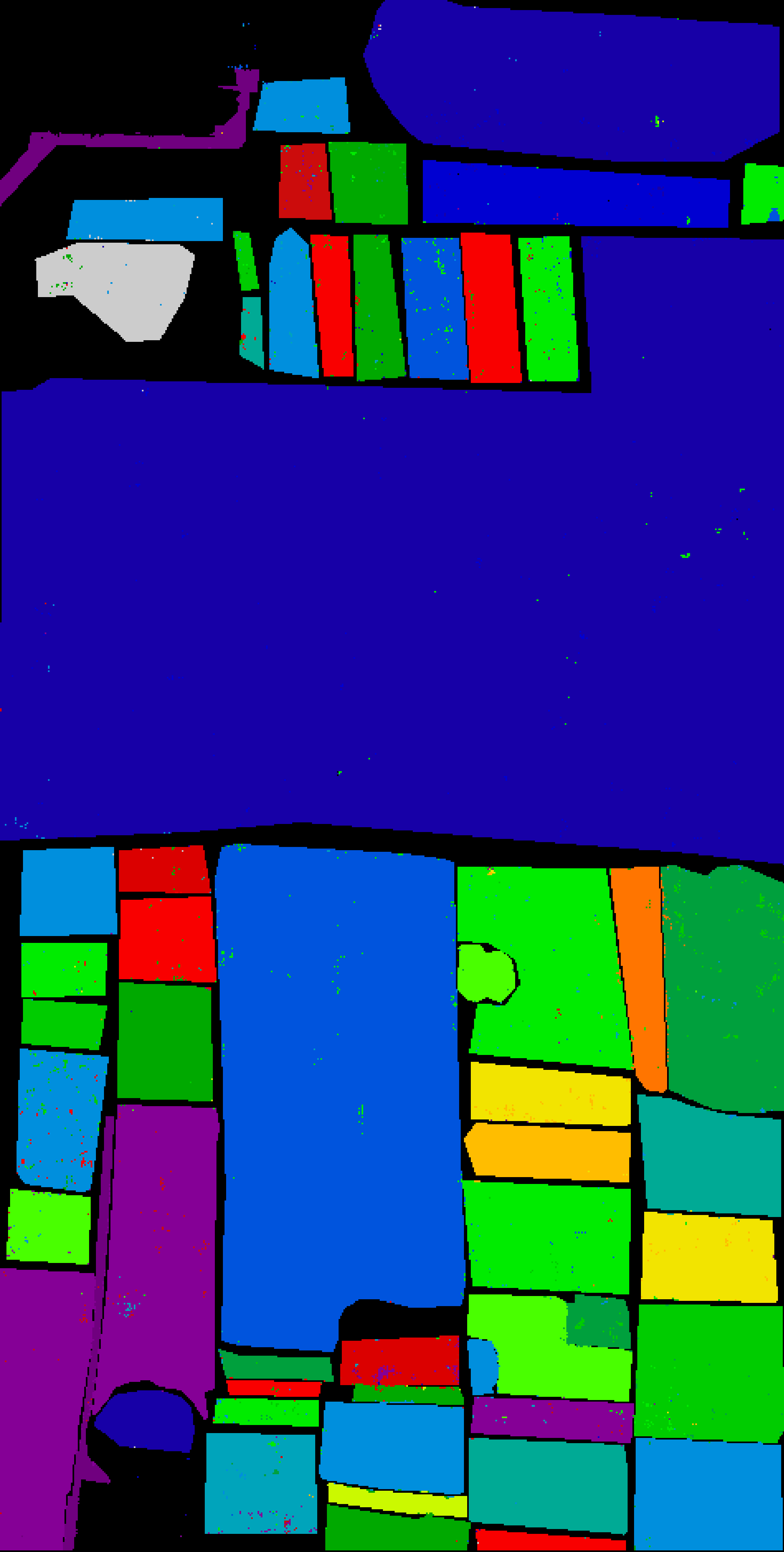}
	\caption*{20322}
    \end{subfigure}
\caption{\textbf{HH dataset}: Ground truth maps corresponding to varying numbers of training samples.}
\label{HHGT}
\end{figure}

Table \ref{HHT} provides computational metrics; despite increasing data, the training and testing times remain relatively stable, showcasing the model's computational efficiency. For example, training times range from 272.05s to 291.68s, and testing times fluctuate minimally between 22.42s and 23.42s. For instance, OA increased from 87.83\% to 98.97\%, while AA rose from 71.14\% to 97.94\%, reflecting the model's robust performance. The $\kappa$ followed a similar upward trajectory, indicating reduced misclassification rates and alignment with ground truth. Figure \ref{HHGT} presents ground truth maps corresponding to the training sample counts. These visualizations show the gradual enrichment of spatial details and class boundaries as training data increases. The progression from 1933 to 20,322 samples demonstrates the model's growing ability to capture intricate spatial and spectral variations.

%%%%%%%%%%%%%%%%%%%%%%%%%%%%%%%
\begin{table}[!hbt]
    \centering
    \caption{\textbf{HH Dataset}: Accuracy improvements as the sample count increases per iteration.}
    \begin{tabular}{c|cccccc} \hline 
        \multirow{2}{*}{\textbf{Class}} & \multicolumn{6}{c}{\textbf{Number of Training Samples}} \\ \cline{2-7}
        & 1933 & 5761 & 9512 & 13188 & 16791 & 20322 \\ \hline 
        
        $\kappa$ & 84.51 & 92.38 & 96.07 & 97.36 & 98.82 & 98.69 \\
        OA & 87.82 & 93.98 & 96.89 & 97.91 & 99.07 & 98.968 \\
        AA & 71.13 & 86.73 & 92.80 & 95.62 & 97.86 & 97.93 \\ \hline 

        Train (s) & 272.05 & 273.63 & 277.96 & 285.60 & 291.45 & 291.68 \\ 
        Test (s) & 23.30 & 22.56 & 22.42 & 22.73 & 23.42 & 22.89 \\ \hline 
        Flops & \multicolumn{6}{c}{\textbf{112128}} \\
        Param & \multicolumn{6}{c}{\textbf{838358}} \\ \hline 
    \end{tabular}
    \label{HHT}
\end{table}
%%%%%%%%%%%%%%%%%%%%%%%%%%%%%%%
%%%%%%%%%%%%%%%%%%%%%%%%%%%%%%%

\begin{figure}[!hbt]
    \centering
    \begin{subfigure}{0.07\textwidth}
	\includegraphics[width=0.99\textwidth]{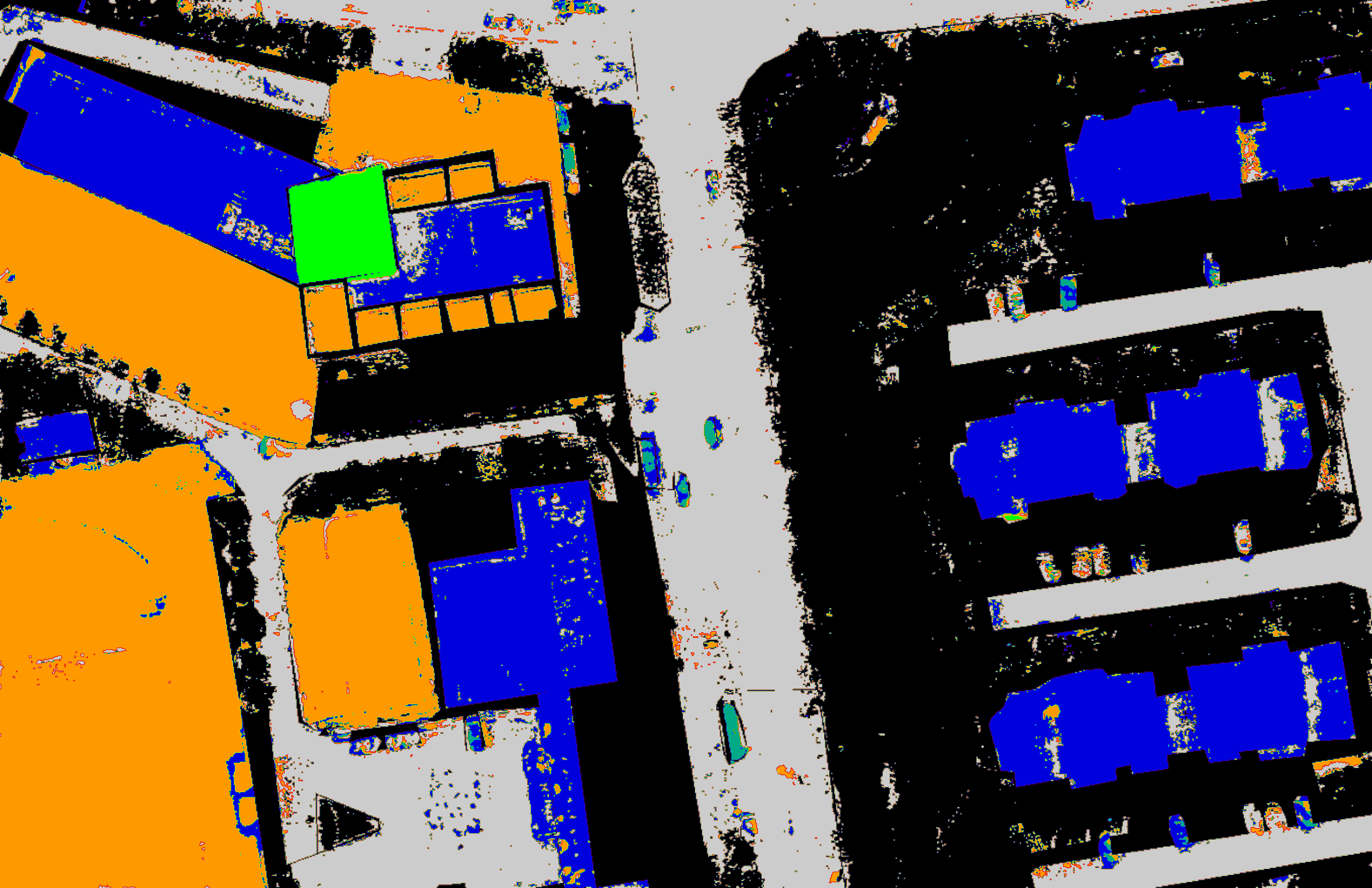}
	\caption*{4774} 
    \end{subfigure}
    \begin{subfigure}{0.07\textwidth}
	\includegraphics[width=0.99\textwidth]{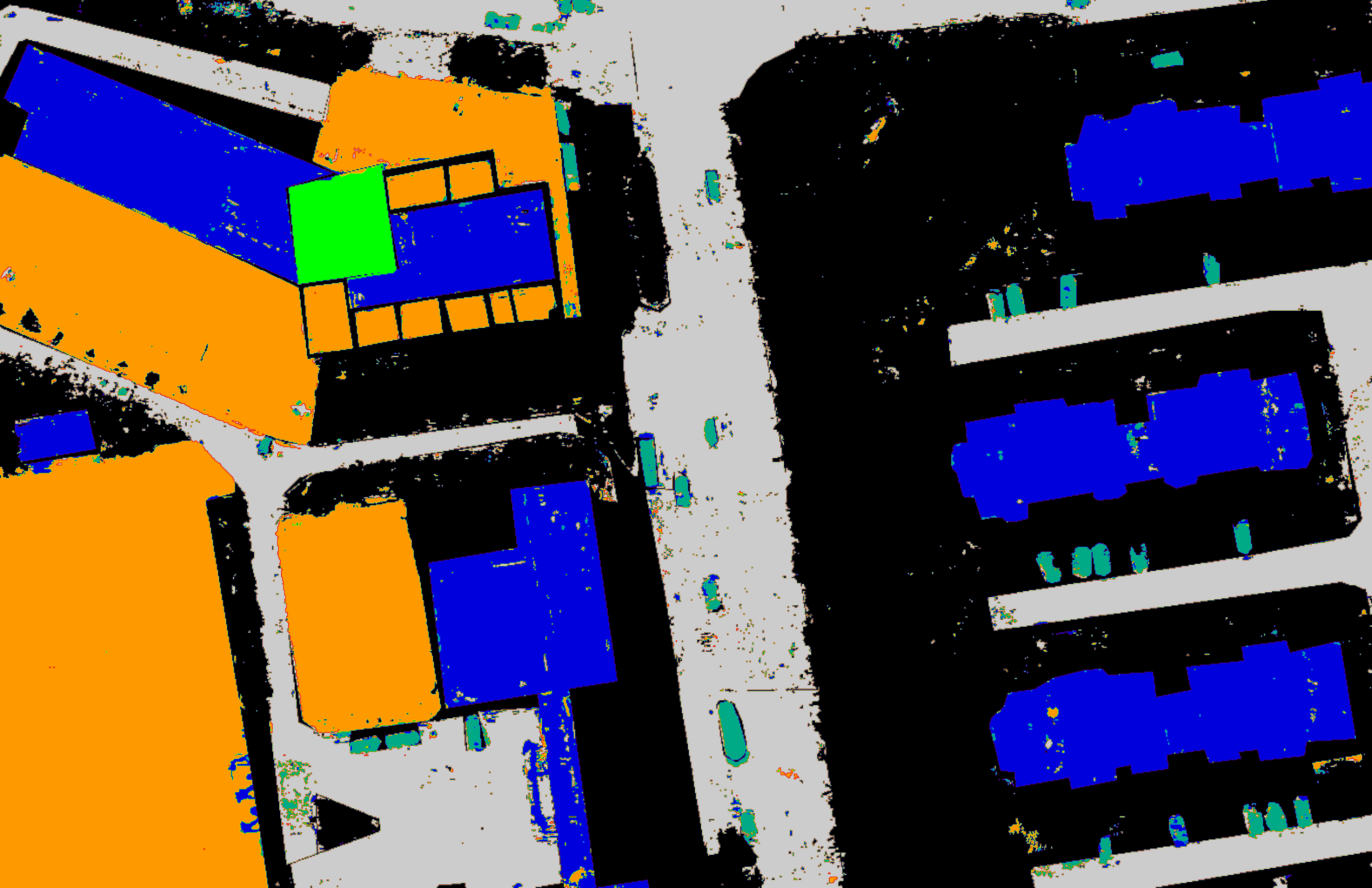}
	\caption*{14227}
    \end{subfigure}
    \begin{subfigure}{0.07\textwidth}
	\includegraphics[width=0.99\textwidth]{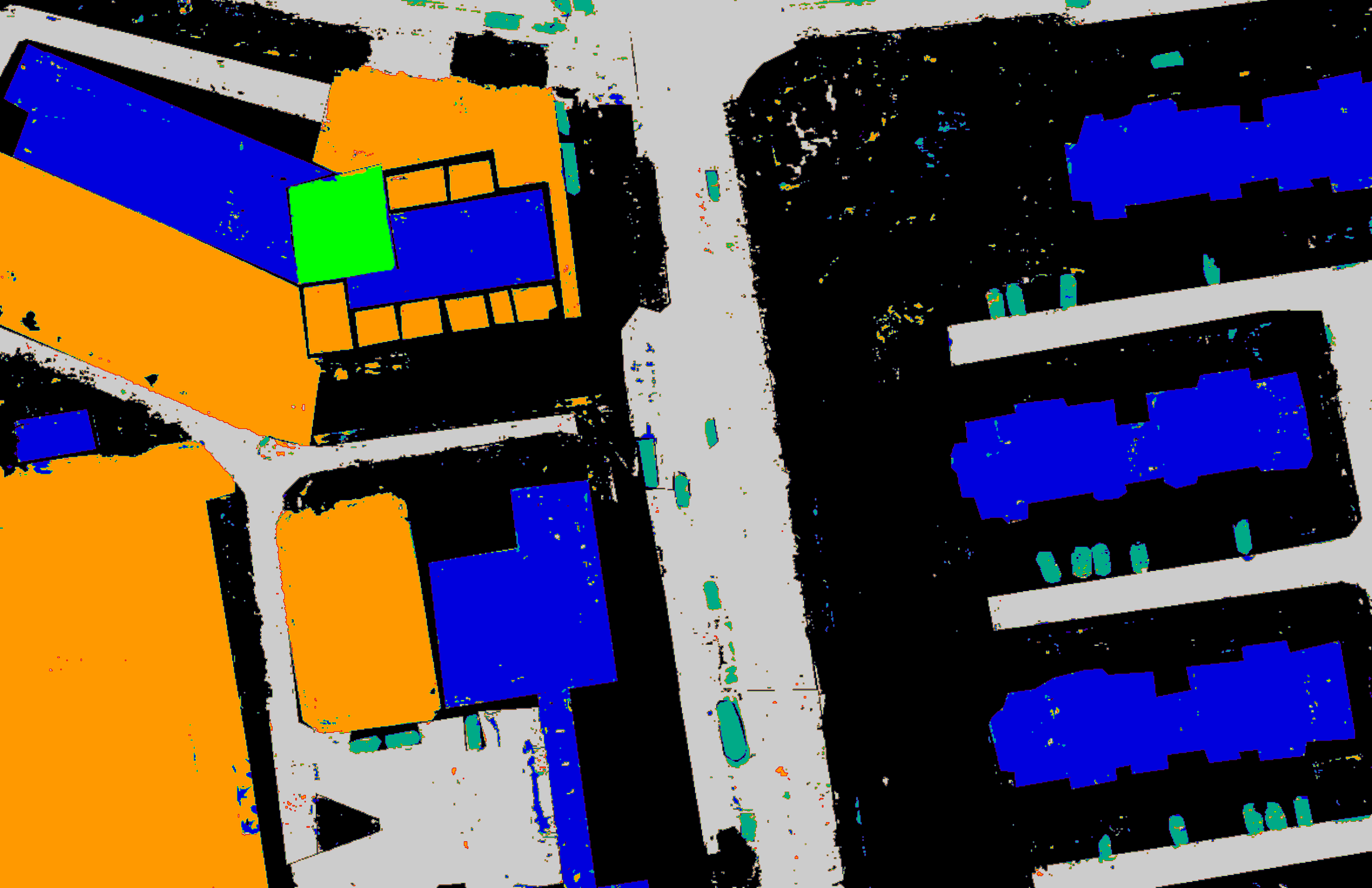}
	\caption*{23491}
    \end{subfigure} 
    \begin{subfigure}{0.07\textwidth}
	\includegraphics[width=0.99\textwidth]{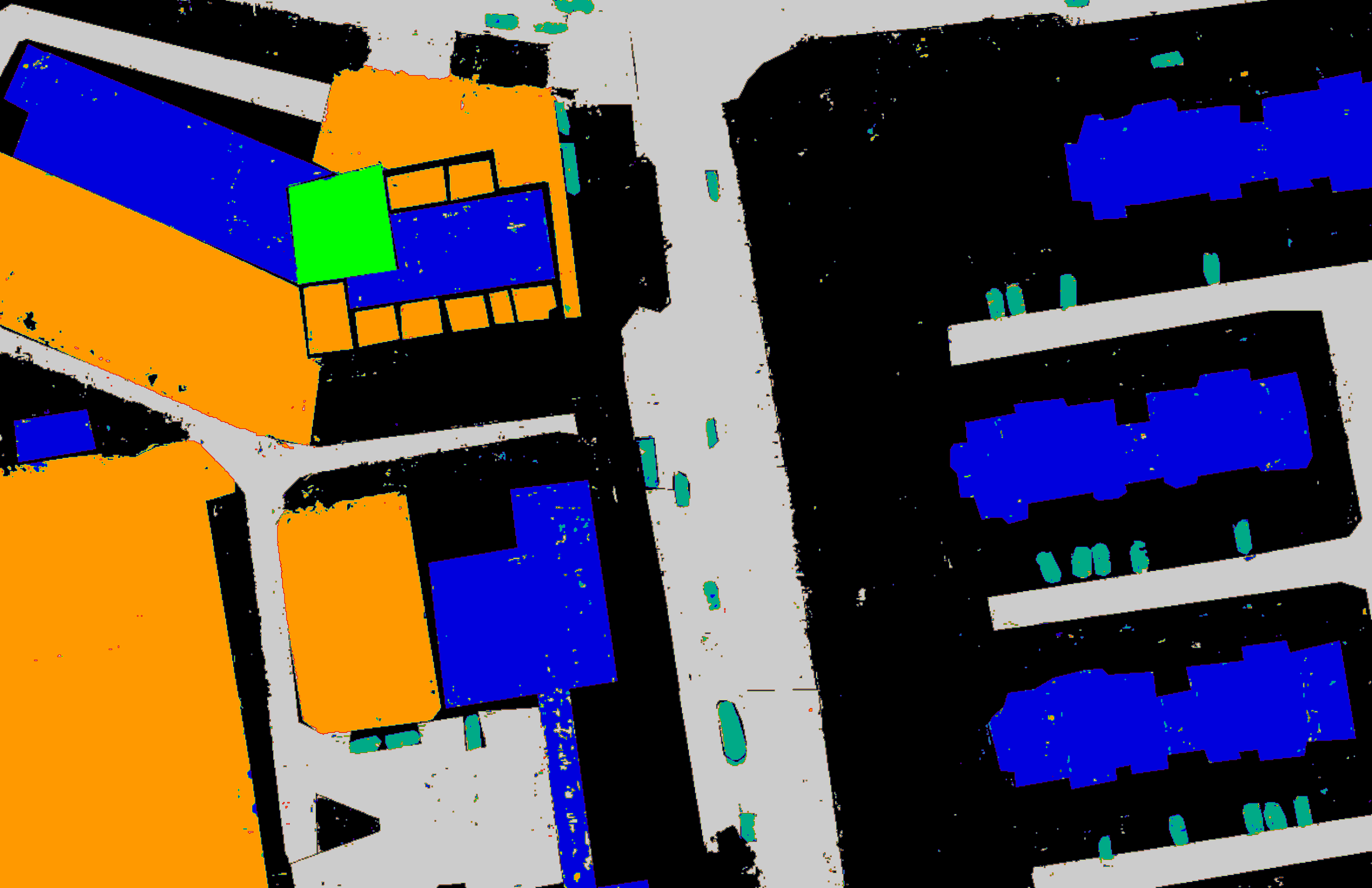}
	\caption*{32570}
    \end{subfigure}
    \begin{subfigure}{0.07\textwidth}
	\includegraphics[width=0.99\textwidth]{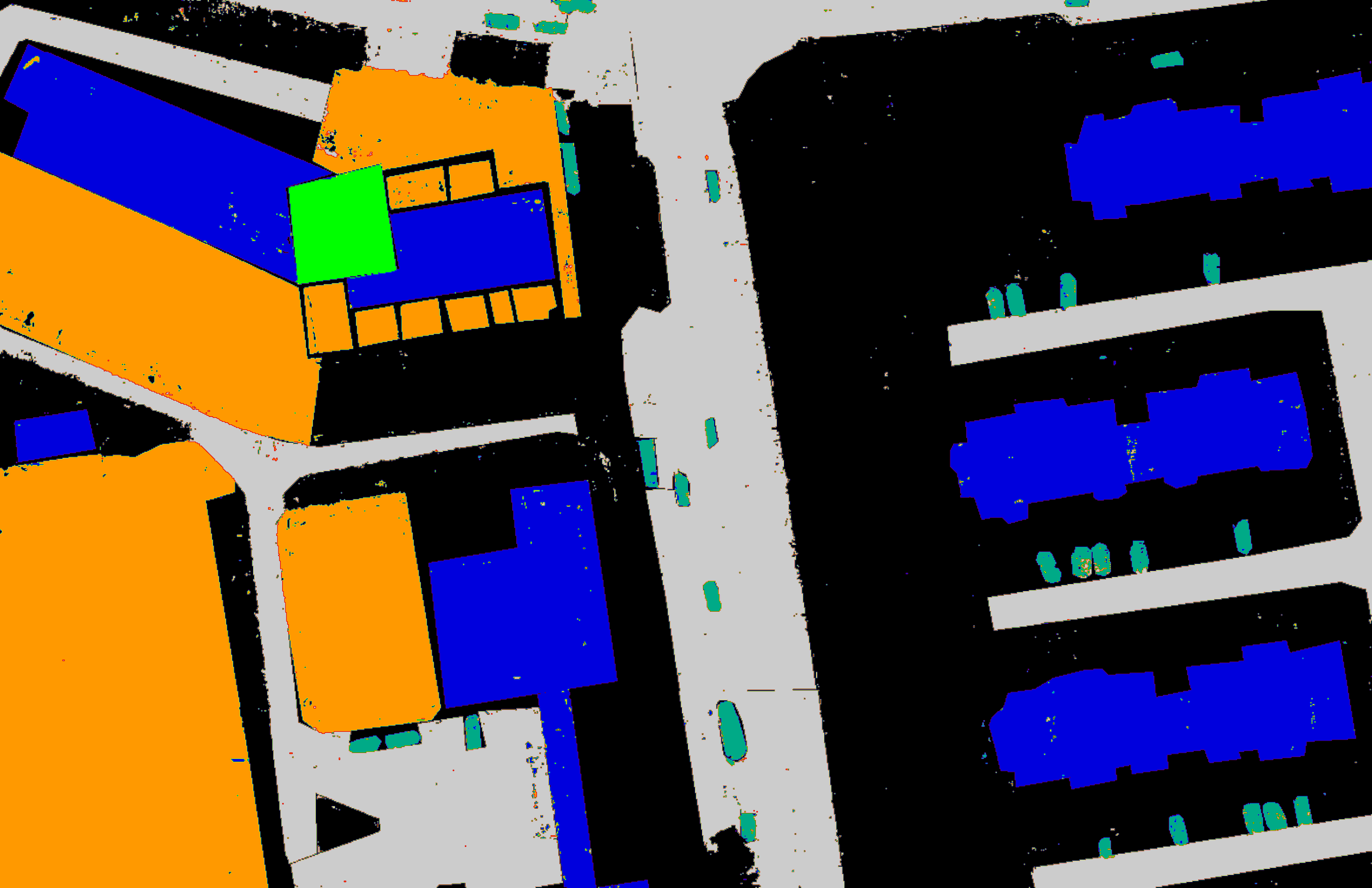}
	\caption*{41467}
    \end{subfigure}
    \begin{subfigure}{0.07\textwidth}
	\includegraphics[width=0.99\textwidth]{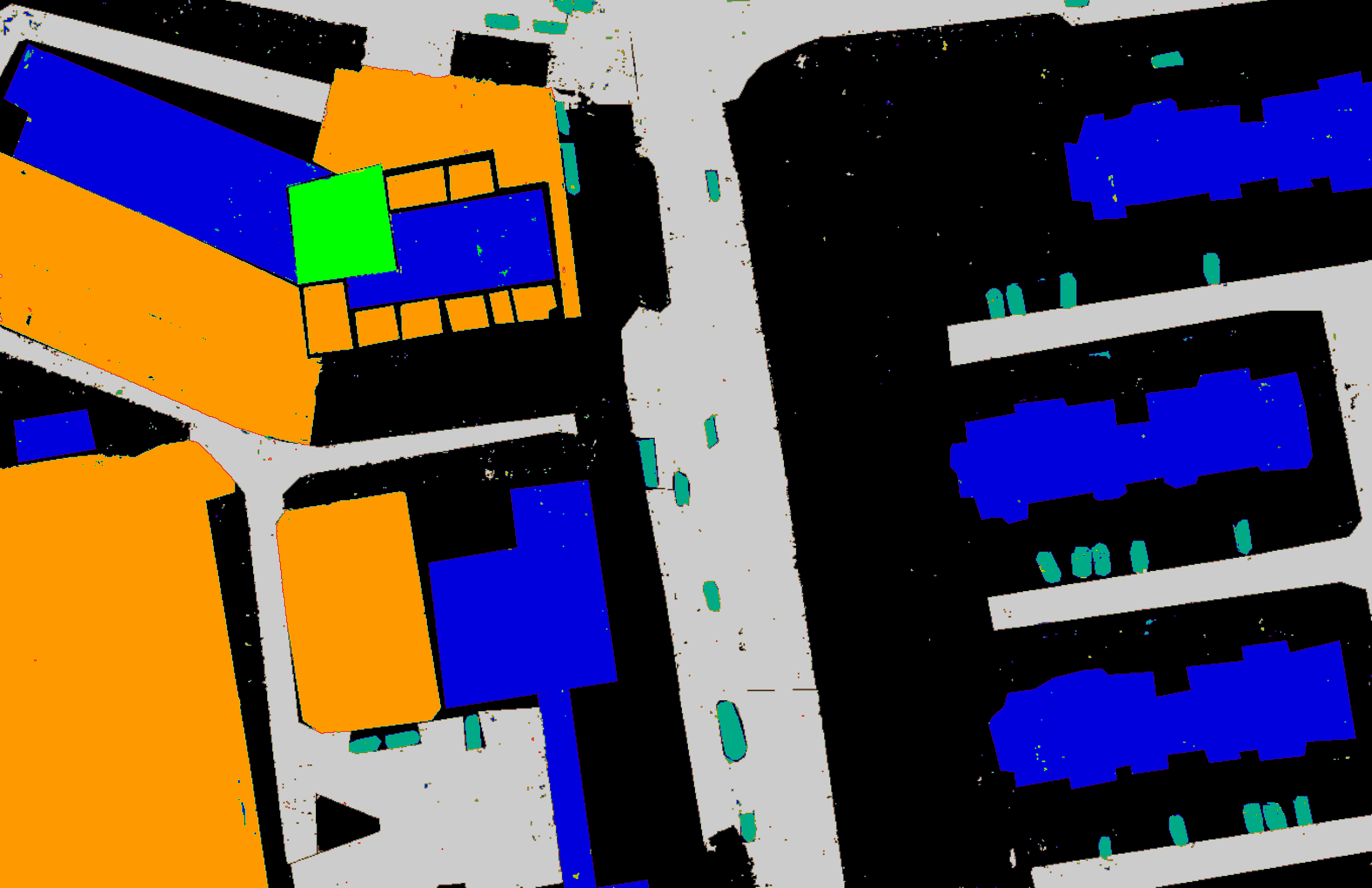}
	\caption*{50186}
    \end{subfigure}
\caption{\textbf{QUH dataset}: Ground truth maps corresponding to varying numbers of training samples.}
\label{QUHGT}
\end{figure}

Table \ref{QUHT} reports training and testing times, which grow modestly with larger sample counts, and provides the model's computational metrics, such as FLOPs and parameters, indicating efficient scaling. Figure \ref{QUHGT} illustrates ground truth maps for the dataset, corresponding to each training sample size. These maps visually demonstrate the enhancement in spatial coverage and label consistency as the number of training samples increases, corroborating the quantitative improvements observed in Table \ref{QUHT}. 

%%%%%%%%%%%%%%%%%%%%%%%%%%%%%%%
\begin{table}[!hbt]
    \centering
    \caption{\textbf{QUH Dataset}: Accuracy improvements as the sample count increases per iteration.}
    \begin{tabular}{c|cccccc} \hline 
        \multirow{2}{*}{\textbf{Class}} & \multicolumn{6}{c}{\textbf{Number of Training Samples}} \\ \cline{2-7}
        & 4774 & 14227 & 23491 & 32570 & 41467 & 50186 \\ \hline 

        $\kappa$ & 88.59 & 95.18 & 96.72 & 98.32 & 98.64 & 99.04 \\
        OA & 91.41 & 96.35 & 97.52 & 98.73 & 98.97 & 99.28 \\
        AA & 81.15 & 94.52 & 96.29 & 97.91 & 97.92 & 98.95 \\ \hline 

        Train (s) & 2000.15 & 2122.98 & 2152.50 & 2248.08 & 2296.25 & 2360.75 \\ 
        Test (s) & 219.09 & 227.46 & 231.50 & 239.75 & 239.00 & 234.79 \\ \hline 
        Flops & \multicolumn{6}{c}{\textbf{108032}} \\
        Param & \multicolumn{6}{c}{\textbf{834246}} \\ \hline 
    \end{tabular}
    \label{QUHT}
\end{table}
%%%%%%%%%%%%%%%%%%%%%%%%%%%%%%%
%%%%%%%%%%%%%%%%%%%%%%%%%%%%%%%

%%%%%%%%%%%%%%%%%%%%%%%%%%%%%%%
\section{ATL-SST Effects on Cross Datasets}

This section explores the impact of ATL in enabling effective cross-dataset classification. By leveraging the actively trained SST on one dataset and fine-tuning it with a limited fraction of samples from other datasets, we assess its ability to generalize and adapt to varying spectral-spatial distributions. The results highlight the significance of ATL in achieving high accuracy with minimal fine-tuning across diverse datasets.

%%%%%%%%%%%%%%%%%%%%%%%%%%%%%%%
\begin{table}[!hbt]
    \centering
    \caption{The OA, AA, and $\kappa$ accuracies for cross-dataset classification were obtained using the actively trained SST on the SA dataset and evaluated across other datasets.}
    \begin{tabular}{c||c||ccccc} \hline 
        \multirow{2}{*}{\textbf{Metric}} & \textbf{SST Trained} & \multicolumn{5}{c}{\textbf{SST Fine-Tuned using 10\% Samples}} \\ \cline{2-7} 
        & SA & PU & UH & HC & HH & QUH \\ \hline 
        $\kappa$ & 99.42 & 96.42 & 97.95 & 93.95 & 93.94 & 95.70 \\ 
        OA & 99.48 & 97.30 & 98.11 & 94.83 & 95.20 & 96.75 \\ 
        AA & 99.76 & 95.47 & 97.89 & 90.14 & 89.30 & 93.94 \\ \hline 
        Test (s) & 11.49 & 10.31 & 5.46 & 61.11 & 128.04 & 252.74 \\ \hline 
    \end{tabular}
    \label{ASAT}
\end{table}
%%%%%%%%%%%%%%%%%%%%%%%%%%%%%%%
\begin{figure}[!hbt]
    \centering
    \begin{subfigure}{0.15\textwidth}
	\includegraphics[width=0.99\textwidth]{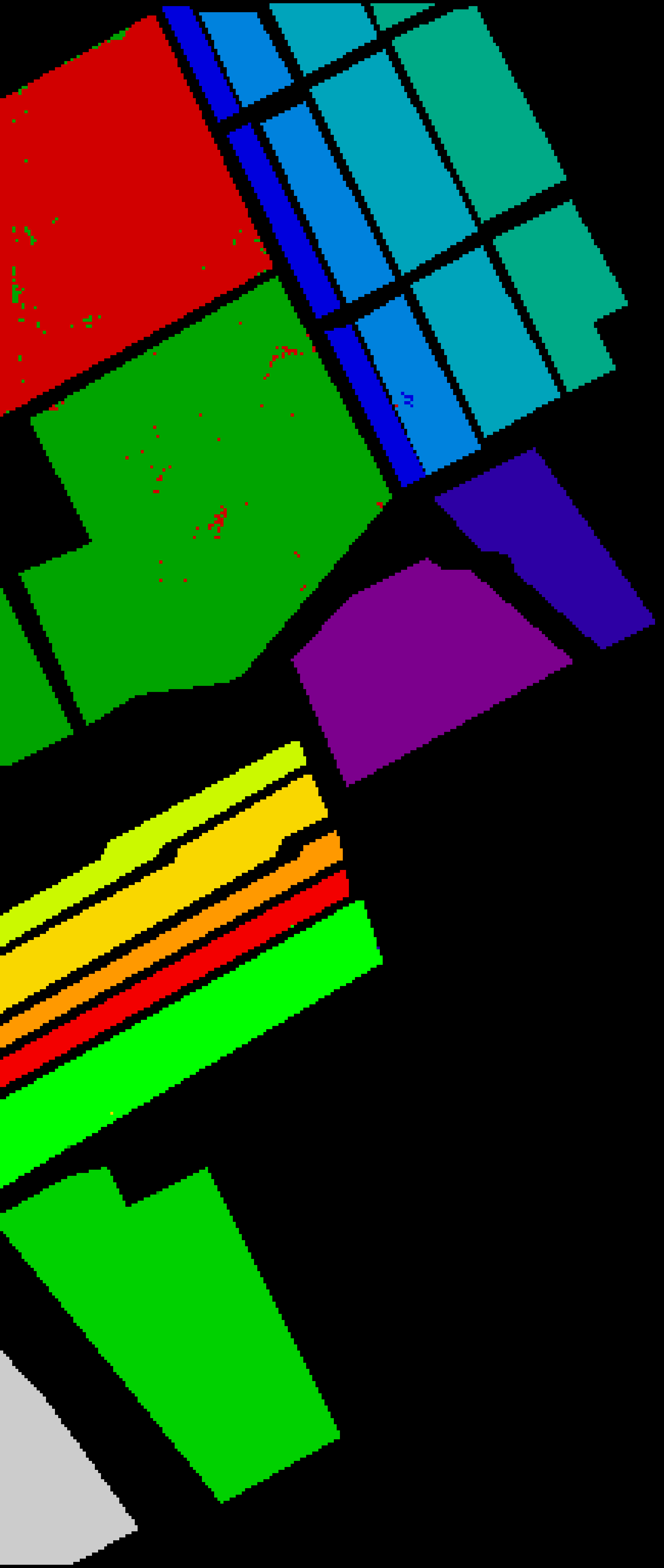}
	\caption*{Trained-SA} 
    \end{subfigure}
    \begin{subfigure}{0.15\textwidth}
	\includegraphics[width=0.99\textwidth]{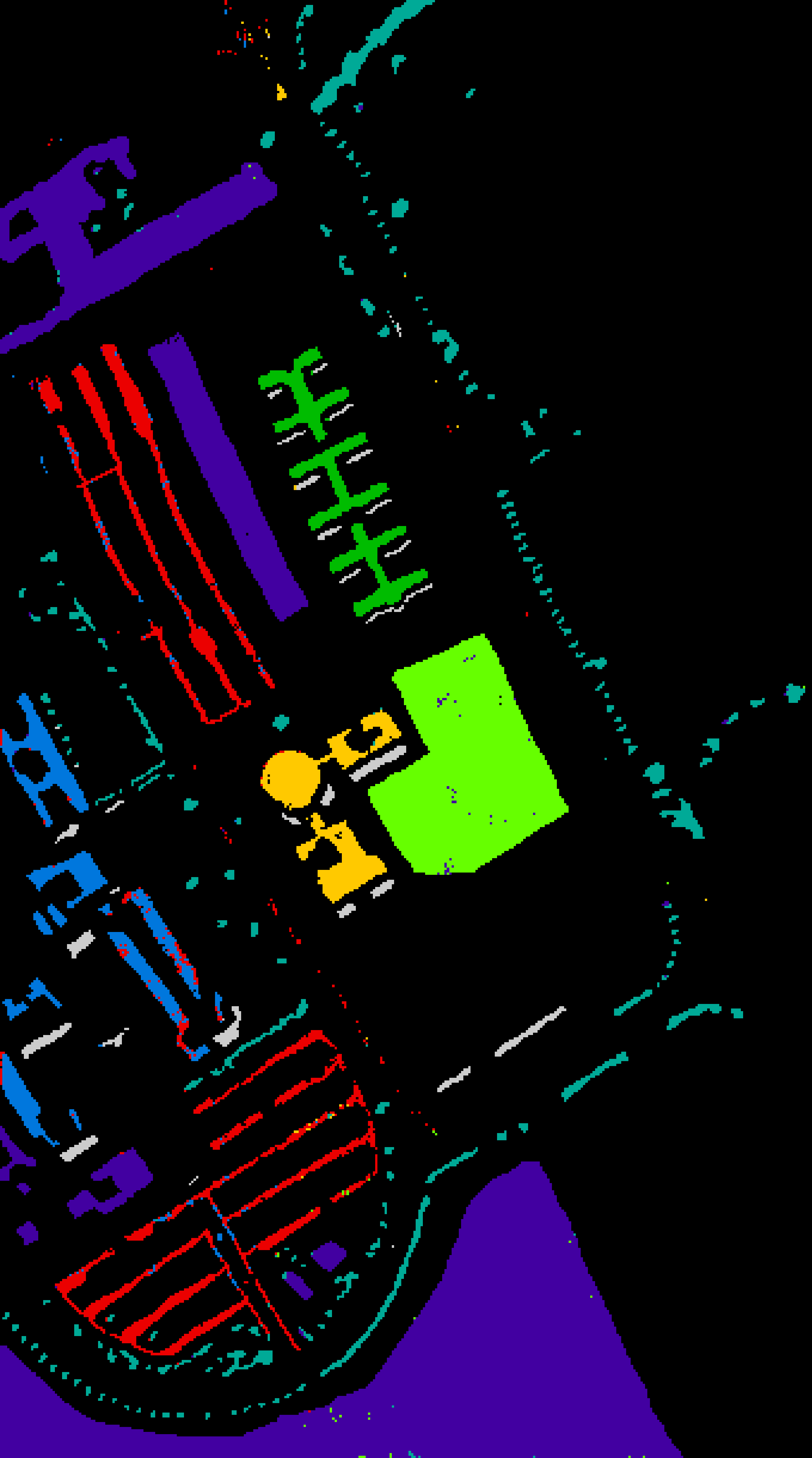}
	\caption*{Test-PU}
    \end{subfigure}
    \begin{subfigure}{0.15\textwidth}
	\includegraphics[width=0.99\textwidth]{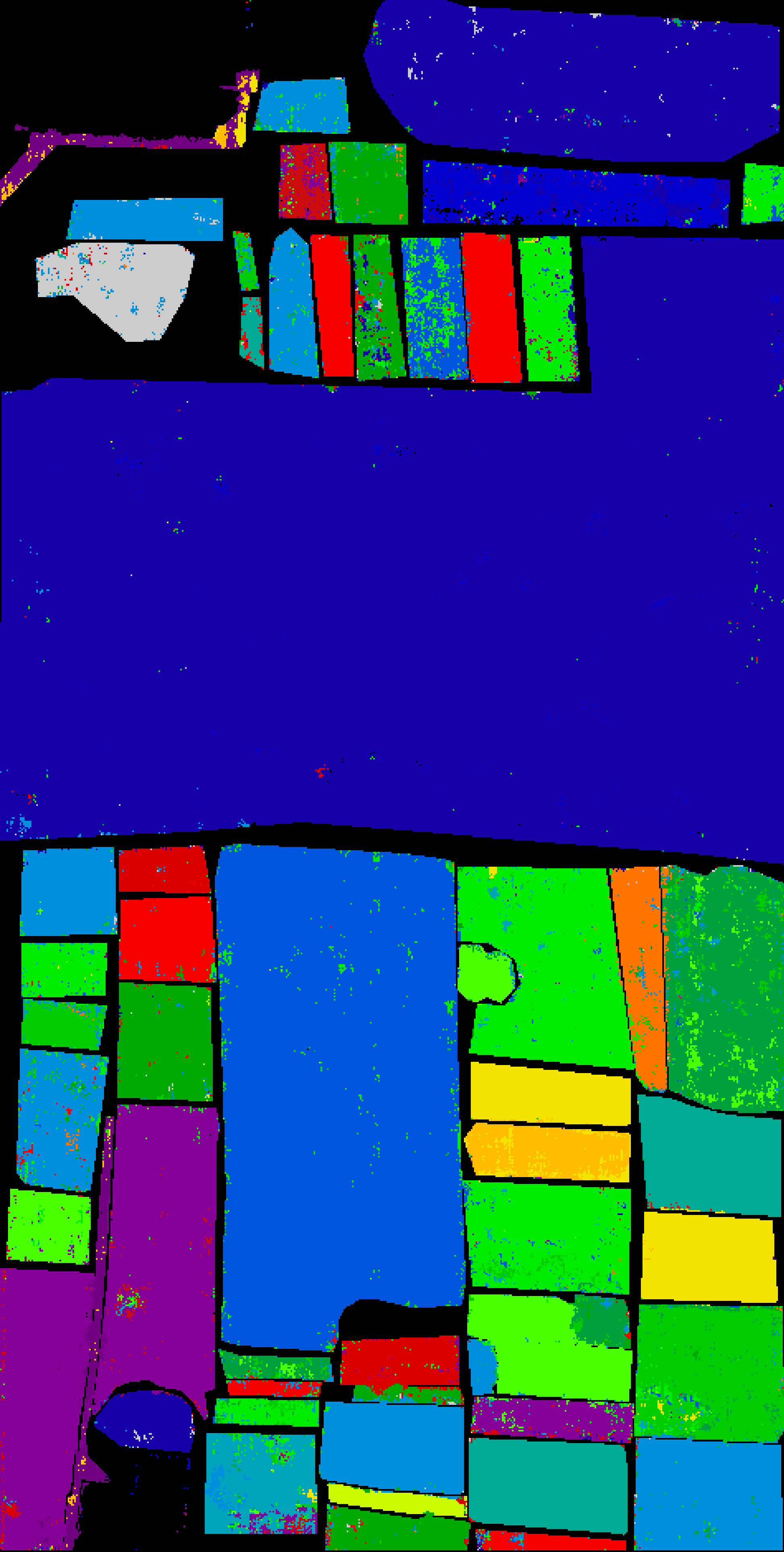}
	\caption*{Test-HH}
    \end{subfigure}
    \begin{subfigure}{0.49\textwidth}
	\includegraphics[width=0.99\textwidth]{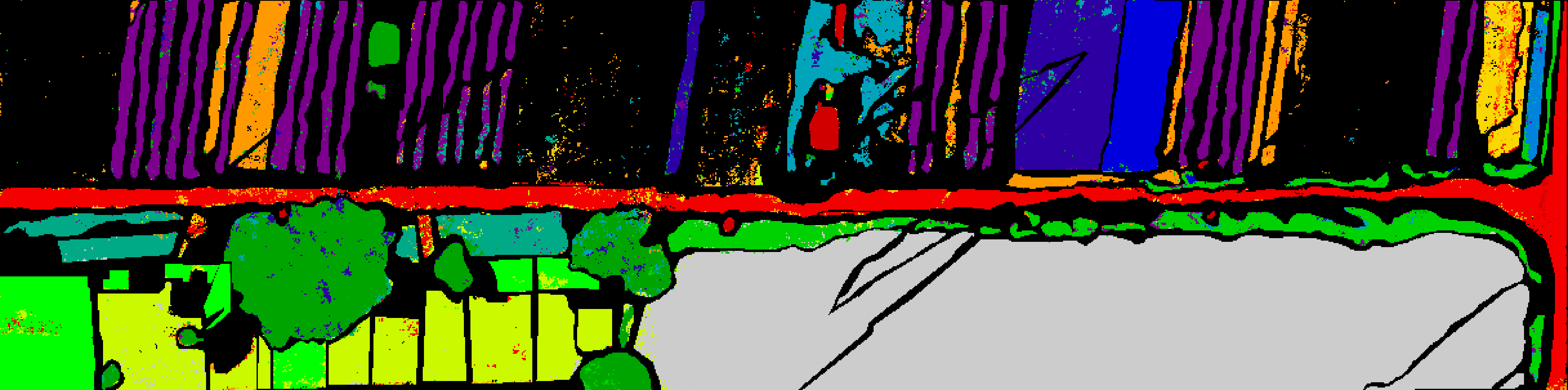}
	\caption*{Test-HC}
    \end{subfigure} 
    \begin{subfigure}{0.49\textwidth}
	\includegraphics[width=0.99\textwidth]{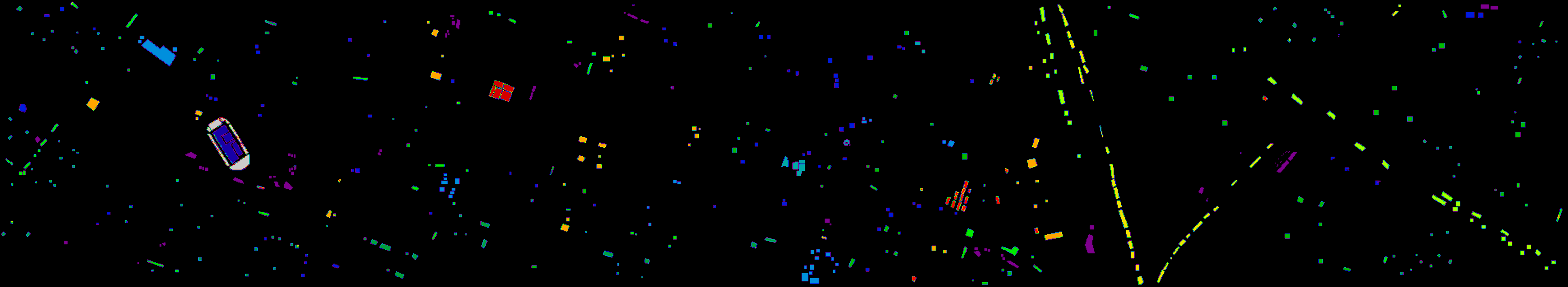}
	\caption*{Test-UH}
    \end{subfigure}
\caption{\textbf{SST trained on SA} was fine-tuned and tested on PU, HC, HH, and UH datasets, respectively.}
\label{ASA}
\end{figure}
%%%%%%%%%%%%%%%%%%%%%%%%%%%%%%%

Table \ref{ASAT} presents the OA, AA, and $\kappa$ for cross-dataset classification, where the ATL-SST is initially trained on the SA dataset and subsequently fine-tuned using 10\% of the samples from other datasets (PU, UH, HC, HH, and QUH). The results indicate that the model achieves high accuracy across datasets, with $\kappa$ values consistently above 93\%, demonstrating effective transfer learning. Testing times vary significantly across datasets due to differences in data volume and complexity, with QUH requiring the longest computation times. Figure \ref{ASA} provides visual representations of classification outcomes, showcasing the trained SST on SA and its fine-tuned performance on PU, HC, HH, and UH datasets. Each subfigure illustrates distinct patterns, emphasizing the model’s adaptability to different data distributions.

\begin{figure}[!hbt]
    \centering
    \begin{subfigure}{0.15\textwidth}
	\includegraphics[width=0.99\textwidth]{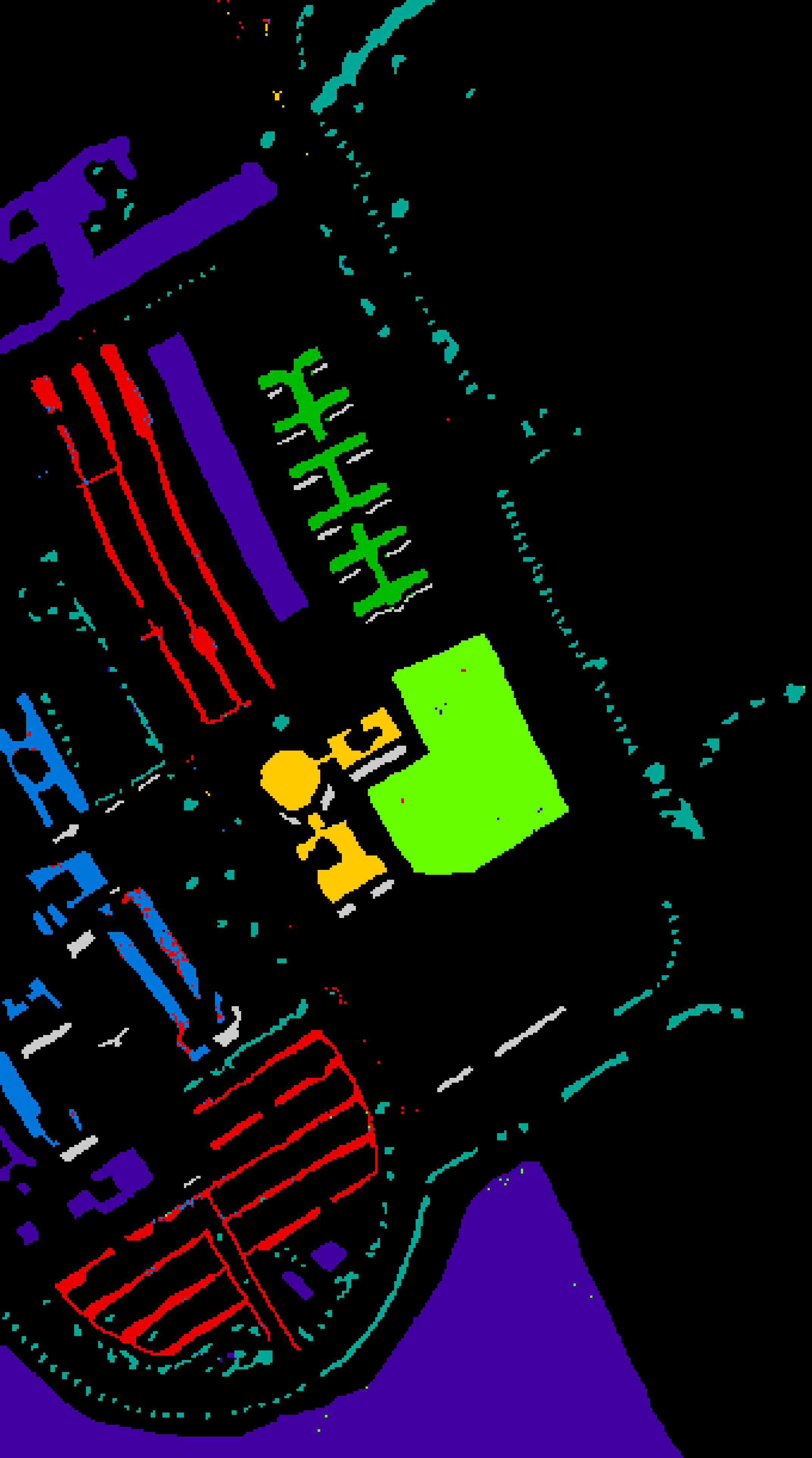}
	\caption*{Trained-PU} 
    \end{subfigure}
    \begin{subfigure}{0.15\textwidth}
	\includegraphics[width=0.99\textwidth]{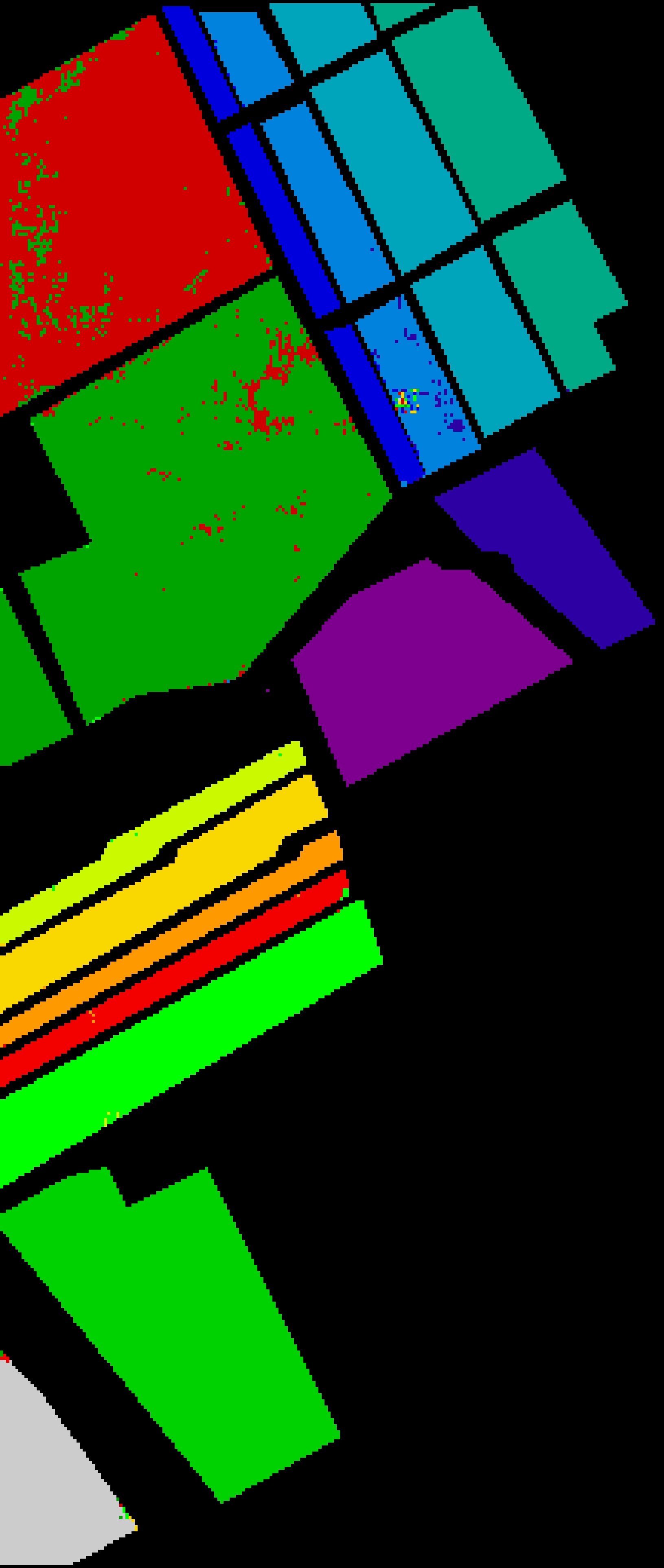}
	\caption*{Test-SA}
    \end{subfigure}
    \begin{subfigure}{0.15\textwidth}
	\includegraphics[width=0.99\textwidth]{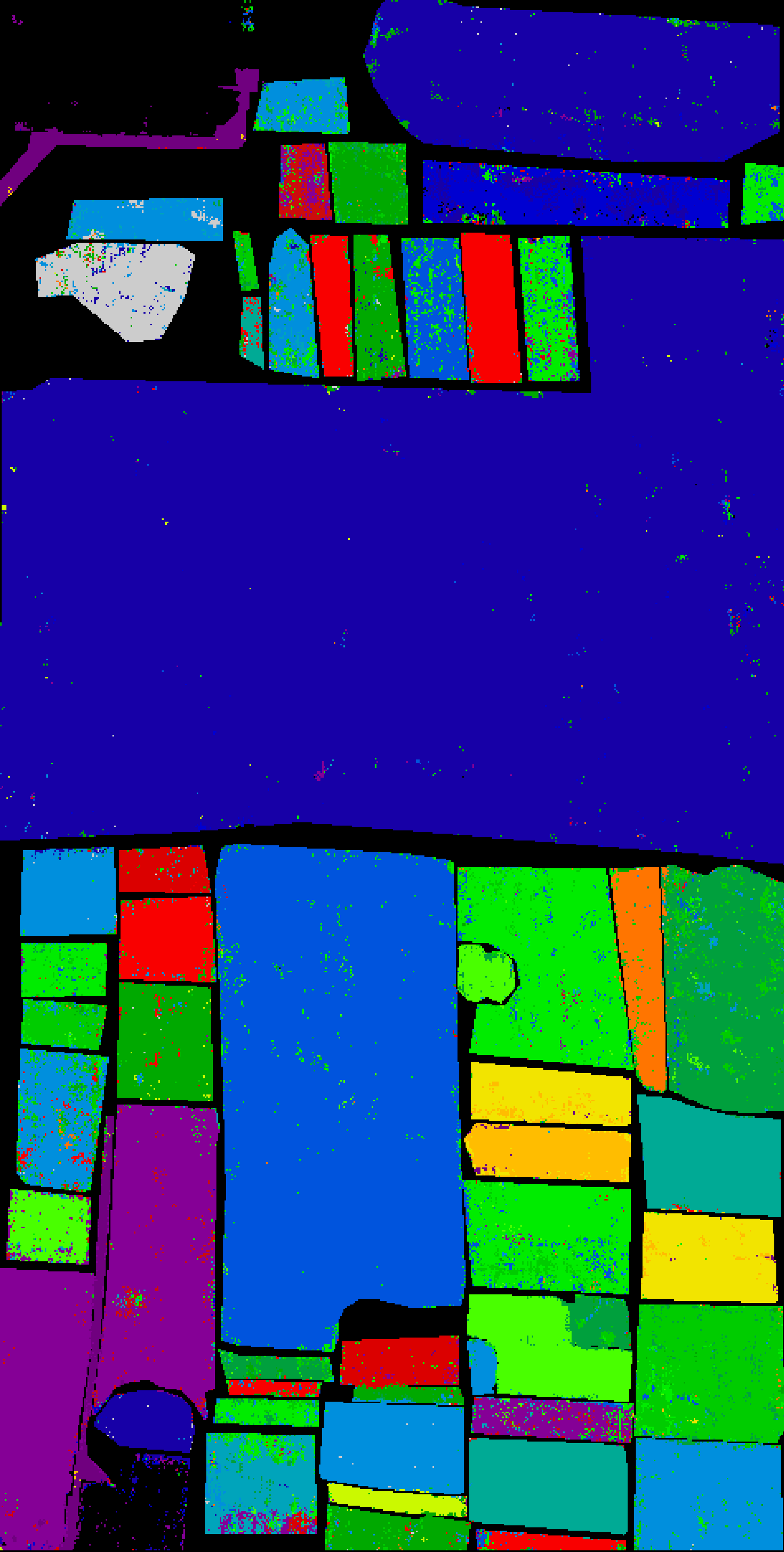}
	\caption*{Test-HH}
    \end{subfigure}
    \begin{subfigure}{0.49\textwidth}
	\includegraphics[width=0.99\textwidth]{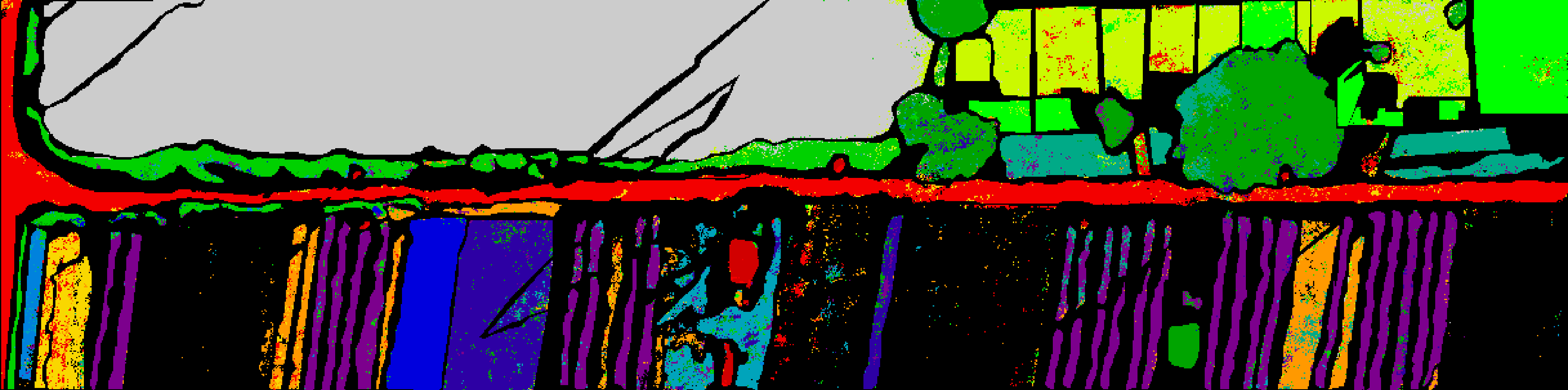}
	\caption*{Test-HC}
    \end{subfigure} 
    \begin{subfigure}{0.49\textwidth}
	\includegraphics[width=0.99\textwidth]{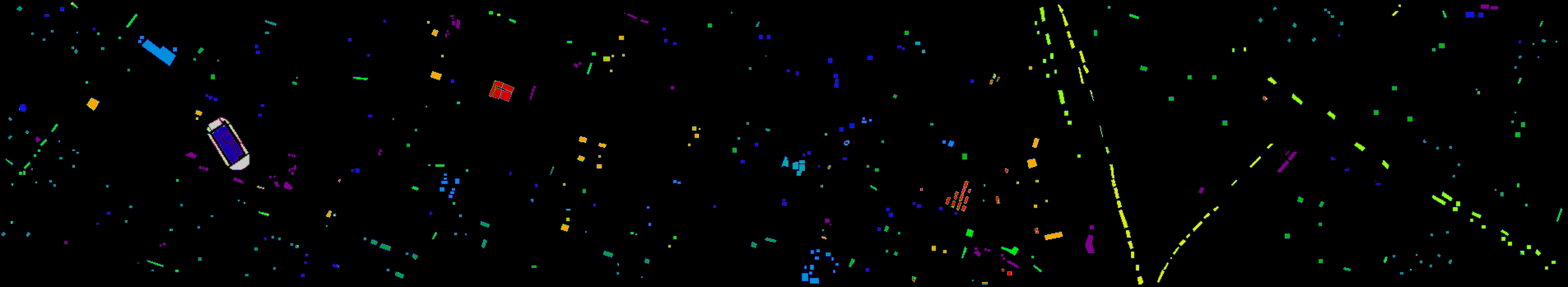}
	\caption*{Test-UH}
    \end{subfigure}
\caption{\textbf{SST trained on PU} was fine-tuned and tested on SA, HC, HH, and UH datasets, respectively.}
\label{APU}
\end{figure}

Table \ref{APUT} and Figure \ref{APU} illustrate the performance of the SST trained on the PU dataset and subsequently fine-tuned and evaluated on other datasets, including SA, UH, HC, and HH. Table \ref{APUT} provides detailed metrics, highlighting the transferability and generalization ability of the SST model. The actively trained SST on PU achieves a $\kappa$ of 98.51\% and an OA of 98.88\% on its test data, demonstrating robust performance on the source dataset. When fine-tuned with only 10\% samples from other datasets, the model maintains competitive accuracy, with OA values exceeding 96\% for most target datasets. Testing times are also reported, indicating computational efficiency during model fine-tuning and evaluation. Figure \ref{APU} visually complements these results by showcasing the SST model's classification maps on the source dataset (PU) and various target datasets (SA, HC, HH, and UH). 

%%%%%%%%%%%%%%%%%%%%%%%%%%%%%%%
\begin{table}[!hbt]
    \centering
    \caption{The OA, AA, and $\kappa$ accuracies for cross-dataset classification were obtained using the actively trained SST on the PU dataset and evaluated across other datasets.}
    \begin{tabular}{c||c||ccccc} \hline 
        \multirow{2}{*}{\textbf{Metric}} & \textbf{SST Trained} & \multicolumn{5}{c}{\textbf{SST Fine-Tuned using 10\% Samples}} \\ \cline{2-7} 
        & PU & SA & UH & HC & HH & QUH \\ \hline 
        $\kappa$ & 98.51 & 97.75 & 96.17 & 91.80 & 92.43 & 94.85 \\ 
        OA & 98.87 & 97.98 & 96.46 & 93.00 & 94.01 & 96.11 \\ 
        AA & 98.17 & 98.83 & 96.06 & 87.55 & 87.71 & 92.79 \\ \hline 
        Test (s) & 9.09 & 12.83 & 3.71 & 59.67 & 89.13 & 239.48 \\ \hline 
    \end{tabular}
    \label{APUT}
\end{table}
%%%%%%%%%%%%%%%%%%%%%%%%%%%%%%%
%%%%%%%%%%%%%%%%%%%%%%%%%%%%%%%

%%%%%%%%%%%%%%%%%%%%%%%%%%%%%%%
\begin{table}[!hbt]
    \centering
    \caption{The OA, AA, and $\kappa$ accuracies for cross-dataset classification were obtained using the actively trained SST on the UH dataset and evaluated across other datasets.}
    \resizebox{\columnwidth}{!}{\begin{tabular}{c||c||ccccc} \hline 
        \multirow{2}{*}{\textbf{Metric}} & \textbf{SST Trained} & \multicolumn{5}{c}{\textbf{SST Fine-Tuned using 10\% Samples}} \\ \cline{2-7} 
        & UH & SA & PU & HC & HH & QUH \\ \hline 
        $\kappa$ & 97.38 & 98.29 & 97.01 & 93.26 & 94.04 & 95.60 \\ 
        OA & 97.57 & 98.47 & 97.74 & 94.25 & 95.29 & 96.67 \\ 
        AA & 96.97 & 99.19 & 96.32 & 90.42 & 89.40 & 94.35 \\ \hline 
        Test (s) & 3.44 & 12.85 & 10.81 & 69.98 & 97.30 & 242.65 \\ \hline
    \end{tabular}}
    \label{AUHT}
\end{table}
%%%%%%%%%%%%%%%%%%%%%%%%%%%%%%%
\begin{figure}[!hbt]
    \centering
    \begin{subfigure}{0.49\textwidth}
	\includegraphics[width=0.99\textwidth]{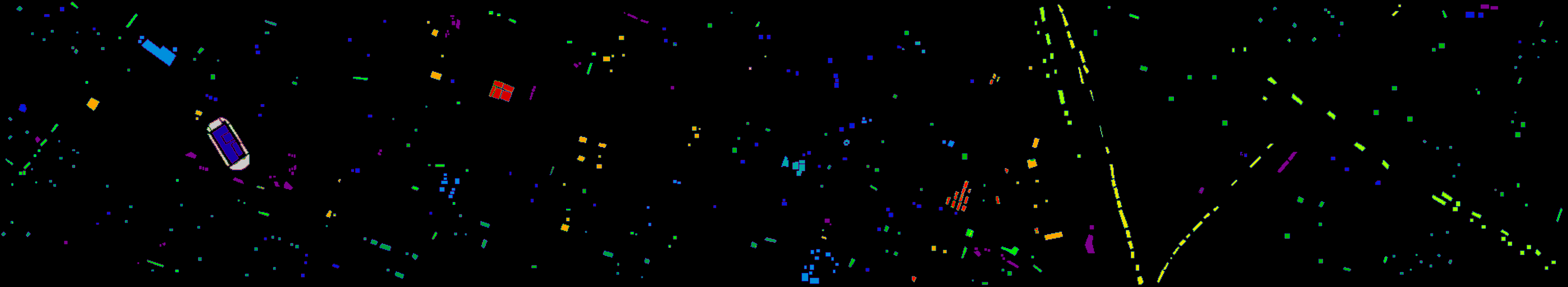}
	\caption*{Trained-UH} 
    \end{subfigure}
    \begin{subfigure}{0.49\textwidth}
	\includegraphics[width=0.99\textwidth]{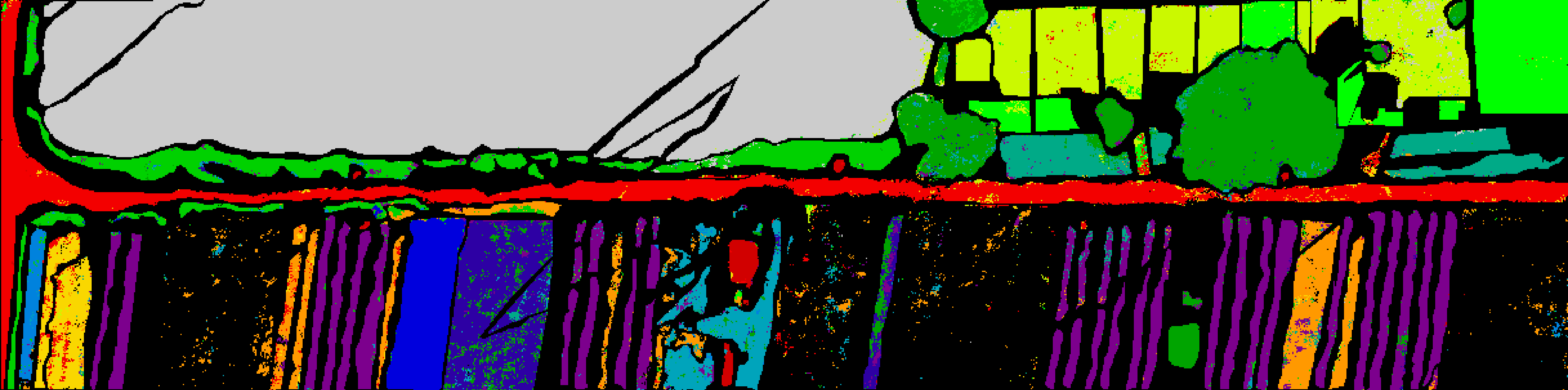}
	\caption*{Test-HC}
    \end{subfigure}
    \begin{subfigure}{0.15\textwidth}
	\includegraphics[width=0.99\textwidth]{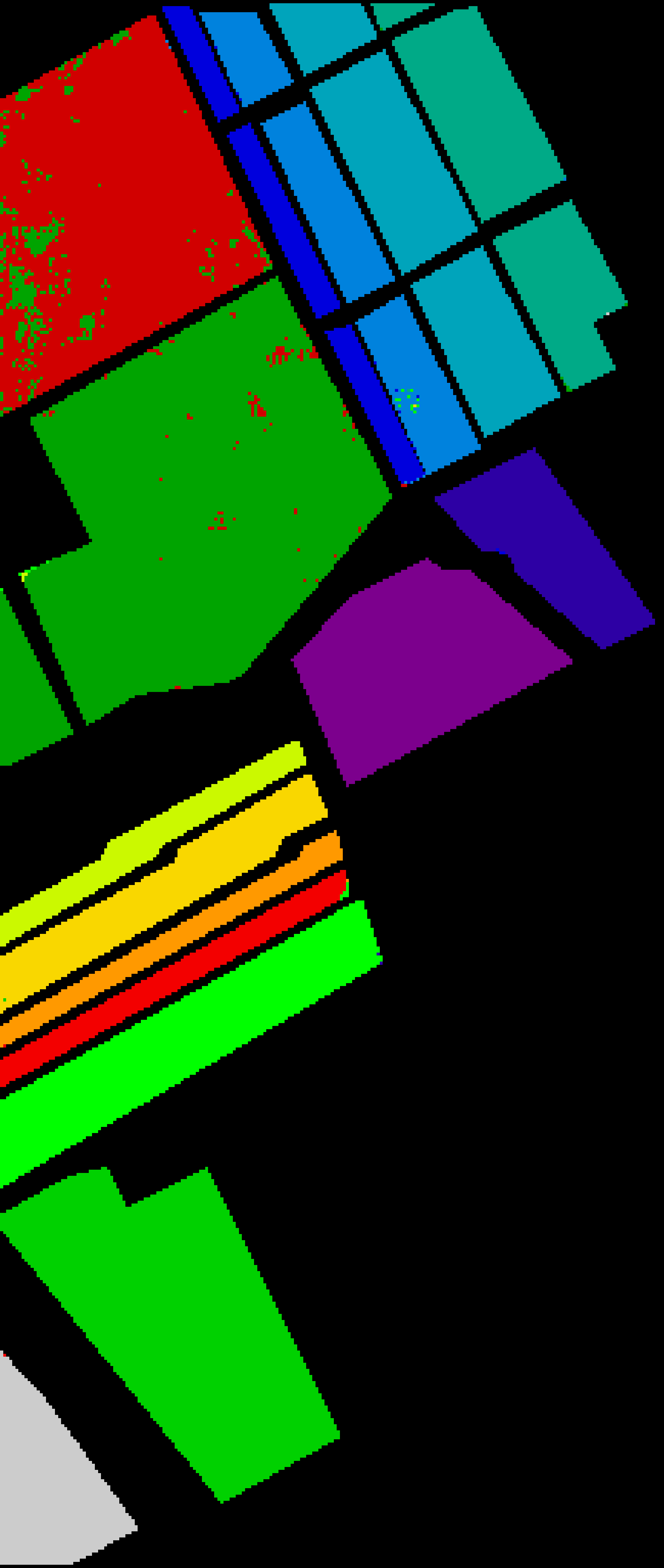}
	\caption*{Test-SA}
    \end{subfigure}
     \begin{subfigure}{0.15\textwidth}
	\includegraphics[width=0.99\textwidth]{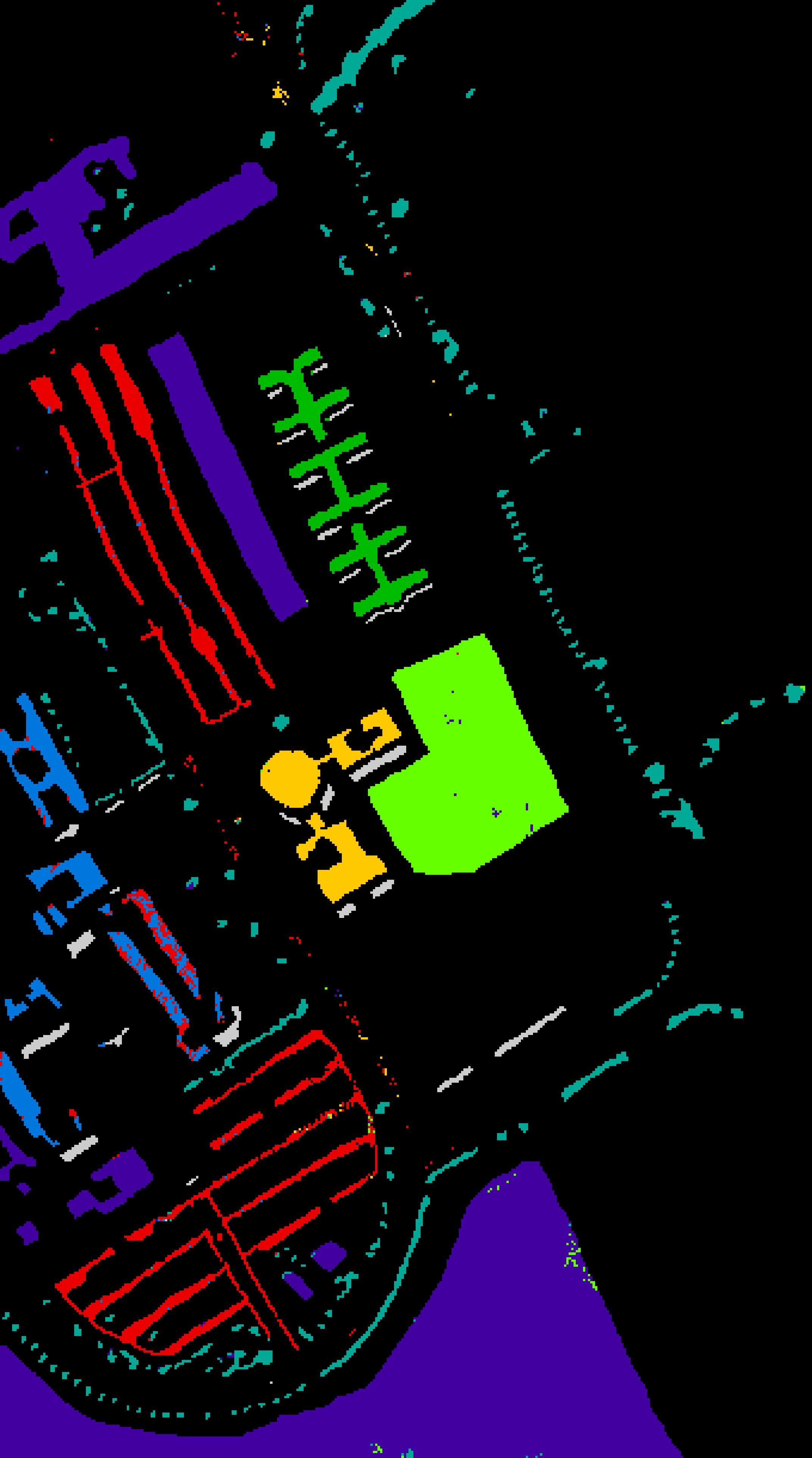}
	\caption*{Test-PU}
    \end{subfigure} 
    \begin{subfigure}{0.15\textwidth}
	\includegraphics[width=0.99\textwidth]{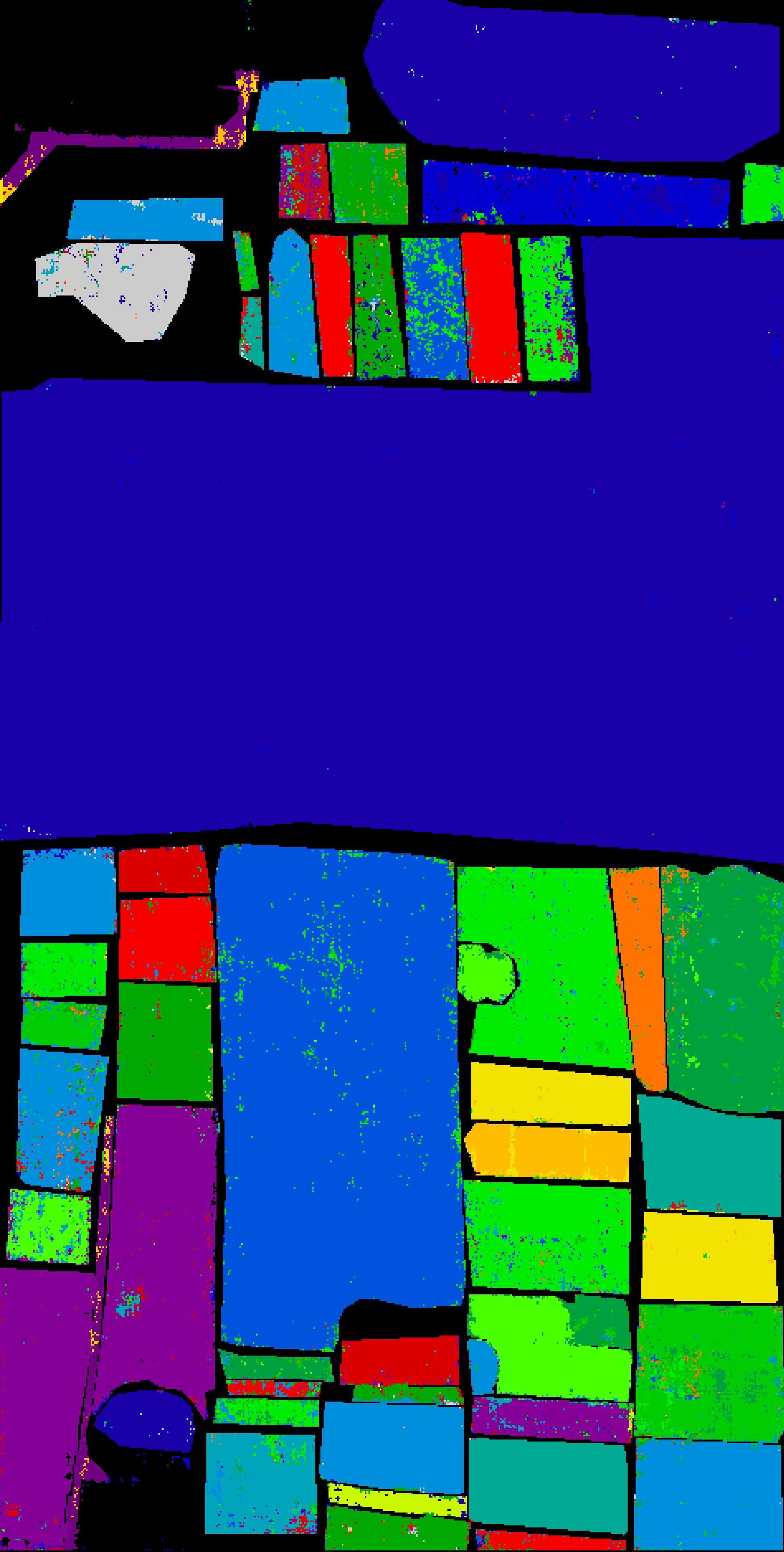}
	\caption*{Test-HH}
    \end{subfigure}
\caption{\textbf{SST trained on UH} was fine-tuned and tested on SA, HC, HH, and PU datasets, respectively.}
\label{AUH}
\end{figure}
%%%%%%%%%%%%%%%%%%%%%%%%%%%%%%%

Table \ref{AUHT} and Figure \ref{AUH} present the performance of the SST actively trained on the UH dataset and fine-tuned with 10\% samples from other target datasets, including SA, PU, HC, and HH. Table \ref{AUHT} outlines key metrics, highlighting the SST model's capability to generalize and adapt to new datasets with minimal fine-tuning. The actively trained SST on UH achieves high baseline performance, with a $\kappa$ of 97.38\% and an OA of 97.58\% on its test data. When fine-tuned with samples from other datasets, the model demonstrates strong transferability, achieving an OA of 98.47\% on SA and 97.75\% on PU. Although the performance is slightly reduced for HC and HH, the OA values remain above 94\%, underscoring the model's robustness. Testing times scale with dataset complexity, with SA and PU requiring less computation compared to HC and HH, as seen in the recorded durations. Figure \ref{AUH} complements the table by visualizing classification results. The maps indicate that the fine-tuned SST effectively captures spatial and spectral features across diverse datasets. Specifically, the SST retains consistent classification quality on target datasets despite using only 10\% of the samples for fine-tuning.

%%%%%%%%%%%%%%%%%%%%%%%%%%%%%%%
\section{Comparison with State-of-the-art Methods}

The following Tables and Figures illustrate a comparative evaluation of various HSI classification methods on the UH, PU, and SA datasets. The methods compared include Attention Graph CNN (AGCN) \cite{10409250}, Wavelet-Based Spatial-Spectral Transformer (WaveFormer) \cite{10399798}, Hybrid Spatial-Spectral Transformer (HybViT) \cite{10636090}, Pyramid-based Spatial-Spectral Transformer (PyFormer) \cite{10681622}, Spatial-Spectral Transformer (SST) \cite{10841983}, Spatial-Spectral Mamba (SSMamba) \cite{10844849}, and Wavelet-Based Spatial-Spectral Mamba (WaveMamba) \cite{10767233}. These state-of-the-art approaches are benchmarked against the proposed method, ATL-SST, under consistent training, validation, and testing configurations, as summarized in the Tables.

%%%%%%%%%%%%%%%%%%%%%%%%%%%%%%%
\begin{table*}[!hbt]
    \centering
    \caption{\textbf{UH dataset:} Comparison of various HSI classification methods (Attention Graph CNN = AGCN, Hybrid Spatial-spectral Transformer = HViT, Spatial-spectral Mamba = SSM, WaveFormer = WF, PyFormer = PF, WaveMamab = WM) highlighting performance metrics.}
    \begin{tabular}{ccc||ccccccc||c} \hline
        \multicolumn{3}{c||}{Samples} & \multicolumn{7}{c||}{\textbf{State-of-the-art Comparative Methods}} & \multirow{2}{*}{\textbf{ATL-SST}} \\ \cline{1-10}
        Tr & Va & Te & AGCN \cite{10409250} & HViT \cite{10636090} & SST \cite{10841983} & WF \cite{10399798} & PF \cite{10681622} & SSM \cite{10844849} & WM \cite{10767233} & \\ \hline 
        43 & 582 & 626 & 95.36 & 93.61 & 94.56 & 94.88 & 85.30 & 91.85 & 93.92 & 98.40 \\
        42 & 585 & 627 & 99.84 & 99.36 & 99.20 & 99.04 & 100 & 96.65 & 98.40 & 99.52 \\
        26 & 323 & 348 & 99.13 & 100 & 99.71 & 98.56 & 95.68 & 93.67 & 97.70 & 100 \\
        44 & 578 & 622 & 94.69 & 97.58 & 97.26 & 97.58 & 98.87 & 96.62 & 94.85 & 97.90 \\
        42 & 579 & 621 & 100 & 99.83 & 100 & 100 & 94.52 & 96.61 & 96.94 & 100 \\
        17 & 146 & 162 & 87.03 & 80.24 & 98.14 & 88.88 & 88.88 & 80.86 & 87.65 & 98.14 \\
        106 & 528 & 634 & 89.74 & 91.95 & 92.27 & 94.47 & 89.27 & 80.44 & 85.48 & 94.47 \\
        81 & 541 & 622 & 87.29 & 90.35 & 92.76 & 96.46 & 87.62 & 83.92 & 90.99 & 95.98 \\
        81 & 545 & 626 & 69.96 & 91.21 & 93.76 & 90.09 & 71.72 & 65.01 & 83.22 & 97.92 \\
        69 & 544 & 614 & 96.09 & 96.74 & 99.51 & 99.02 & 84.85 & 80.29 & 89.73 & 100 \\
        70 & 547 & 618 & 94.66 & 95.95 & 96.76 & 94.82 & 90.93 & 86.08 & 92.88 & 99.02 \\
        68 & 548 & 617 & 81.03 & 99.02 & 98.86 & 97.40 & 80.87 & 92.70 & 92.54 & 97.56 \\
        55 & 180 & 234 & 87.60 & 82.05 & 79.48 & 72.22 & 57.69 & 34.61 & 73.50 & 77.77 \\
        15 & 199 & 214 & 95.32 & 96.26 & 97.19 & 97.19 & 97.19 & 92.52 & 98.59 & 100 \\
        28 & 302 & 330 & 94.54 & 95.45 & 100 & 100 & 100 & 81.51 & 95.75 & 97.87 \\ \hline 
        \multicolumn{3}{c||}{Train (s)} & 101.61 & 50.85 & 46.74 & 47.73 & 397.21 & 67.21 & 563.52 & 47.39 \\ 
        \multicolumn{3}{c||}{$\kappa$} & 90.64 & 94.61 & 96.01 & 95.39 & 87.62 & 84.26 & 91.19 & \textbf{97.38} \\
        \multicolumn{3}{c||}{OA} & 91.35 & 95.02 & 96.31 & 95.74 & 88.55 & 85.46 & 91.85 & \textbf{97.57} \\
        \multicolumn{3}{c||}{AA} & 91.49 & 93.97 & 95.96 & 94.71 & 88.23 & 83.56 & 91.48 & \textbf{96.97}\\ \hline 
    \end{tabular}
    \label{UHC}
\end{table*}
%%%%%%%%%%%%%%%%%%%%%%%%%%%%%%%
\begin{figure*}[!hbt]
    \centering
    \begin{subfigure}{0.49\textwidth}
	\includegraphics[width=0.99\textwidth]{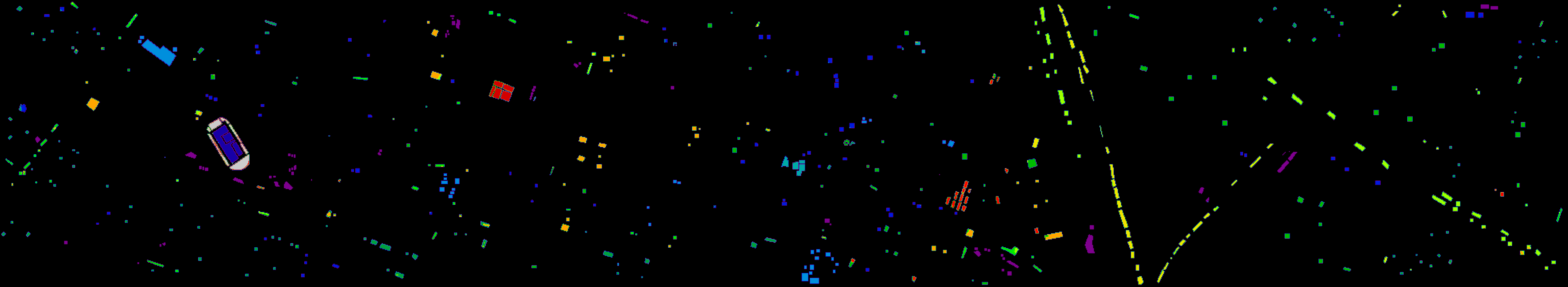}
	\caption*{AGCN \cite{10409250}} 
    \end{subfigure}
    \begin{subfigure}{0.49\textwidth}
	\includegraphics[width=0.99\textwidth]{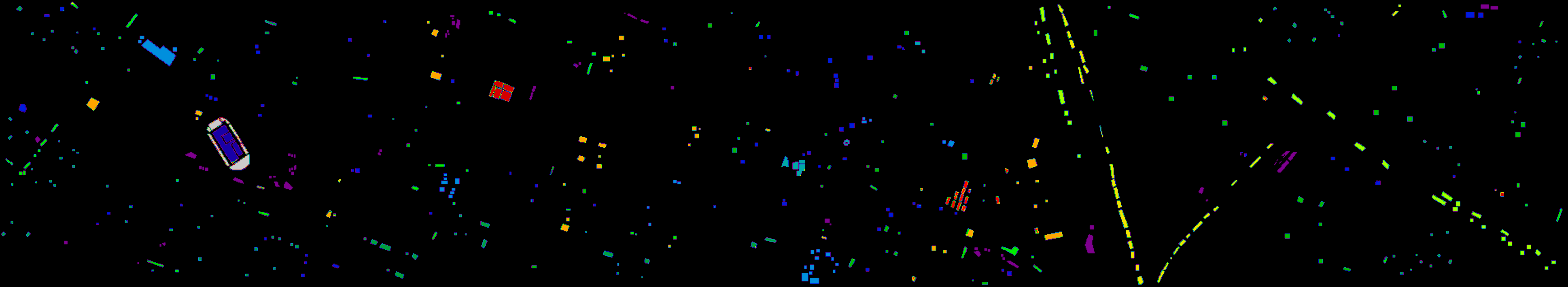}
	\caption*{SST \cite{10841983}}
    \end{subfigure}
    \begin{subfigure}{0.49\textwidth}
	\includegraphics[width=0.99\textwidth]{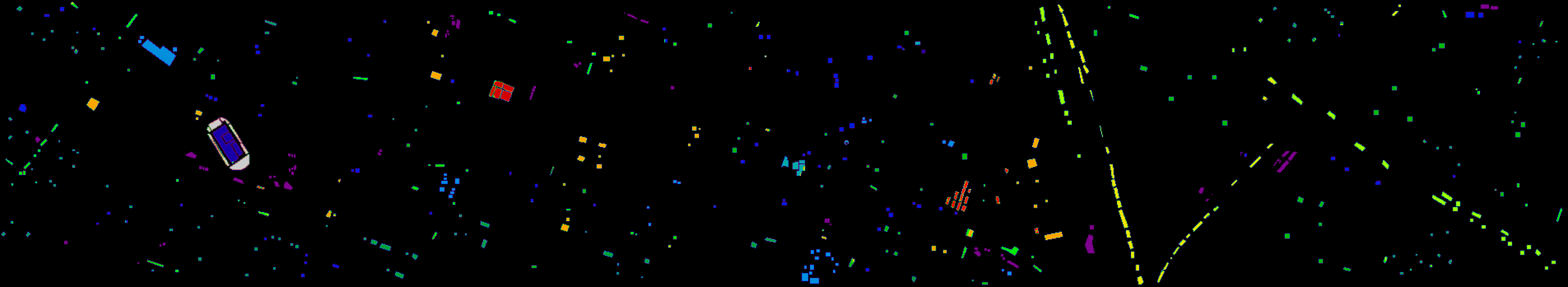}
	\caption*{HViT \cite{10636090}}
    \end{subfigure}
     \begin{subfigure}{0.49\textwidth}
	\includegraphics[width=0.99\textwidth]{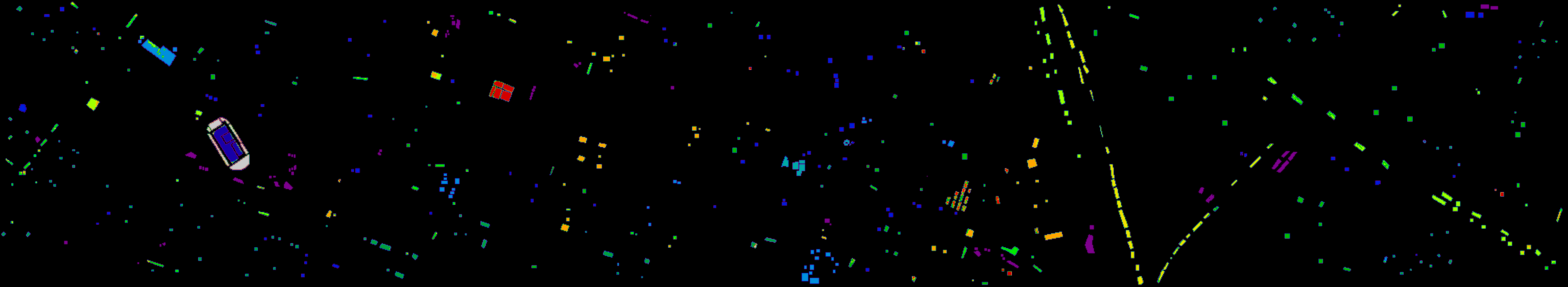}
	\caption*{PF \cite{10681622}}
    \end{subfigure} 
    \begin{subfigure}{0.49\textwidth}
	\includegraphics[width=0.99\textwidth]{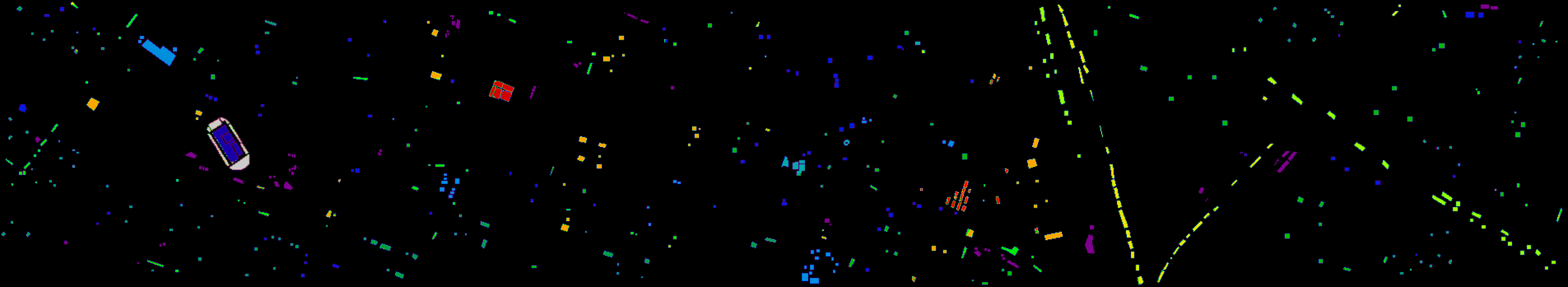}
	\caption*{WF \cite{10399798}}
    \end{subfigure}
    \begin{subfigure}{0.49\textwidth}
	\includegraphics[width=0.99\textwidth]{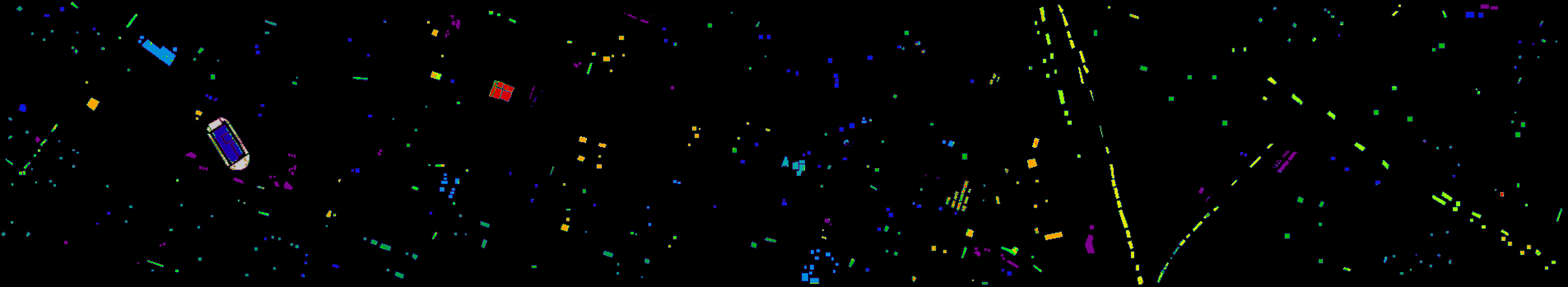}
	\caption*{SSM \cite{10844849}}
    \end{subfigure}
    \begin{subfigure}{0.49\textwidth}
	\includegraphics[width=0.99\textwidth]{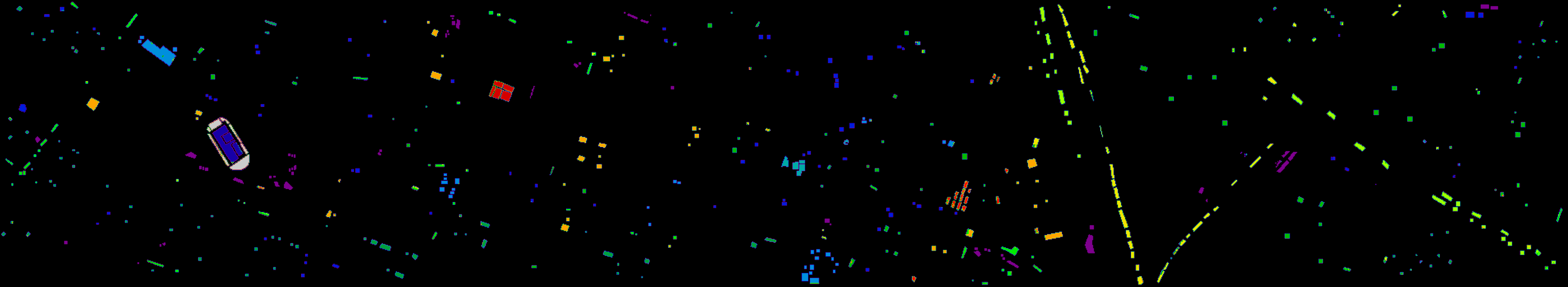}
	\caption*{WM \cite{10767233}}
    \end{subfigure}
    \begin{subfigure}{0.49\textwidth}
	\includegraphics[width=0.99\textwidth]{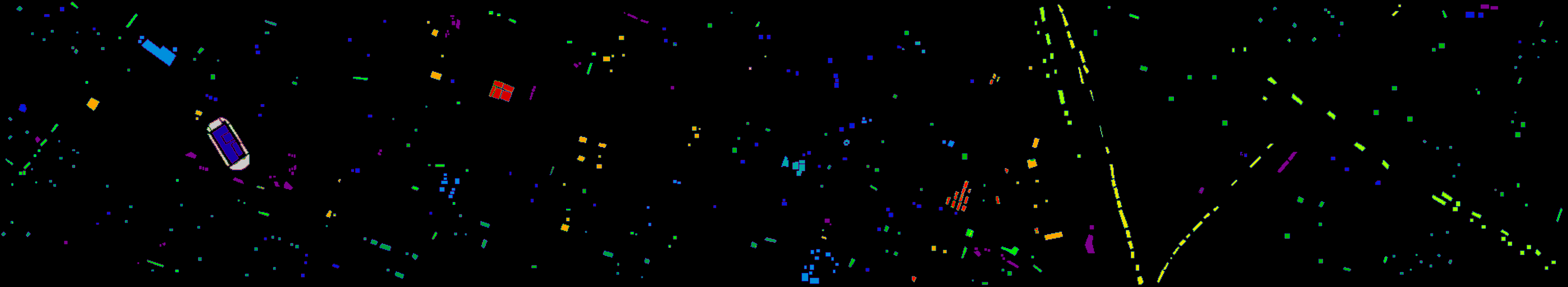}
	\caption*{Proposed}
    \end{subfigure}
\caption{\textbf{UH dataset:} Ground truth maps for comparative methods alongside the proposed ATL-SST.}
\label{UHCG}
\end{figure*}
%%%%%%%%%%%%%%%%%%%%%%%%%%%%%%%

%%%%%%%%%%%%%%%%%%%%%%%%%%%%%%%
\begin{table*}[!hbt]
    \centering
    \caption{\textbf{PU dataset:} Comparison of various HSI classification methods highlighting performance metrics.}
    \begin{tabular}{ccc||ccccccc||c} \hline
        \multicolumn{3}{c||}{Samples} & \multicolumn{7}{c||}{\textbf{State-of-the-art Comparative Methods}} & \multirow{2}{*}{\textbf{ATL-SST}} \\ \cline{1-10}
        Tr & Va & Te & AGCN \cite{10409250} & HViT \cite{10636090} & SST \cite{10841983} & WF \cite{10399798} & PF \cite{10681622} & SSM \cite{10844849} & WM \cite{10767233} & \\ \hline 
        422 & 2893 & 3316 & 99.45 & 97.13 & 97.40 & 97.55 & 96.86 & 88.75 & 97.64 & 98.79 \\
        535 & 8790 & 9324 & 98.01 & 99.14 & 99.02 & 99.53 & 98.93 & 94.74 & 99.53 & 99.90 \\
        218 & 832 & 1049 & 69.01 & 84.65 & 84.74 & 82.93 & 93.51 & 74.73 & 75.97 & 91.32 \\
        161 & 1371 & 1532 & 97.51 & 96.27 & 96.01 & 95.16 & 97.32 & 92.23 & 94.38 & 99.54 \\
        15 & 657 & 673 & 97.77 & 100 & 100 & 100 & 100 & 98.95 & 98.21 & 100 \\
        311 & 2204 & 2514 & 99.20 & 97.73 & 97.13 & 96.26 & 94.35 & 93.35 & 98.60 & 99.64 \\
        114 & 551 & 665 & 87.96 & 88.12 & 92.93 & 91.27 & 75.03 & 82.25 & 92.33 & 99.54 \\
        405 & 1436 & 1841 & 93.91 & 89.57 & 90.05 & 92.28 & 63.06 & 81.96 & 88.48 & 95.87 \\
        64 & 409 & 474 & 97.04 & 93.45 & 96.83 & 94.93 & 89.45 & 89.45 & 90.50 & 98.94 \\ \hline 
        \multicolumn{3}{c||}{Train (s)} & 72.49 & 117.58 & 114.56 & 112.68 & 1181.18 & 154.76 & 1382.70 & 124.78 \\
        \multicolumn{3}{c||}{$\kappa$} & 95.00 & 95.33 & 95.56 & 95.68 & 91.62 & 88.16 & 94.94 & \textbf{98.51} \\
        \multicolumn{3}{c||}{OA} & 96.22 & 96.48 & 96.65 & 96.75 & 93.68 & 91.01 & 96.19 & \textbf{98.87} \\
        \multicolumn{3}{c||}{AA} & 93.32 & 94.01 & 94.90 & 94.44 & 89.83 & 88.49 & 92.85 & \textbf{98.17} \\ \hline 
    \end{tabular}
    \label{PUC}
\end{table*}
%%%%%%%%%%%%%%%%%%%%%%%%%%%%%%%
\begin{figure*}[!hbt]
    \centering
    \begin{subfigure}{0.11\textwidth}
	\includegraphics[width=0.99\textwidth]{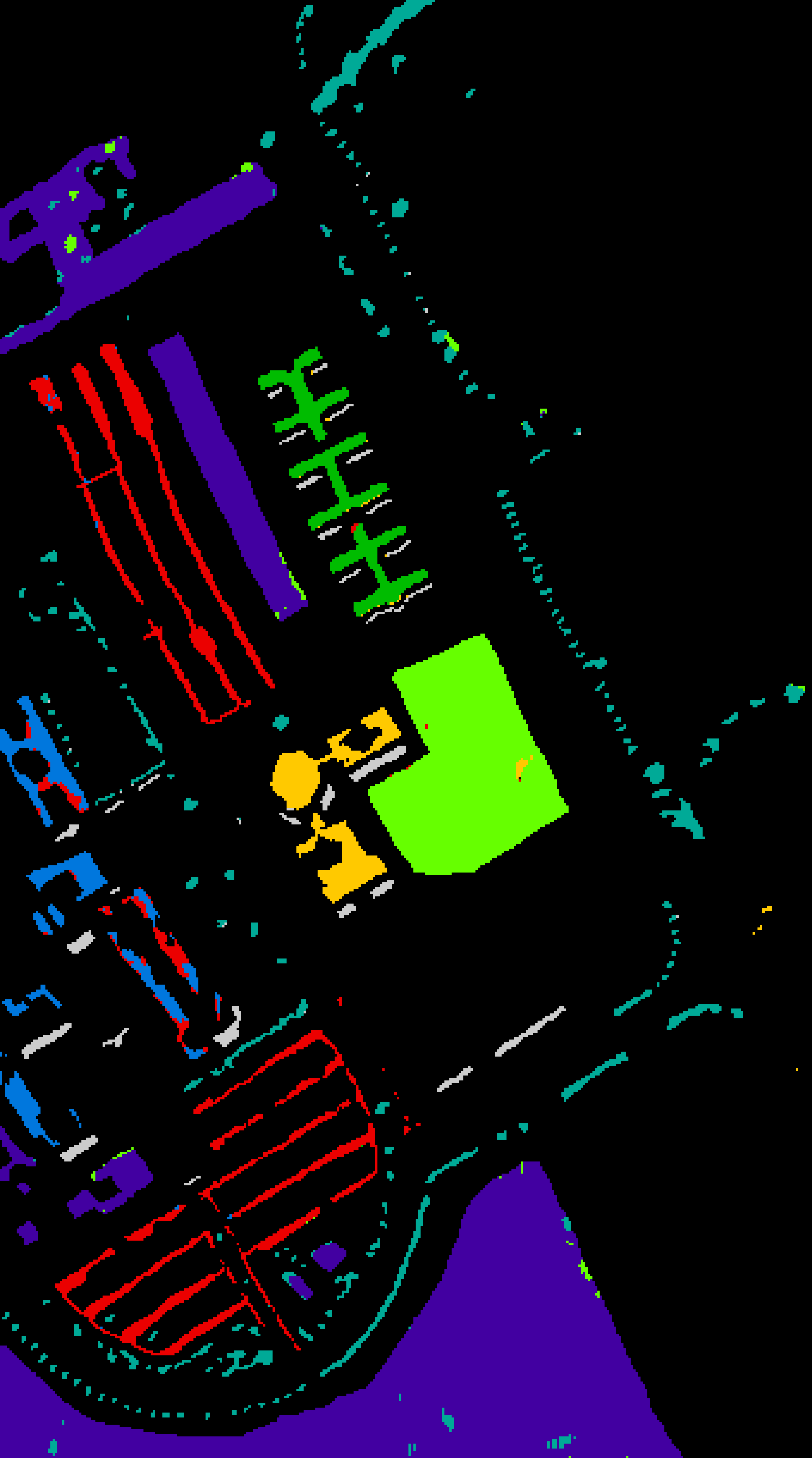}
	\caption*{AGCN \cite{10409250}}
    \end{subfigure}
    \begin{subfigure}{0.11\textwidth}
	\includegraphics[width=0.99\textwidth]{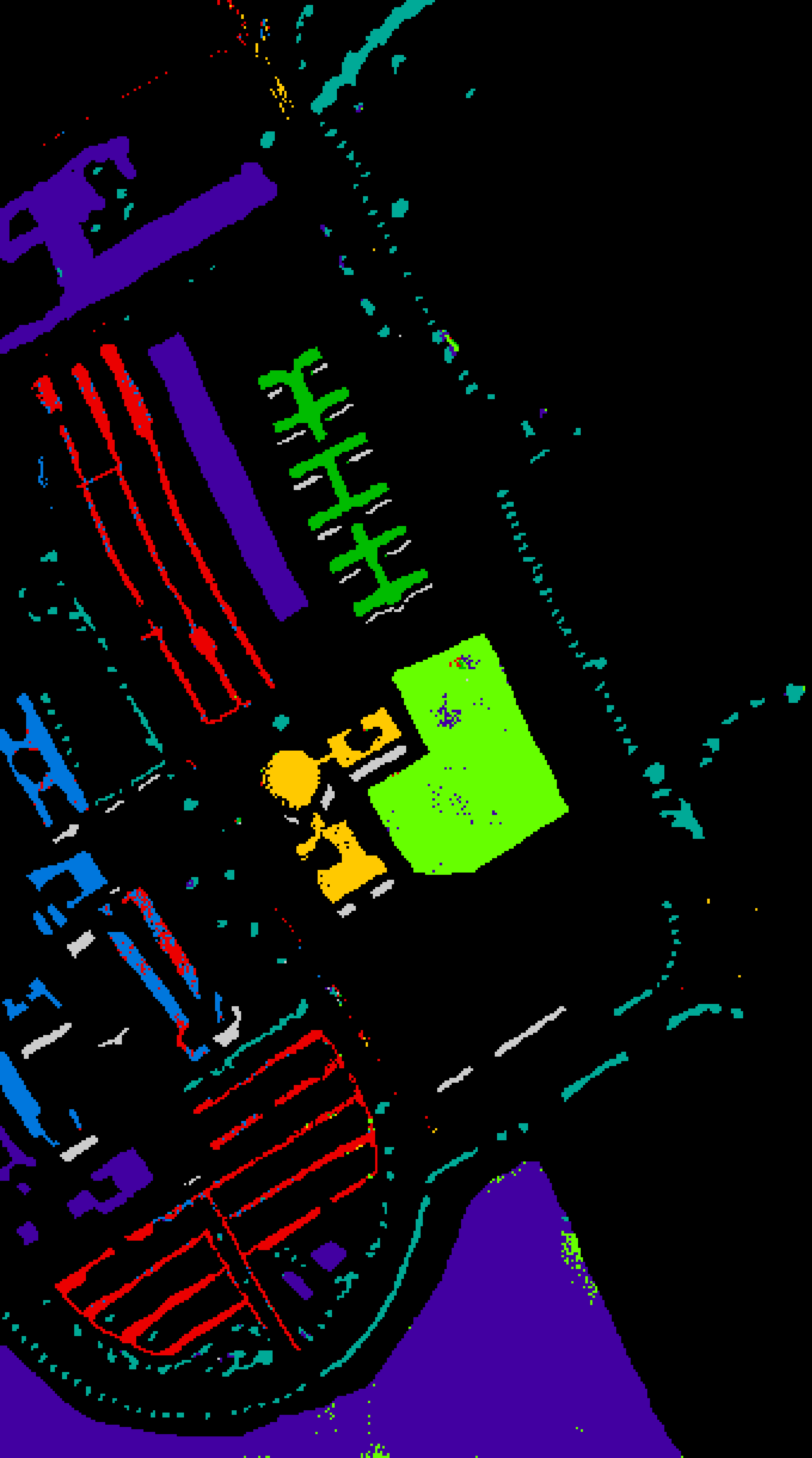}
	\caption*{SST \cite{10841983}}
    \end{subfigure}
    \begin{subfigure}{0.11\textwidth}
	\includegraphics[width=0.99\textwidth]{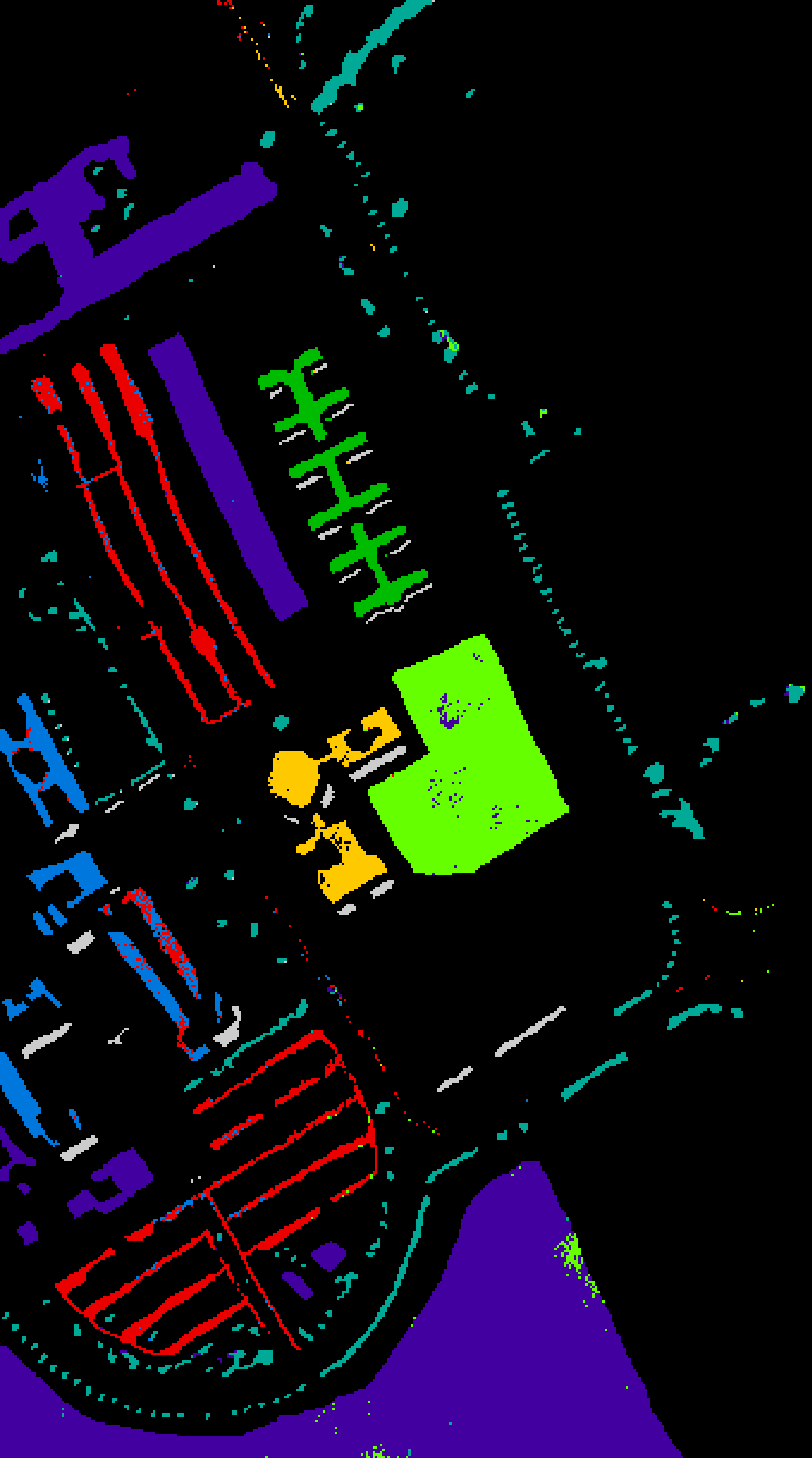}
	\caption*{HViT \cite{10636090}}
    \end{subfigure}
     \begin{subfigure}{0.11\textwidth}
	\includegraphics[width=0.99\textwidth]{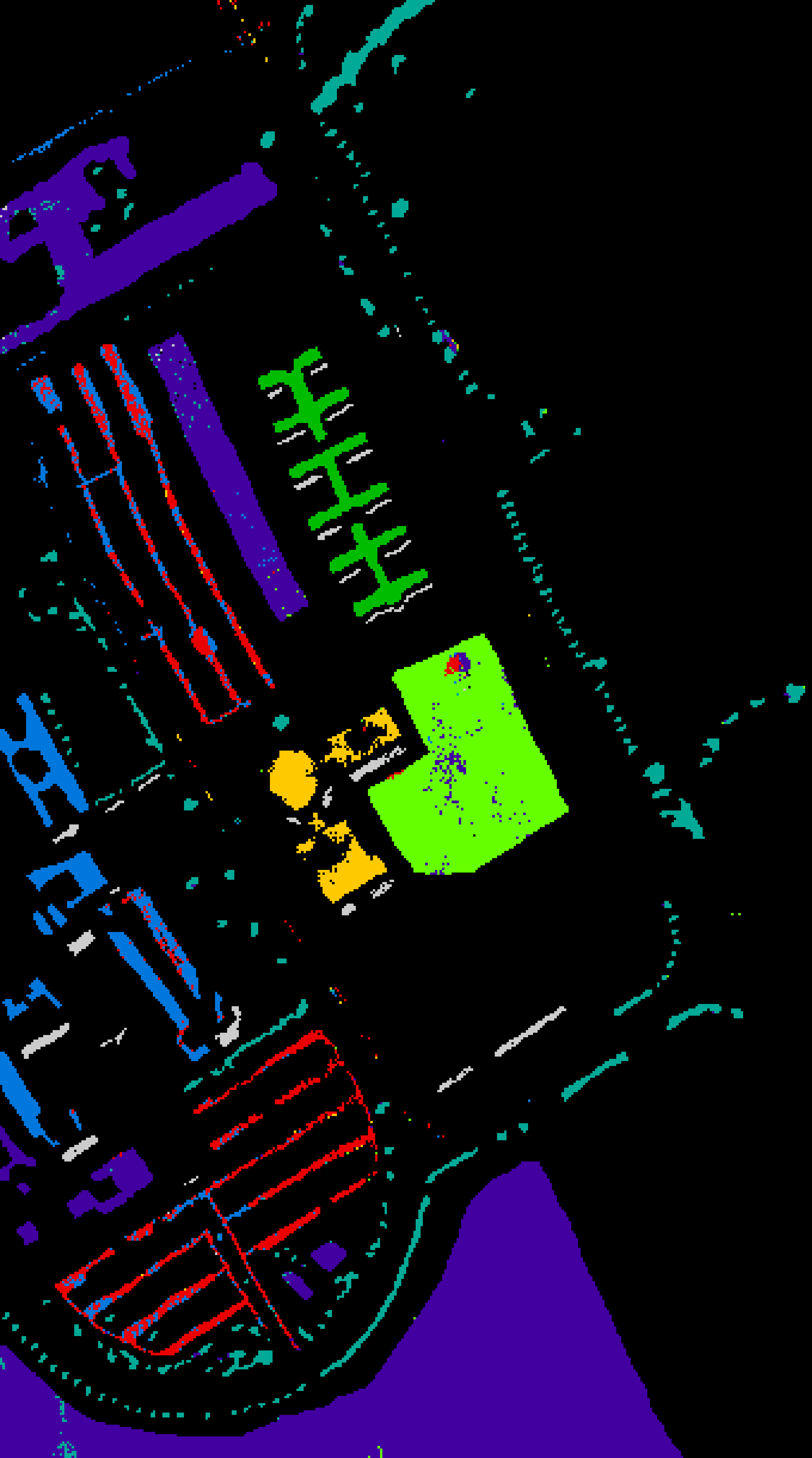}
	\caption*{PF \cite{10681622}}
    \end{subfigure} 
    \begin{subfigure}{0.11\textwidth}
	\includegraphics[width=0.99\textwidth]{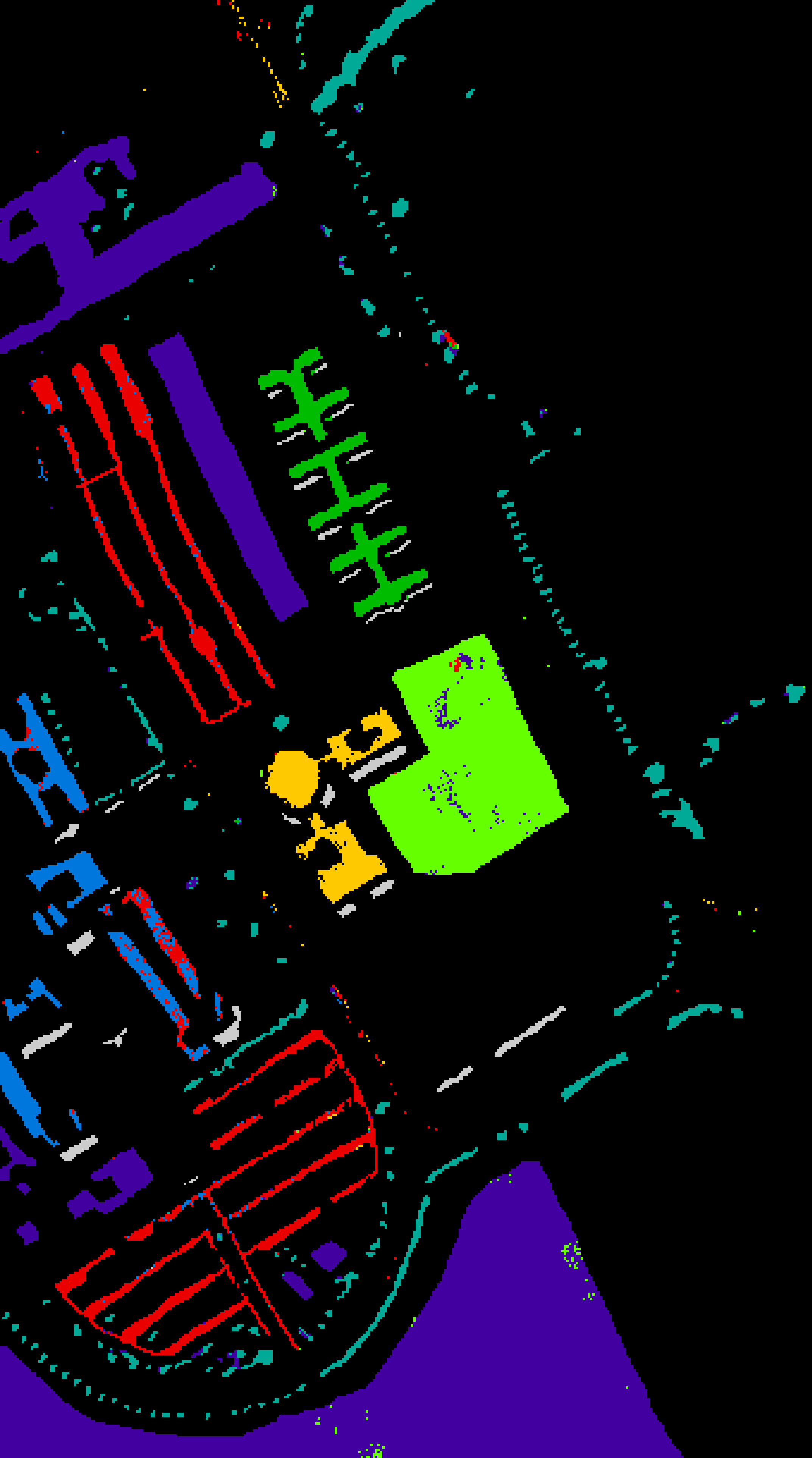}
	\caption*{WF \cite{10399798}}
    \end{subfigure}
    \begin{subfigure}{0.11\textwidth}
	\includegraphics[width=0.99\textwidth]{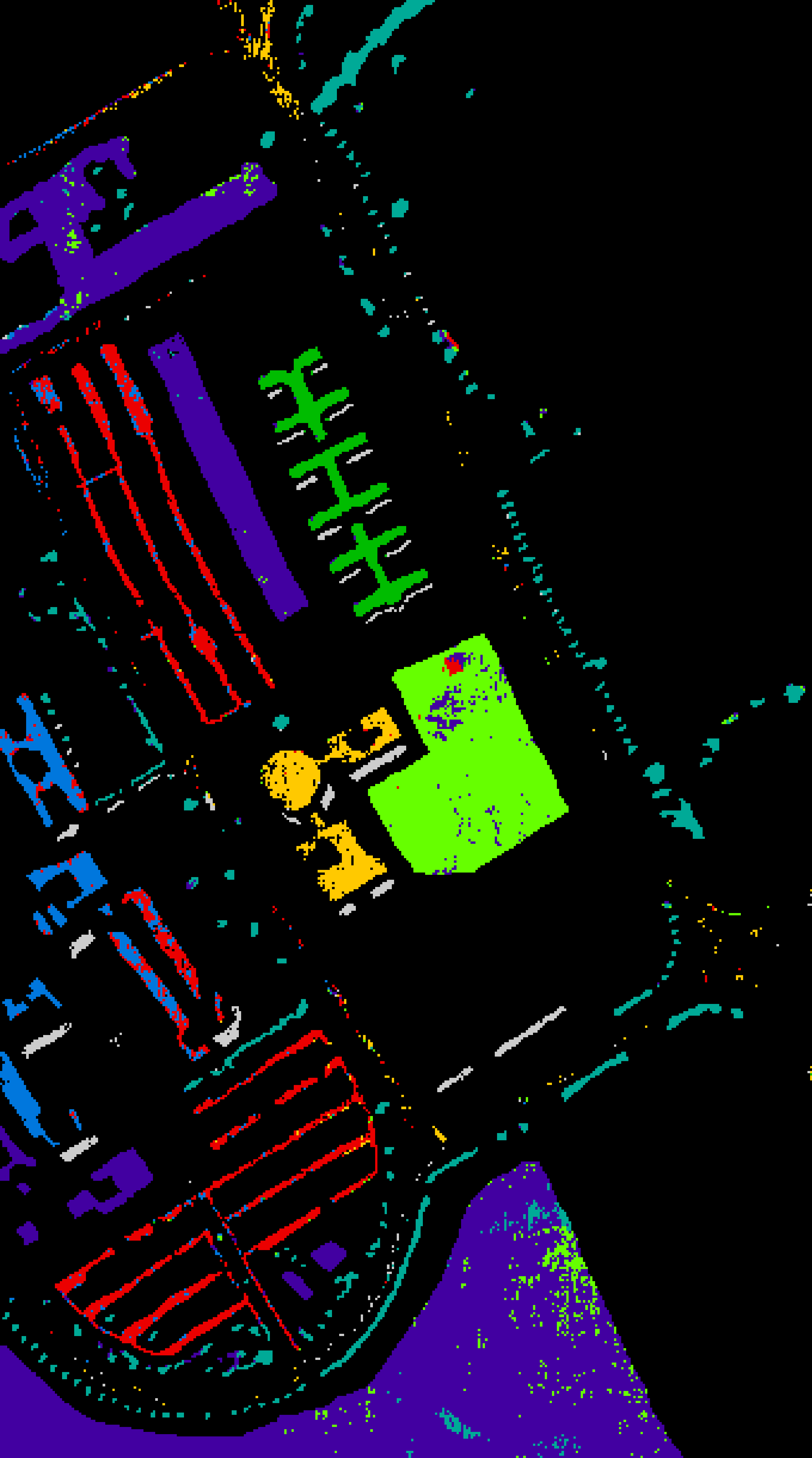}
	\caption*{SSM \cite{10844849}}
    \end{subfigure}
    \begin{subfigure}{0.11\textwidth}
	\includegraphics[width=0.99\textwidth]{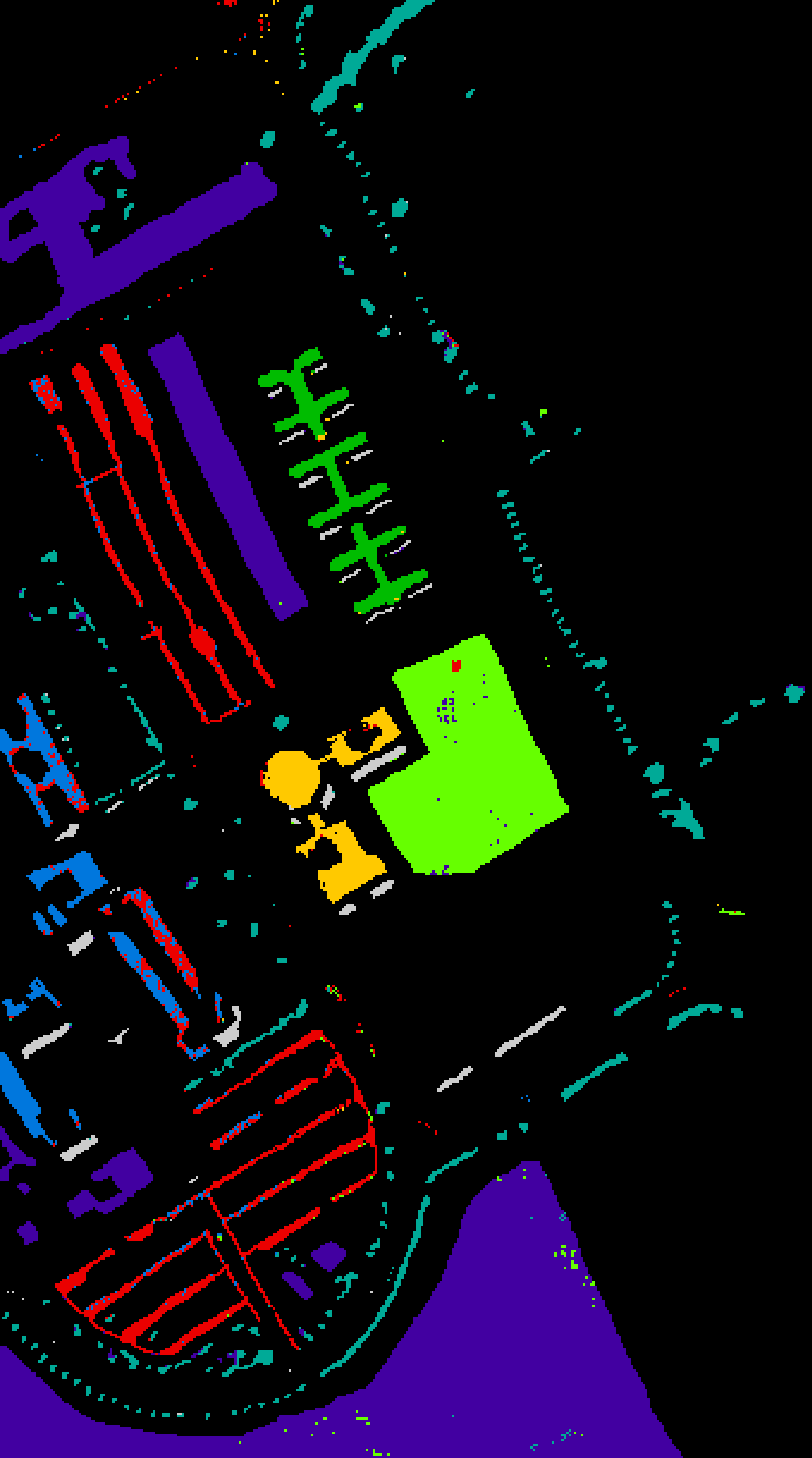}
	\caption*{WM \cite{10767233}}
    \end{subfigure}
    \begin{subfigure}{0.11\textwidth}
	\includegraphics[width=0.99\textwidth]{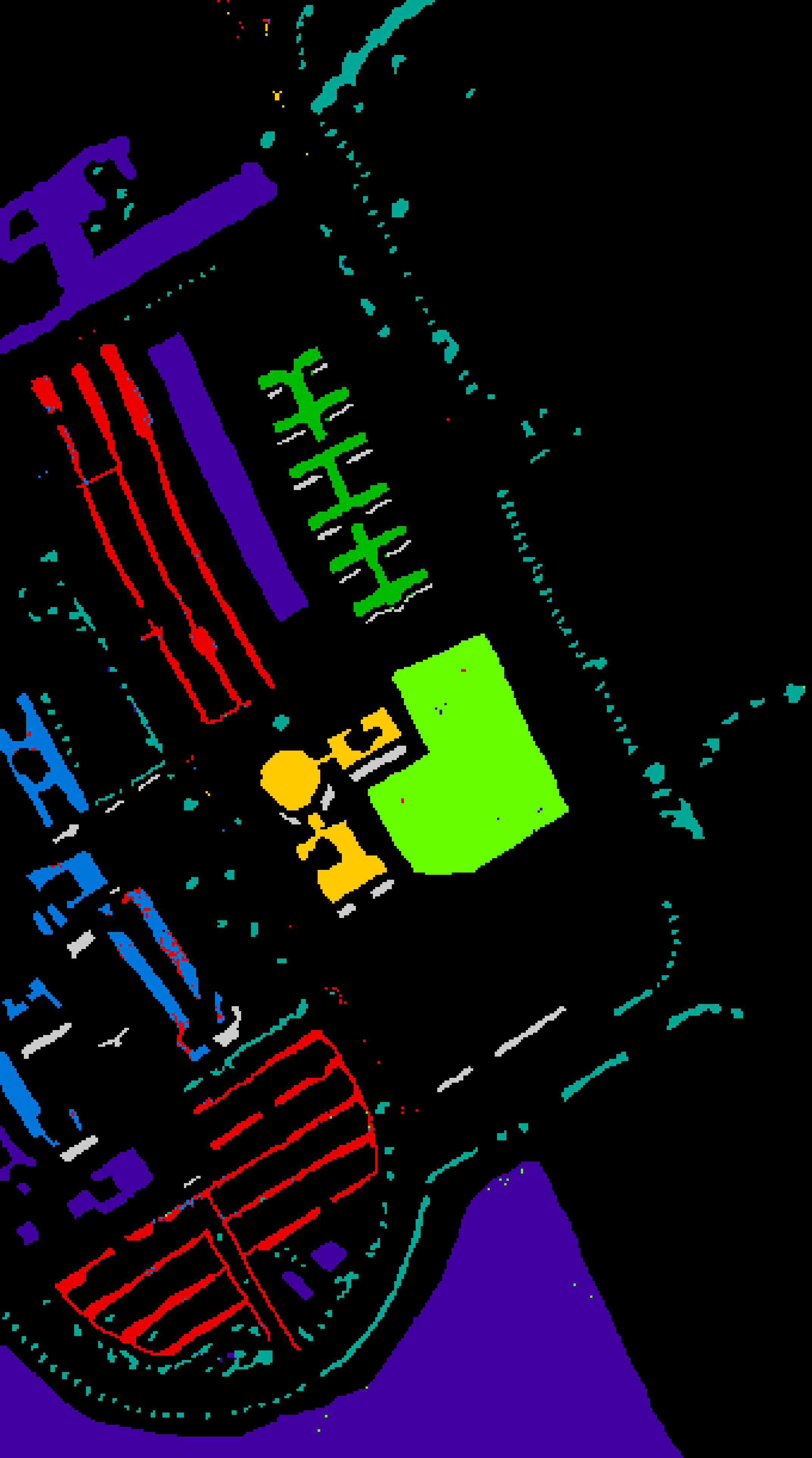}
	\caption*{Proposed}
    \end{subfigure}
\caption{\textbf{PU dataset:} Ground truth maps for comparative methods alongside the proposed ATL-SST.}
\label{PUCG}
\end{figure*}
%%%%%%%%%%%%%%%%%%%%%%%%%%%%%%%

%%%%%%%%%%%%%%%%%%%%%%%%%%%%%%%
\begin{table*}[!hbt]
    \centering
    \caption{\textbf{SA} Comparison of various HSI classification methods highlighting performance metrics.}
    \begin{tabular}{ccc||ccccccc||c} \hline
        \multicolumn{3}{c||}{Samples} & \multicolumn{7}{c||}{\textbf{State-of-the-art Comparative Methods}} & \multirow{2}{*}{\textbf{ATL-SST}} \\ \cline{1-10}
        Tr & Va & Te & AGCN \cite{10409250} & HViT \cite{10636090} & SST \cite{10841983} & WF \cite{10399798} & PF \cite{10681622} & SSM \cite{10844849} & WM \cite{10767233} & \\ \hline 
        41 & 964 & 1004 & 99.50 & 99.90 & 99.00 & 98.50 & 100 & 99.40 & 100 & 100 \\
        93 & 1770 & 1863 & 100 & 100 & 100 & 100 & 100 & 100 & 100 & 100 \\
        80 & 908 & 988 & 100 & 99.39 & 99.59 & 99.49 & 100 & 97.36 & 98.68 & 100 \\
        62 & 635 & 697 & 96.41 & 99.85 & 99.71 & 100 & 100 & 98.4218 & 99.13 & 100 \\
        118 & 1221 & 1339 & 86.03 & 99.85 & 99.70 & 99.17 & 94.69 & 98.65 & 98.20 & 99.40 \\
        77 & 1902 & 1980 & 100 & 100 & 100 & 100 & 100 & 99.84 & 99.84 & 100 \\
        64 & 1725 & 1790 & 99.88 & 99.88 & 100 & 100 & 100 & 98.82 & 99.21 & 100 \\
        1004 & 4631 & 5636 & 86.12 & 93.98 & 94.96 & 95.28 & 88.25 & 87.63 & 94.25 & 98.86 \\
        109 & 2992 & 3102 & 100 & 100 & 100 & 100 & 99.96 & 99.83 & 100 & 99.96 \\
        94 & 1545 & 1639 & 98.16 & 99.51 & 99.14 & 99.75 & 93.89 & 97.37 & 97.62 & 99.93 \\
        58 & 476 & 534 & 91.94 & 100 & 98.87 & 99.81 & 88.95 & 96.81 & 98.31 & 100 \\
        56 & 908 & 963 & 97.92 & 100 & 99.58 & 99.89 & 100 & 98.65 & 99.89 & 100 \\
        54 & 404 & 458 & 96.72 & 100 & 100 & 100 & 96.28 & 99.12 & 100 & 100 \\
        40 & 495 & 535 & 99.62 & 99.62 & 100 & 99.06 & 46.16 & 94.20 & 98.13 & 99.81 \\
        842 & 2792 & 3634 & 92.90 & 91.24 & 90.72 & 90.03 & 92.43 & 76.63 & 91.16 & 98.23 \\
        50 & 854 & 903 & 99.44 & 100 & 99.66 & 100 & 100 & 99.66 & 99.55 & 100 \\ \hline 
        \multicolumn{3}{c||}{Train (s)} & 91.07 & 147.54 & 139.82 & 144.16 & 1469.81 & 194.06 & 1734.50 & 157.85 \\
        \multicolumn{3}{c||}{$\kappa$} & 94.35 & 97.20 & 97.25 & 97.24 & 93.94 & 92.80 & 96.84 & \textbf{99.42} \\
        \multicolumn{3}{c||}{OA} & 94.92 & 97.49 & 97.53 & 97.52 & 94.55 & 93.54 & 97.16 & \textbf{99.48} \\
        \multicolumn{3}{c||}{AA} & 96.54 & 98.95 & 98.81 & 98.81 & 93.79 & 96.40 & 98.37 & \textbf{99.76} \\ \hline 
    \end{tabular}
    \label{SAC}
\end{table*}
%%%%%%%%%%%%%%%%%%%%%%%%%%%%%%%
\begin{figure*}[!hbt]
    \centering
    \begin{subfigure}{0.11\textwidth}
	\includegraphics[width=0.99\textwidth]{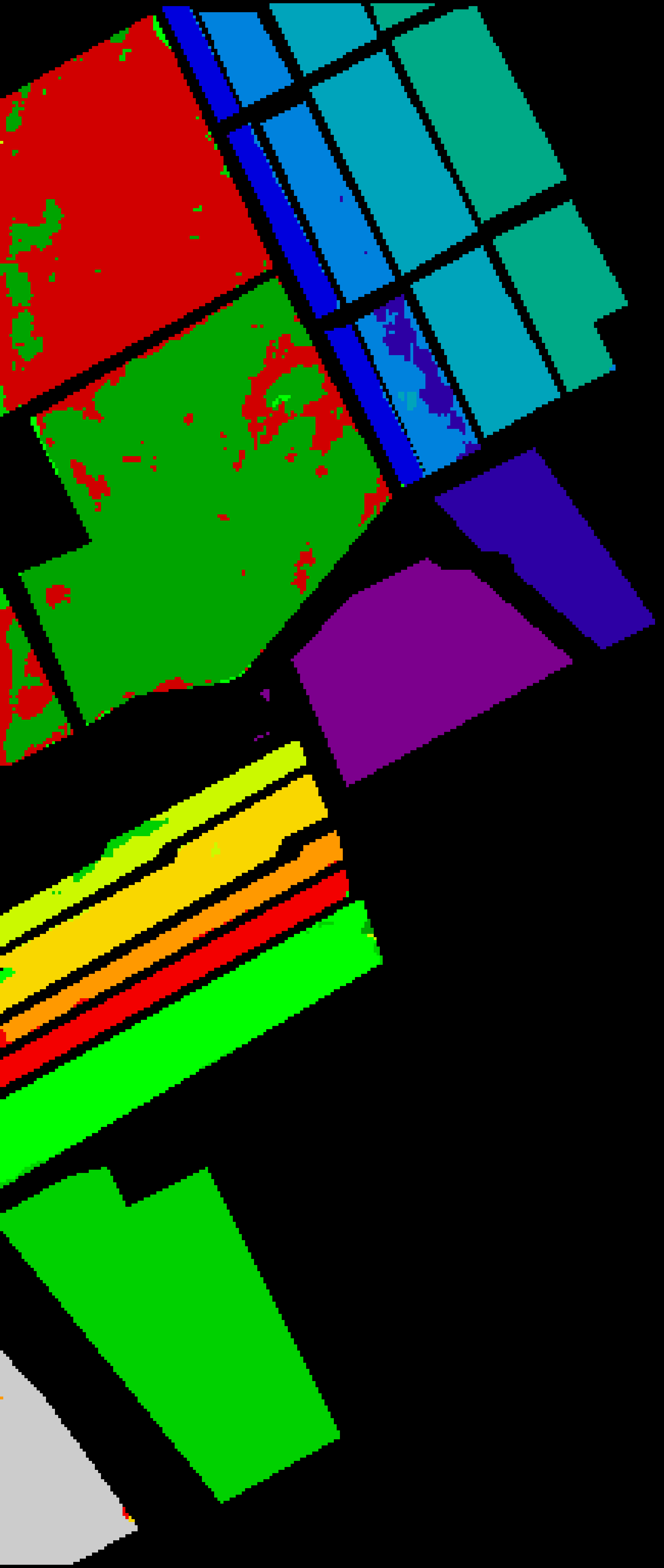}
	\caption*{AGCN \cite{10409250}}
    \end{subfigure}
    \begin{subfigure}{0.11\textwidth}
	\includegraphics[width=0.99\textwidth]{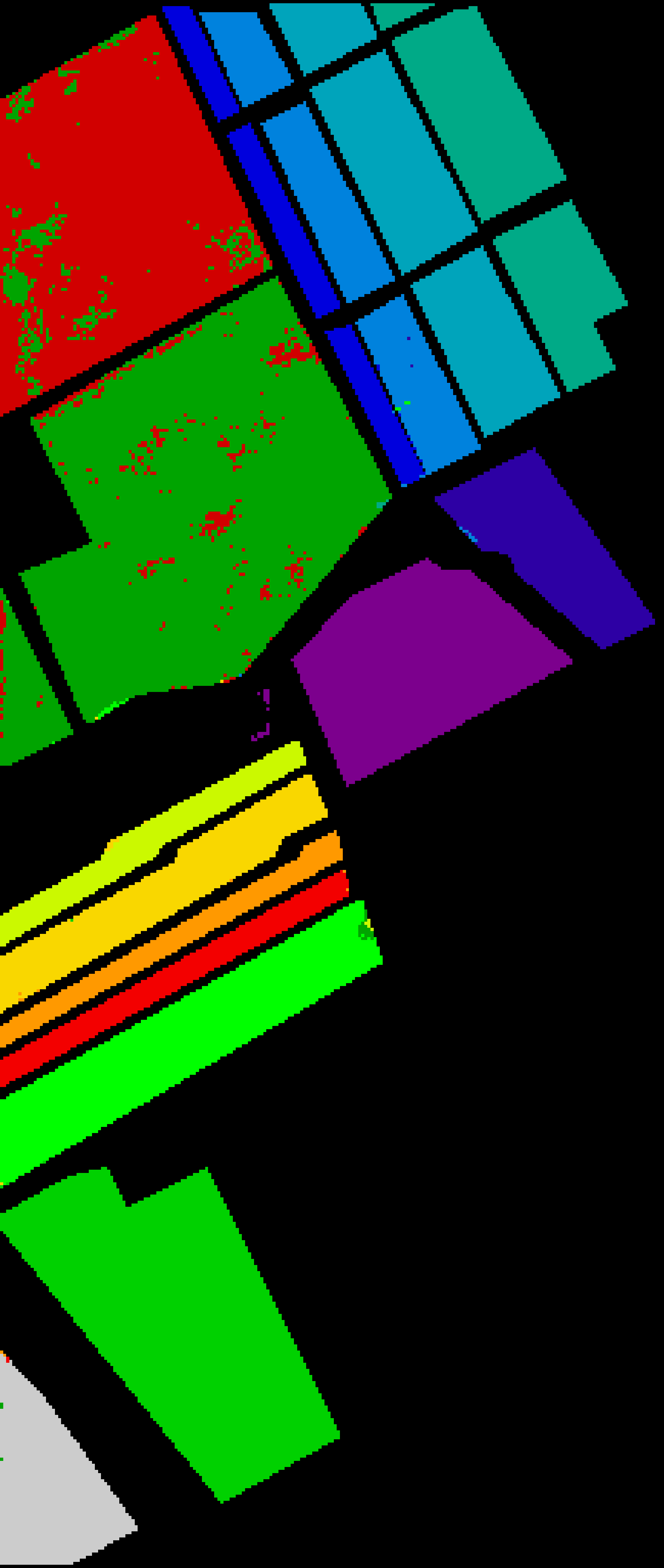}
	\caption*{SST \cite{10841983}}
    \end{subfigure}
    \begin{subfigure}{0.11\textwidth}
	\includegraphics[width=0.99\textwidth]{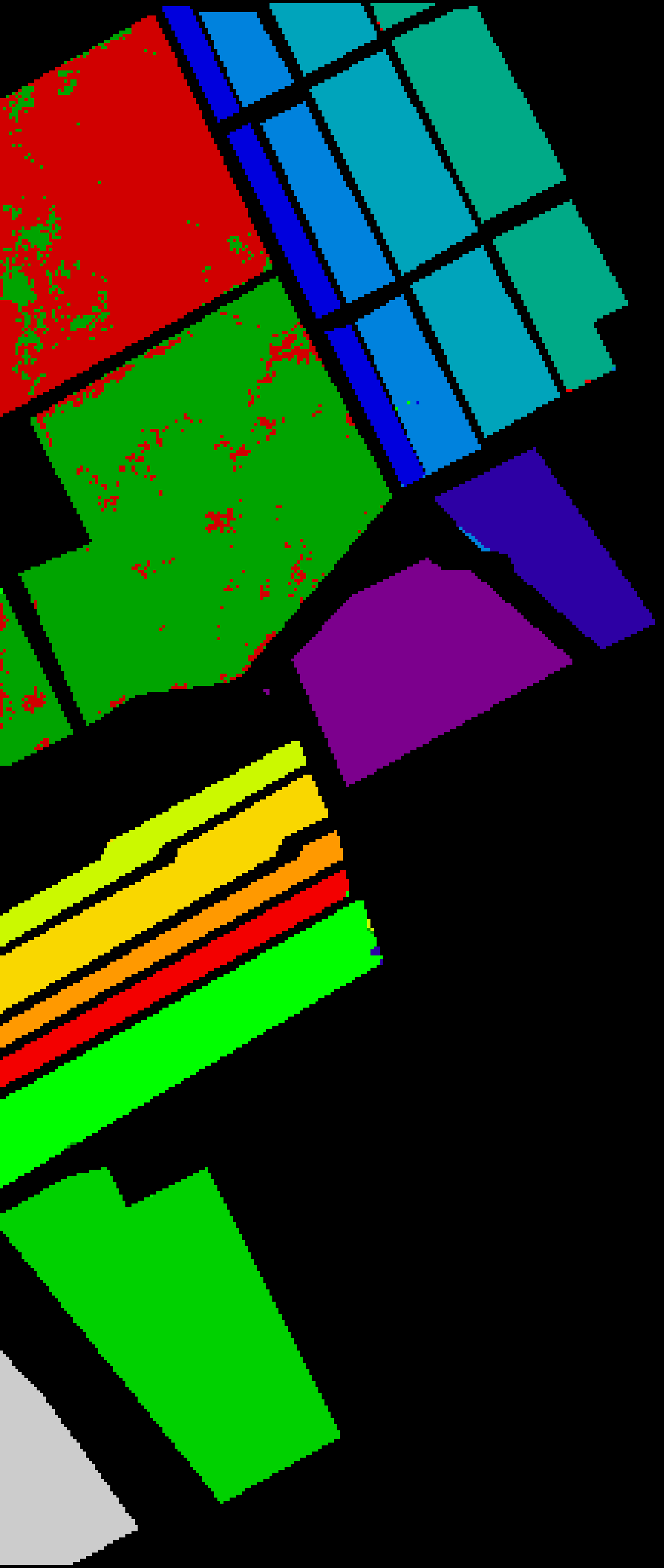}
	\caption*{HViT \cite{10636090}}
    \end{subfigure}
     \begin{subfigure}{0.11\textwidth}
	\includegraphics[width=0.99\textwidth]{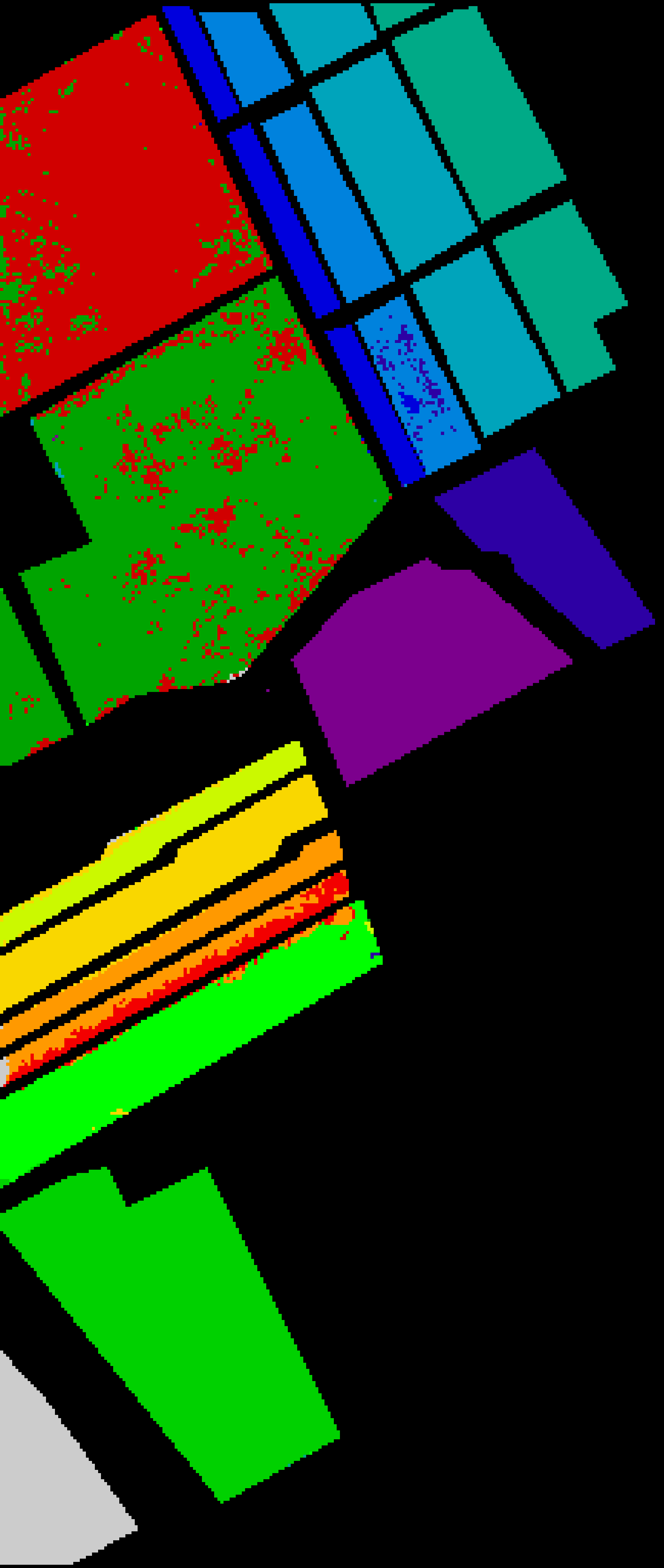}
	\caption*{PF \cite{10681622}}
    \end{subfigure} 
    \begin{subfigure}{0.11\textwidth}
	\includegraphics[width=0.99\textwidth]{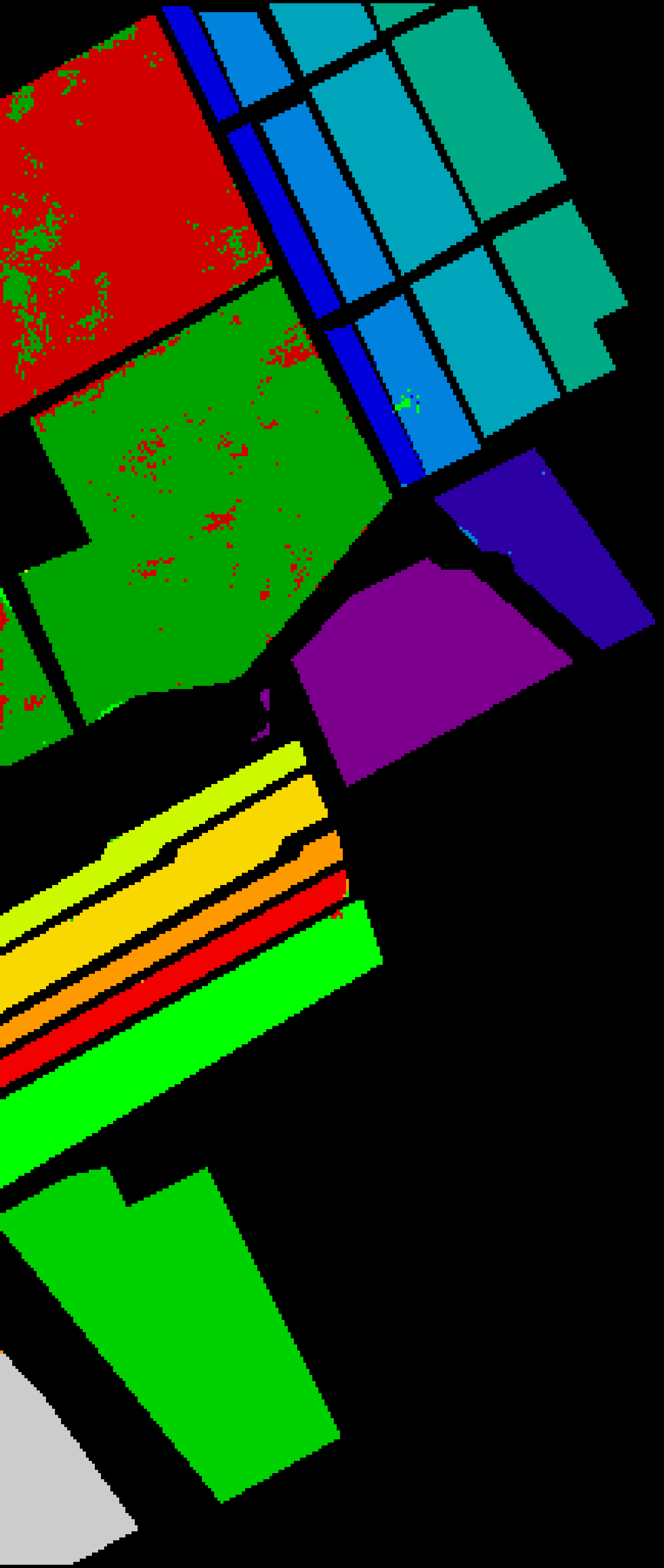}
	\caption*{WF \cite{10399798}}
    \end{subfigure}
    \begin{subfigure}{0.11\textwidth}
	\includegraphics[width=0.99\textwidth]{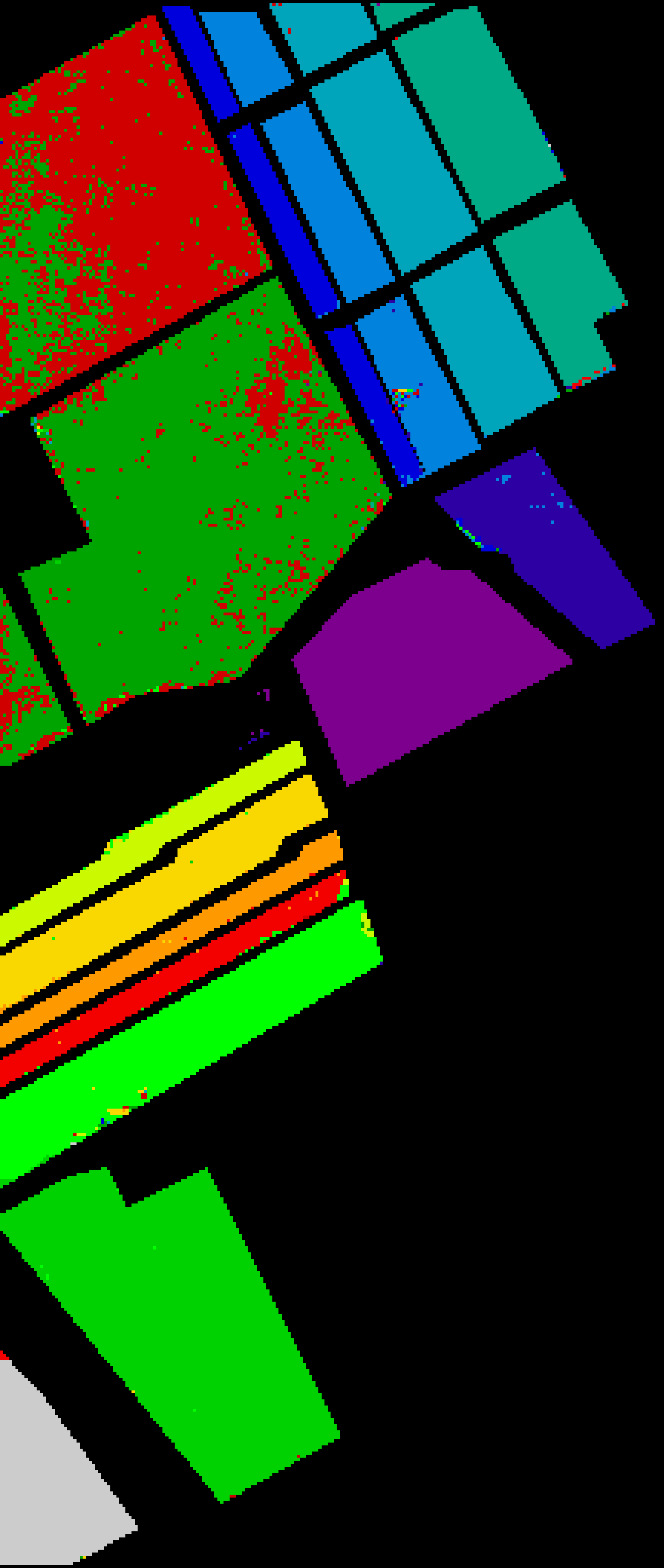}
	\caption*{SSM \cite{10844849}}
    \end{subfigure}
    \begin{subfigure}{0.11\textwidth}
	\includegraphics[width=0.99\textwidth]{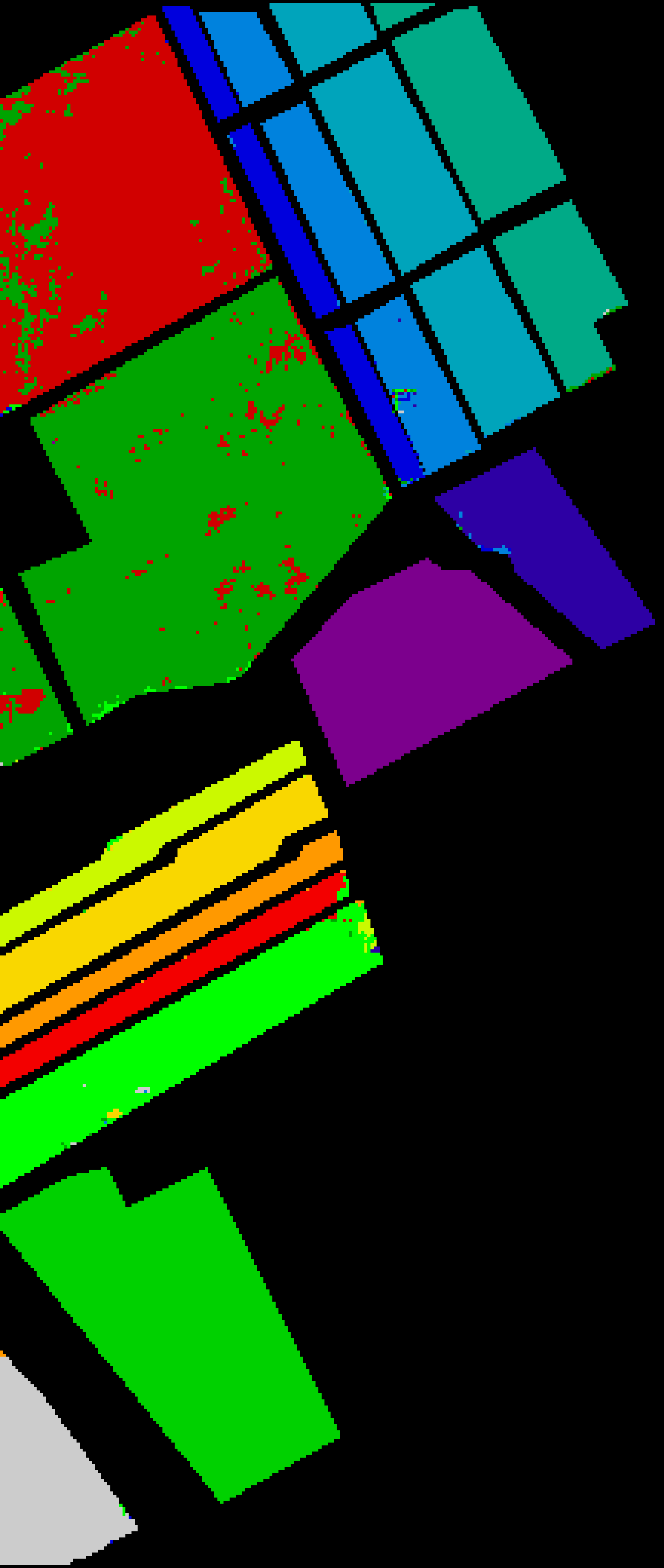}
	\caption*{WM \cite{10767233}}
    \end{subfigure}
    \begin{subfigure}{0.11\textwidth}
	\includegraphics[width=0.99\textwidth]{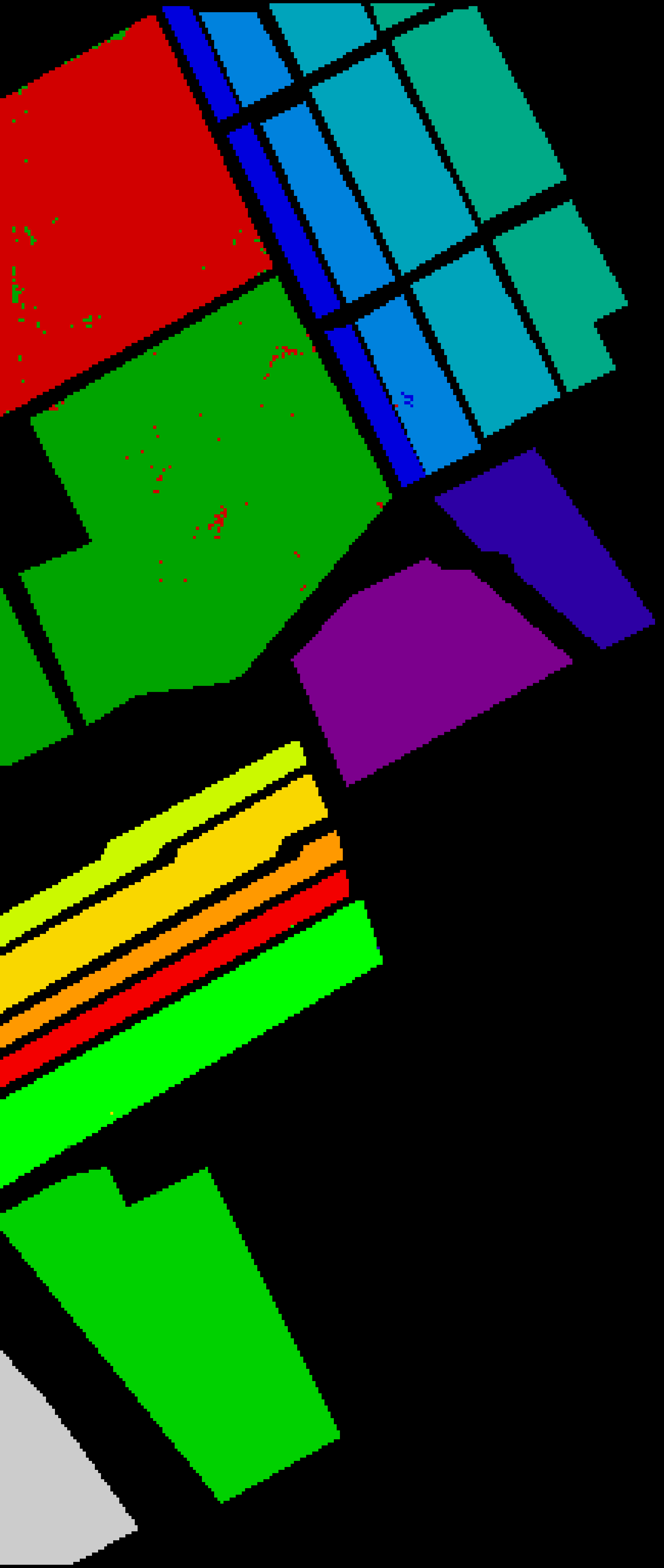}
	\caption*{Proposed}
    \end{subfigure}
\caption{\textbf{SA dataset:} Ground truth maps for comparative methods alongside the proposed ATL-SST.}
\label{SACG}
\end{figure*}
%%%%%%%%%%%%%%%%%%%%%%%%%%%%%%%

As shown in Table \ref{UHC}, the results highlight the superior performance of ATL-SST across most experimental settings. For OA, ATL-SST consistently achieves the highest values, with a maximum of 100\% in multiple scenarios, underscoring its robustness and effectiveness. Similarly, ATL-SST outperforms all comparative methods in terms of $\kappa$ and AA, achieving top scores of 97.38\% and 96.97\%, respectively. This reflects the method's ability to effectively classify diverse classes within the hyperspectral dataset. Notably, ATL-SST demonstrates significant improvement over AGCN and WaveMamba, with AGCN yielding lower performance metrics (e.g., 91.35\% OA and 90.64\% $\kappa$) in many cases. While HybViT and SST exhibit strong results, particularly in individual settings, their performance is generally less consistent compared to ATL-SST. The training time analysis also reveals ATL-SST's computational efficiency, with an average training time comparable to SST and WaveFormer, and significantly lower than methods like WaveMamba and PyFormer. The ground truth maps presented in Figure \ref{UHCG} visually depict the classification results for the UH dataset. It is evident from the maps that ATL-SST provides more precise and coherent classification results, with fewer misclassified pixels and better-defined class boundaries compared to the comparative methods. For instance, methods like AGCN and WaveMamba exhibit noticeable misclassifications and noise, whereas ATL-SST maintains high spatial and spectral accuracy. 

The results presented in Table \ref{PUC} and Figure \ref{PUCG} comprehensively compare the performance. In terms of accuracy, ATL-SST consistently outperforms all other methods, achieving the highest OA of 98.8778\%, AA of 98.1747\%, and $\kappa$ of 98.5130\%. These results are a significant improvement over the next best-performing methods, such as WaveMamba, which achieves an OA of 96.1941\%. From a computational perspective, ATL-SST maintains competitive efficiency, with a training time of 124.78s, comparable to HybViT and SST while significantly outperforming computationally intensive models like WaveMamba and WaveFormer, which require 1382.70s and 112.68s, respectively. Qualitative results presented in Figure \ref{PUCG} further validate the quantitative findings. The classification maps generated by ATL-SST are visually superior, exhibiting precise boundary delineation and accurate region classifications that closely match the ground truth. In contrast, competing methods, such as AGCN and SSMamba, show noticeable errors, particularly in regions with complex class boundaries or high spectral variability.

The comparison of HSI classification methods on the SA dataset, as detailed in Table \ref{SAC} and illustrated in Figure \ref{SACG}, highlights the superior performance of the proposed ATL-SST model. Quantitatively, ATL-SST demonstrates outstanding accuracy, achieving the highest OA of 99.4864\%, AA of 99.7641\%, and $\kappa$ of 99.4281\%. These metrics significantly surpass the next best-performing models, such as HybViT and SST, which attain OAs of 97.4912\% and 97.5318\%, respectively. From a computational perspective, ATL-SST achieves a balanced trade-off between accuracy and training efficiency. Its training time of 157.85s is competitive with models like HybViT and SST, while being significantly faster than WaveMamba and PyFormer, which require 1734.50s and 1469.81s, respectively. Qualitative results in Figure \ref{SACG} further validate ATL-SST's efficacy. The classification maps produced by ATL-SST exhibit superior boundary delineation and region specificity, closely mirroring the ground truth. In comparison, models such as SSMamba and WaveMamba show noticeable misclassifications in regions with complex spectral patterns, while transformer-based models like HybViT and PyFormer, although performing well overall, display minor boundary smoothing. 

%%%%%%%%%%%%%%%%%%%%%%%%%%%%%%%
\section{Ablation Study}

To evaluate the individual contributions of each module within the SST-ATL framework, we conduct a comprehensive ablation study on three representative datasets: PU, UH, and SA. The ablation analysis is divided into two categories: component-level evaluation and query strategy assessment.

%%%%%%%%%%%%%%%%%%%%%%%%%%%%%%%
\paragraph{\textbf{Component-Level Ablation}} First, to measure the effect of AL, we replace the hybrid uncertainty-diversity sampling strategy with random sampling. Second, to isolate the impact of diversity sampling, we retain only uncertainty-based selection. Third, to test the influence of selective model adaptation, we disable dynamic freezing and allow all transformer layers to be uniformly fine-tuned. Fourth, we examine the value of the self-calibrated attention mechanism by reverting to standard MHSA without uncertainty-guided refinement.

The results presented in Table \ref{Component} indicate that each component significantly contributes to the final performance of the model. The full SST-ATL achieves the highest OA, AA, and $\kappa$ across all datasets. The removal of AL causes the most severe degradation, OA drops by 2.66\%, 2.65\%, and 2.70\% for PU, UH, and SA respectively, highlighting the effectiveness of the proposed hybrid sampling. Excluding diversity sampling also leads to performance drops, particularly in the AA metric, reflecting reduced class coverage. The absence of dynamic freezing results in slightly reduced accuracy but increased training instability, while the removal of self-calibrated attention degrades spatial-spectral feature alignment, as evidenced by reduced $\kappa$ values. 

%%%%%%%%%%%%%%%%%%%%%%%%%%%%%%%
\begin{table}[!htb]
\centering
\caption{Component-level ablation results.}
\resizebox{\columnwidth}{!}{\begin{tabular}{l|c c c|c c c|c c c} \hline
\multirow{2}{*}{\textbf{Variant}} & \multicolumn{3}{c|}{\textbf{PU}} & \multicolumn{3}{c|}{\textbf{UH}} & \multicolumn{3}{c}{\textbf{SA}} \\ \cline{2-10}
 & OA & AA & $\kappa$ & OA & AA & $\kappa$ & OA & AA & $\kappa$ \\ \hline
\textbf{Full SST-ATL} & \textbf{98.88} & \textbf{98.17} & \textbf{98.51} & \textbf{97.58} & \textbf{96.97} & \textbf{97.38} & \textbf{99.49} & \textbf{99.76} & \textbf{99.43} \\
w/o Active Learning & 96.22 & 94.60 & 95.00 & 94.93 & 93.48 & 94.51 & 96.79 & 97.90 & 96.90 \\
w/o Diversity & 97.14 & 96.13 & 96.47 & 95.72 & 94.36 & 95.08 & 97.85 & 98.40 & 97.86 \\
w/o Dynamic Freezing & 97.31 & 96.23 & 96.60 & 96.14 & 94.72 & 95.52 & 98.12 & 98.83 & 98.10 \\
w/o Self-Calibrated Attention & 97.53 & 96.74 & 97.12 & 96.80 & 95.93 & 96.61 & 98.71 & 99.12 & 98.63 \\ \hline 
\end{tabular}}
\label{Component}
\end{table}
%%%%%%%%%%%%%%%%%%%%%%%%%%%%%%%

%%%%%%%%%%%%%%%%%%%%%%%%%%%%%%%
\paragraph{\textbf{Query Strategy Evaluation}} In addition to component analysis, we evaluate the effectiveness of our hybrid uncertainty-diversity sampling strategy against four widely used alternatives: random sampling, entropy-based uncertainty sampling, margin-based uncertainty sampling, and diversity-only sampling. All strategies are tested across four AL rounds (250 to 1000 labeled samples) on the PU, UH, and SA datasets.

As shown in Table \ref{Query}, our proposed hybrid query strategy outperforms all others in every setting. The hybrid method achieves up to 98.14\% OA on PU, 97.29\% on UH, and 98.97\% on SA with only 1000 labeled samples. The advantage is particularly evident in the early rounds, where the model learns from as few as 250 samples. For example, at 250 labels, the hybrid method exceeds random sampling by over 3\% OA on all datasets, showcasing its superior label efficiency and informativeness. These results emphasize that incorporating both uncertainty and spectral-spatial diversity into the query process leads to faster convergence, better generalization, and significantly lower annotation costs than existing AL approaches.

%%%%%%%%%%%%%%%%%%%%%%%%%%%%%%%
\begin{table}[!hbt]
\centering
\caption{Comparison of query strategies across AL iterations.}
\resizebox{\columnwidth}{!}{\begin{tabular}{l|cccc|cccc|cccc} \hline
\multirow{2}{*}{\textbf{Strategy}} & \multicolumn{4}{c|}{\textbf{PU (OA\%)}} & \multicolumn{4}{c|}{\textbf{UH (OA\%)}} & \multicolumn{4}{c}{\textbf{SA (OA\%)}} \\ \cline{2-13}
 & 250 & 500 & 750 & 1000 & 250 & 500 & 750 & 1000 & 250 & 500 & 750 & 1000 \\ \hline
Random Sampling & 90.81 & 93.54 & 95.02 & 96.22 & 88.13 & 91.95 & 93.41 & 94.93 & 91.74 & 94.27 & 95.89 & 96.79 \\
Entropy-Based Uncertainty & 91.54 & 94.38 & 95.73 & 97.14 & 89.97 & 93.22 & 94.78 & 95.72 & 92.90 & 95.83 & 96.61 & 97.85 \\
Margin Sampling & 91.23 & 94.07 & 95.49 & 96.80 & 89.51 & 92.68 & 94.33 & 95.31 & 92.46 & 95.30 & 96.20 & 97.43 \\
Diversity-Only Sampling & 92.14 & 94.96 & 96.34 & 97.04 & 90.62 & 93.75 & 95.11 & 95.96 & 93.33 & 96.08 & 96.95 & 98.00 \\
\textbf{Proposed (Hybrid)} & \textbf{93.92} & \textbf{96.23} & \textbf{97.51} & \textbf{98.14} & \textbf{92.01} & \textbf{95.04} & \textbf{96.30} & \textbf{97.29} & \textbf{94.71} & \textbf{97.42} & \textbf{98.13} & \textbf{98.97} \\ \hline
\end{tabular}}
\label{Query}
\end{table}
%%%%%%%%%%%%%%%%%%%%%%%%%%%%%%%

%%%%%%%%%%%%%%%%%%%%%%%%%%%%%%%
\section{Conclusions and Future Research Directions}

This work proposes a novel multi-stage ATL framework that integrates the SST for efficient HSI classification. The proposed framework leverages the strengths of transfer learning, AL, and SST to address key challenges in HSI classification, such as high spectral dimensionality and limited labeled data. A major contribution of this work is developing an uncertainty-diversity querying mechanism that adaptively selects the most informative and diverse samples for iterative model refinement. This not only optimizes labeling efficiency but also improves the model’s ability to generalize across varying spectral profiles. Additionally, this work introduced a dynamic freezing strategy to selectively freeze and unfreeze SST layers during the transfer learning process, ensuring an optimal balance between computational efficiency and adaptability to new spectral variations. This mechanism significantly reduces computational overhead while preserving critical learned representations, making the approach scalable for resource-constrained environments.

While this study makes substantial progress in advancing HSI classification through ATL and SST, several avenues for future exploration remain: for instance, future extensions could focus on self-supervised learning approaches to reduce reliance on labeled data. This would be particularly beneficial in HSI classification, where obtaining labeled samples is costly. Leveraging self-supervised pretext tasks or pseudo-labeling strategies could further enhance the model's learning capabilities in scenarios with limited labeled data. Although the current uncertainty-diversity-based querying has proven effective, dynamic querying strategies could be explored. For example, combining reinforcement learning with AL may allow the system to learn optimal sampling strategies over time, potentially improving both sample selection and overall learning efficiency. As HSI data can vary significantly between different sensors or geographical regions, exploring domain adaptation techniques within the ATL framework would be a valuable direction. This would allow models to adapt better to unseen datasets or shifts in data distribution, improving the robustness of the classification process in real-world scenarios.

%%%%%%%%%%%%%%%%%%%%%%%%%%%%%%%
\bibliographystyle{IEEEtran}
\bibliography{IEEEabrv,Sam}
%%%%%%%%%%%%%%%%%%%%%%%%%%%%%
\end{document}